\RequirePackage{fix-cm}
\documentclass[smallextended,twocolumn,natbib]{svjour3}
\usepackage[colorlinks,citecolor=blue,linkcolor=blue]{hyperref}
\smartqed  
\makeatletter
\def\cl@chapter{\cl@chapter \@elt {theorem}}
\def\cl@chapter{\@elt {theorem}}

\patchcmd{\NAT@citex}
  {\@citea\NAT@hyper@{%
     \NAT@nmfmt{\NAT@nm}%
     \hyper@natlinkbreak{\NAT@aysep\NAT@spacechar}{\@citeb\@extra@b@citeb}%
     \NAT@date}}
  {\@citea\NAT@nmfmt{\NAT@nm}%
   \NAT@aysep\NAT@spacechar\NAT@hyper@{\NAT@date}}{}{}

\patchcmd{\NAT@citex}
  {\@citea\NAT@hyper@{%
     \NAT@nmfmt{\NAT@nm}%
     \hyper@natlinkbreak{\NAT@spacechar\NAT@@open\if*#1*\else#1\NAT@spacechar\fi}%
       {\@citeb\@extra@b@citeb}%
     \NAT@date}}
  {\@citea\NAT@nmfmt{\NAT@nm}%
   \NAT@spacechar\NAT@@open\if*#1*\else#1\NAT@spacechar\fi\NAT@hyper@{\NAT@date}}
  {}{}
\makeatother

\usepackage{footmisc} 
\usepackage[dvipsnames]{xcolor,colortbl}
\usepackage{soul}
\usepackage{changes}
\usepackage{graphicx}
\usepackage{amssymb}
\usepackage{amsmath}
\usepackage{array}
\usepackage{lineno}
\usepackage{comment}
\usepackage{multirow}
\usepackage{cite}
\usepackage{threeparttable,booktabs}
\usepackage{expl3,xparse}
\usepackage{etoolbox}
\renewcommand{\tablename}{Table}
\usepackage{subcaption}
\renewcommand{\figurename}{Fig.} 
\usepackage[export]{adjustbox}
\usepackage{diagbox}
\usepackage{tikz}
\usepackage{totcount}
\usepackage{overpic}
\newtotcounter{citenum}
\usepackage{footnotebackref}
\usepackage{pifont}
\usepackage{bbm}
\usepackage{tabstackengine}
\usepackage{makecell}

\usepackage{cleveref}
\definecolor{applegreen}{rgb}{0.0, 0.5, 0.0}
\definecolor{cadmiumred}{rgb}{0.89, 0.0, 0.13}
\definecolor{LightGrey}{rgb}{0.9,0.9,0.9}
\definecolor{babyblue}{HTML}{7EA6E0}
\definecolor{fadedred}{HTML}{EA6B66}
\definecolor{pastelorange}{HTML}{FFB570}
\definecolor{pastelteal}{HTML}{67AB9F}
\definecolor{pastelpurple}{HTML}{B5739D}

\newcommand\tstrut{\rule{0pt}{2.4ex}}
\newcommand\bstrut{\rule[-1.0ex]{0pt}{0pt}}

\newcommand\blfootnote[1]{%
  \begingroup
  \renewcommand\thefootnote{}\footnote{#1}%
  \addtocounter{footnote}{-1}%
  \endgroup
}

\newtotcounter{citnum} 
\def\oldbibitem{} \let\oldbibitem=\bibitem
\def\bibitem{\stepcounter{citnum}\oldbibitem}

\newcommand{\pcite}[1]{\textcolor{gray}{\citep{#1}}}
\newcommand{\tcite}[1]{\textcolor{gray}{\citet{#1}}}

\usepackage{balance}
\usepackage{newpxtext}

\makeatletter
\newcommand{\ie}{\emph{i.e.}\@ifnextchar.{\!\@gobble}{}}
\newcommand{\eg}{\emph{e.g.}\@ifnextchar.{\!\@gobble}{}}
\newcommand{\etc}{etc\@ifnextchar.{}{.\@}}

\makeatother

\begin{document}

\sloppy

\title{About Time: Advances, Challenges, and Outlooks of Action Understanding}

\author{Alexandros Stergiou \and Ronald Poppe}
\institute{A. Stergiou \at Faculty of Electrical Engineering, Mathematics and Computer Science at the University of Twente, Drienerlolaan 5, 7522 NB Enschede, The Netherlands   
\and
R.Poppe \at Department of Information and Computing Sciences at Utrecht University, Princetonplein 5, 3584 CC Utrecht, The Netherlands
}

\maketitle

\begin{abstract}
We have witnessed impressive advances in video action understanding. Increased dataset sizes, variability, and computation availability have enabled leaps in performance and task diversification. Current systems can provide coarse- and fine-grained descriptions of video scenes, extract segments corresponding to queries, synthesize unobserved parts of videos, and predict context across multiple modalities. This survey comprehensively reviews advances in uni- and multi-modal action understanding across a range of tasks. We focus on prevalent challenges, overview widely adopted datasets, and survey seminal works with an emphasis on recent advances. We broadly distinguish between three temporal scopes: (1) recognition tasks of actions observed in full, (2) prediction tasks for ongoing partially observed actions, and (3) forecasting tasks for subsequent unobserved action(s). This division allows us to identify specific action modeling and video representation challenges. Finally, we outline future directions to address current shortcomings.
\end{abstract}

\keywords{Action Understanding \and Action Recognition \and Action Prediction \and Action Anticipation}

\newpage

\begin{figure*}[t]
    \centering
    \begin{overpic}[width=\linewidth]{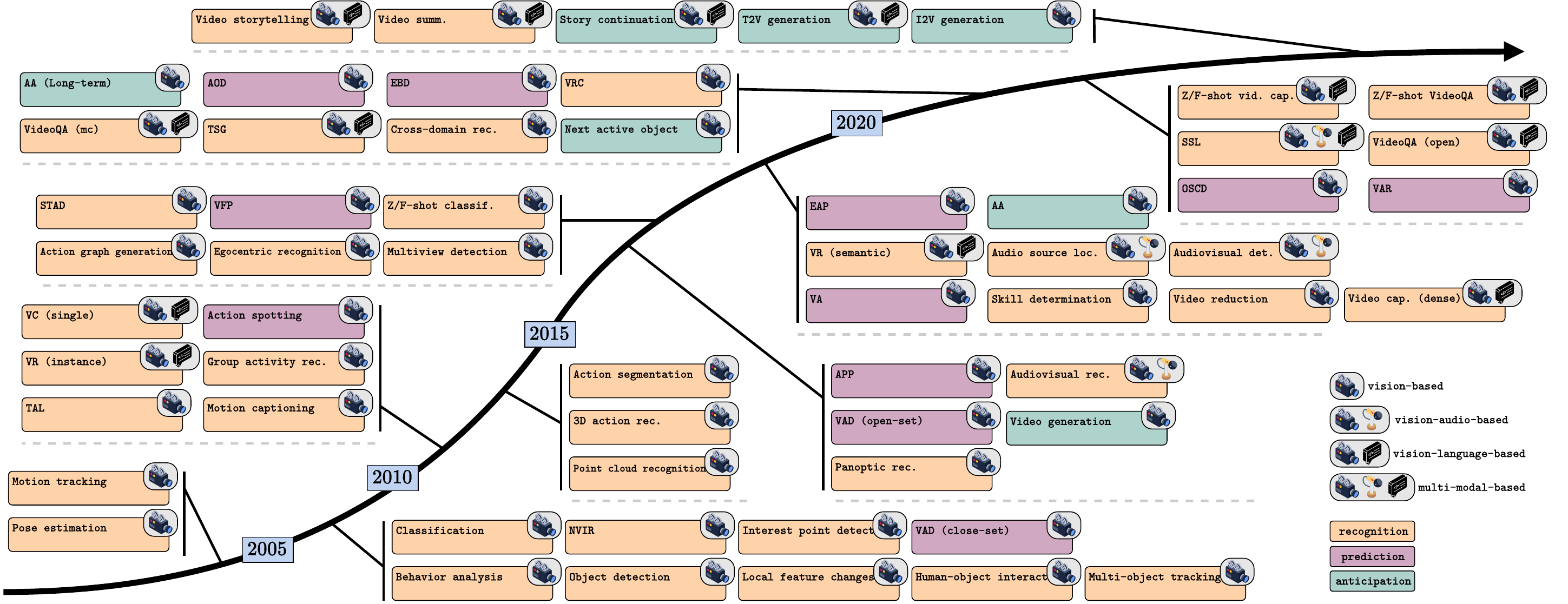}
    \put (0.8,3.53) {\fontsize{4.3}{4}\selectfont \citet{wren1997pfinder}} 
    \put (0.8,6.55) {\fontsize{4.3}{4}\selectfont \citet{bobick2001recognition}} 

    \put (25.25,3.38) {\fontsize{4.3}{4}\selectfont \citet{schuldt2004recognizing}} 
    \put (36.3,3.38) {\fontsize{4.3}{4}\selectfont \citet{vinciarelli2012bridging}} 
    \put (47.35,3.38) {\fontsize{4.3}{4}\selectfont \citet{weinland2010makingaction}} 
    \put (58.4,3.38) {\fontsize{4.3}{4}\selectfont \citet{kim2009observe}} 
    \put (36.3,0.36) {\fontsize{4.3}{4}\selectfont \citet{rehg2013decoding}} 
    \put (25.25,0.36) {\fontsize{4.3}{4}\selectfont \citet{jain2015whatdo}} 
    \put (47.35,0.36) {\fontsize{4.3}{4}\selectfont \citet{wang2013dense}} 
    \put (58.4,0.36) {\fontsize{4.3}{4}\selectfont \citet{yao2010modeling}} 
    \put (69.45,0.36) {\fontsize{4.3}{4}\selectfont \citet{benfold2011stable}} 

    \put (1.65,11.2) {\fontsize{4.3}{4}\selectfont \citet{gaidon2013temporal}} 
    \put (13.2,11.2) {\fontsize{3.7}{4}\selectfont \citet{takano2015statistical}} 
    \put (1.65,14.2) {\fontsize{4.3}{4}\selectfont \citet{song2011multiple}} 
    \put (13.2,14.2) {\fontsize{4.3}{4}\selectfont \citet{choi2012unified}} 
    \put (1.65,17.1) {\fontsize{4.3}{4}\selectfont \citet{guadarrama2013youtube2text}} 
    \put (13.2,17.1) {\fontsize{3.9}{4}\selectfont \citet{hoai2014max}} 

    \put (36.55,13.32) {\fontsize{4.3}{4}\selectfont \citet{zhou2013hierarchical}} 
    \put (36.55,10.35) {\fontsize{4.3}{4}\selectfont \citet{wang2014learning}} 
    \put (36.55,7.39) {\fontsize{4.3}{4}\selectfont \citet{oreifej2013hon4d}} 

    \put (53.25,13.32) {\fontsize{4.3}{4}\selectfont \citet{kataoka2016recognition}} 
    \put (53.25,10.34) {\fontsize{4.3}{4}\selectfont \citet{xu2015learning}} 
    \put (53.24,7.38) {\fontsize{4.3}{4}\selectfont \citet{joo2015panoptic}} 
    \put (64.38,13.32) {\fontsize{4.3}{4}\selectfont \citet{aytar2016soundnet}} 
    \put (64.38,10.34) {\fontsize{4.3}{4}\selectfont \citet{vondrick2016generating}} 

    \put (2.50,21.16) {\fontsize{4.3}{4}\selectfont \citet{shang2017video}} 
    \put (13.65,21.16) {\fontsize{4.3}{4}\selectfont \citet{damen2018scaling}} 
    \put (24.65,21.16) {\fontsize{4.3}{4}\selectfont \citet{baque2017deep}} 
    \put (2.50,24.17) {\fontsize{4.3}{4}\selectfont \citet{singh2017online}} 
    \put (13.65,24.17) {\fontsize{4.3}{4}\selectfont \citet{liang2017dual}} 
    \put (24.65,24.17) {\fontsize{4.3}{4}\selectfont \citet{zhou2018towards}} 
    
    \put (51.65,24.12) {\fontsize{4.3}{4}\selectfont \citet{gammulle2019predicting}} 
    \put (63.2,24.12) {\fontsize{4.0}{4}\selectfont \citet{furnari2019would}} 
    \put (51.65,21.11) {\fontsize{4.3}{4}\selectfont \citet{wray2021semantic}} 
    \put (63.2,21.11) {\fontsize{3.1}{4}\selectfont \citet{arandjelovic2018objects}} 
    \put (74.9,21.11) {\fontsize{4.3}{4}\selectfont \citet{gao2020listen}} 
    \put (51.65,18.18) {\fontsize{4.3}{4}\selectfont \citet{dwibedi2018temporal}} 
    \put (63.2,18.18) {\fontsize{4.3}{4}\selectfont \citet{doughty2018s}} 
    \put (74.85,18.18) {\fontsize{4.3}{4}\selectfont \citet{korbar2019scsampler}} 
    \put (85.95,18.18) {\fontsize{4.3}{4}\selectfont \citet{mun2019streamlined}} 

    \put (1.5,31.95) {\fontsize{4.3}{4}\selectfont \citet{gong2022future}} 
    \put (13.18,31.95) {\fontsize{4.3}{4}\selectfont \citet{ragusa2021meccano}} 
    \put (24.88,31.95) {\fontsize{4.3}{4}\selectfont \citet{shou2021generic}} 
    \put (35.98,31.95) {\fontsize{4.3}{4}\selectfont \citet{hu2022transrac}} 
    \put (1.5,28.95) {\fontsize{4.3}{4}\selectfont \citet{yang2021just}} 
    \put (13.18,28.95) {\fontsize{4.3}{4}\selectfont \citet{wang2022negative}} 
    \put (24.88,28.95) {\fontsize{4.3}{4}\selectfont \citet{pan2020adversarial}} 
    \put (35.98,28.95) {\fontsize{4.3}{4}\selectfont \citet{dessalene2021forecasting}} 

    \put (75.3,31.11) {\fontsize{4.3}{4}\selectfont \citet{tewel2022zero}} 
    \put (75.3,28.17) {\fontsize{4.3}{4}\selectfont \citet{bachmann2022multimae}} 
    \put (75.3,25.23) {\fontsize{4.3}{4}\selectfont \citet{souvcek2022look}} 
    \put (87.5,31.11) {\fontsize{4.3}{4}\selectfont \citet{yang2022zero}} 
    \put (87.5,28.17) {\fontsize{4.3}{4}\selectfont \citet{ko2023open}} 
    \put (87.5,25.23) {\fontsize{4.3}{4}\selectfont \citet{liang2022visual}} 

    \put (12.45,35.94) {\fontsize{4.3}{4}\selectfont \citet{han2024autoadiii}} 
    \put (24.1,35.94) {\fontsize{4.3}{4}\selectfont \citet{wang2024omnivid}} 
    \put (35.75,35.94) {\fontsize{4.3}{4}\selectfont \citet{pan2024synthesizing}} 
    \put (47.4,35.94) {\fontsize{4.3}{4}\selectfont \citet{videoworldsimulators2024}} 
    \put (58.45,35.94) {\fontsize{4.3}{4}\selectfont \citet{renconsisti2v}} 
    \end{overpic}
    \vspace{1em}
    \resizebox{\linewidth}{!}{
    \begin{tabular}{l l l l l}
        \multicolumn{5}{l}{\textbf{Abbreviations used}} \\
         \textbf{NVIR}: Non-Verbal Interaction Recognition & 
         \textbf{VAD}: Video Anomaly Detection & 
         \textbf{VC}: Video Captioning &
         \textbf{VR}: Video Retrieval &
         \textbf{TAL}: Temporal Action Localization \\
         \textbf{APP}: Action Progress Prediction & 
         \textbf{EAP}: Early Action Prediction &
         \textbf{VA}: Video Alignment &
         \textbf{AA}: Action Anticipation & 
         \textbf{STAD}: SpetaoTemporal Action Detection \\ 
         \textbf{VFP}: Video Frame Prediction & 
         \textbf{Z/F}: Zero- and Few-shot & 
         \textbf{AOD}: Active Object Detection & 
         \textbf{TSG}: Temporal Sentence Grounding & 
         \textbf{EBD}: Event Boundary Detection \\ 
         \textbf{VRC}: Video Repetition Counting & 
         \textbf{OSCD}: Object State Change Detection & 
         \textbf{VAR}: Video Abductive Reasoning &
         \textbf{T2V}: Text to Video Generation &
         \textbf{I2V}: Image to Video Generation \\ 
    \end{tabular}
    }
    \caption{\textbf{Action understanding historical overview}. We present popular tasks over time. Landmark papers are selected by their relevance to the period's trends. Most tasks remain popular today.}
    \label{fig:roadmap}
\end{figure*}
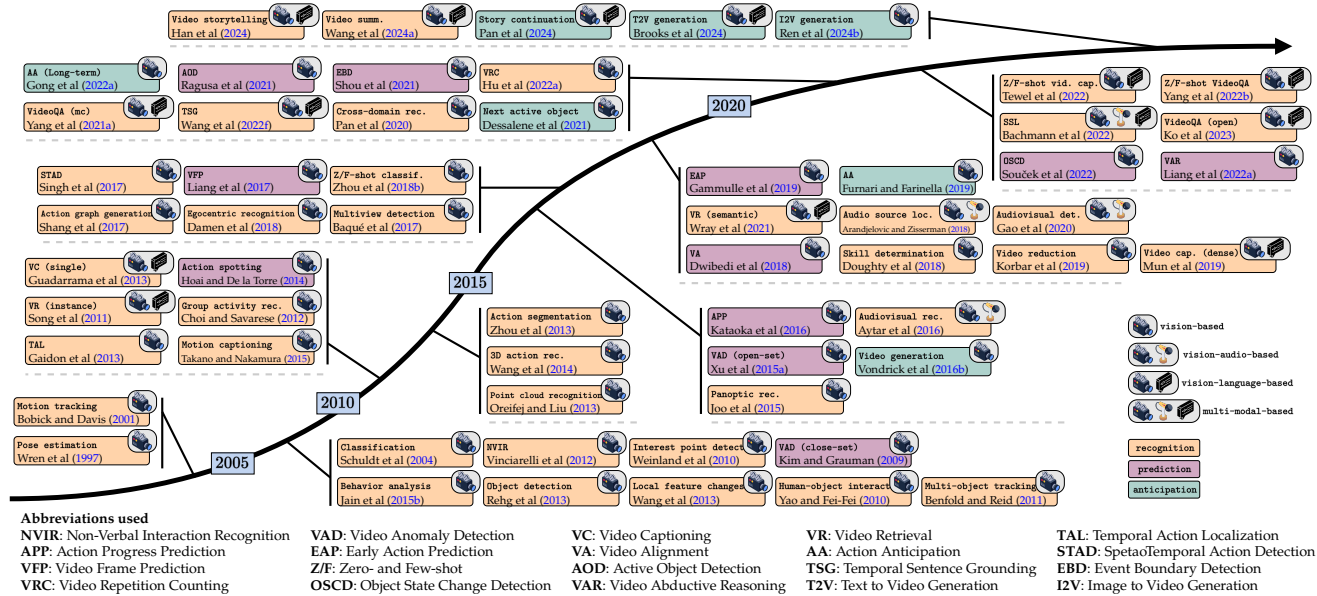

\tableofcontents

\section{Introduction}
\label{sec:intro}

For decades, analyzing human actions in videos has been of particular interest to the computer vision community. Videos are prominent in both our social and professional lives. Over time, the analysis of actions has shifted from the well-understood task of action recognition towards the fundamental and broader area of action understanding. Shown in \Cref{fig:roadmap}, action understanding now includes diverse tasks based on prediction and anticipation with multimodal inputs. The unique challenges and novel computation paradigms are the core focus of our survey. 

In developmental psychology, action understanding has been explored across several psychological aspects \pcite{thompson2019conceptualizing}: 

\noindent
\textbf{The ability to understand the action performed} relates to differentiating between analogous actions \pcite{gallese1996action,jeannerod1994representing} and conceptualizing \emph{how} an action is performed \pcite{spunt2011identifying}. 

\noindent
\textbf{Determining the goal of the action} has been studied in the context of immediate goals \pcite{calvo2005action,kohler2002hearing,rizzolatti2001neurophysiological} in relation to motor functions for the execution of actions and the sensory perception of actions performed by others.

\noindent
\textbf{Determining the actor's intention} refers to identifying high-level goals and motivations to perform actions \pcite{kilner2011more}. Intentions have been defined as the sequential grouping of individual actions \pcite{fogassi2005parietal} and their abstract associated target \pcite{uithol2011understanding}.

\subsection{Taxonomy of this survey}
\label{sec:intro::taxonomy}

\begin{figure*}[t]
    \centering
    \includegraphics[width=\linewidth,trim={1cm 0 6cm 0},clip]{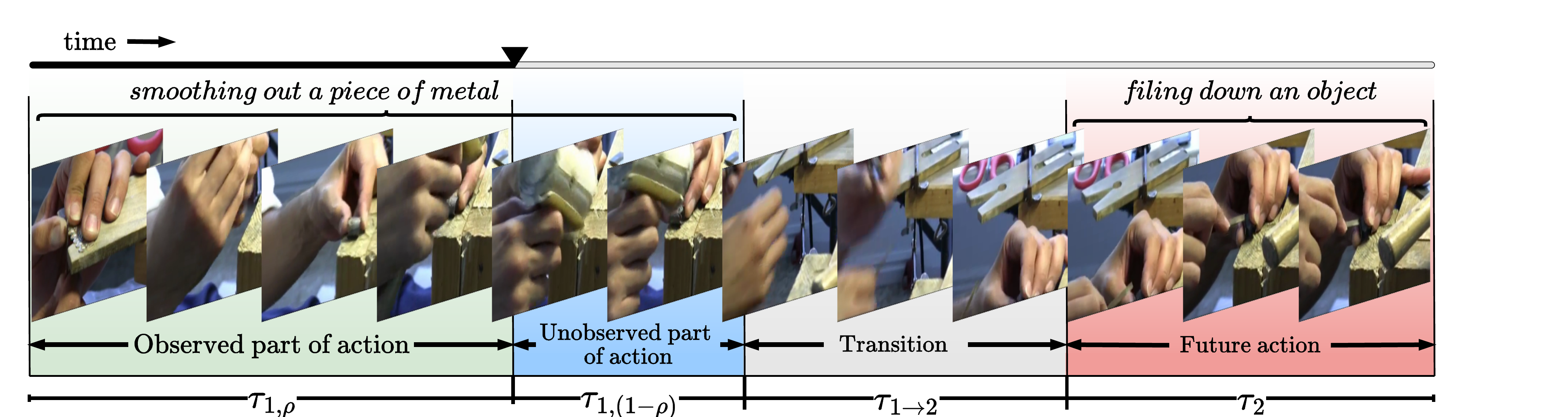}
    \caption{\textbf{Action understanding tasks}. The progress of the video is indicated by the top bar. From the currently performed action of total duration $\tau_1$, only the $\tau_{1,\rho}<\tau_1 $ part is readily observable. After a transition period $0\leq\tau_{1 \rightarrow 2}$, another action is performed with duration $\tau_2$. \textbf{Action recognition} tasks consider full observations of the action at $\tau_1$. \textbf{Action prediction} uses only part $\tau_{1,\rho}$ of the ongoing action. \textbf{Action forecasting} uses current action at $\tau_1$ to predict future actions. Video example sourced from \tcite{wang2019vatex}.} 
    \label{fig:tasks}
\end{figure*}

Inspired by the cognitive aspects of action understanding, we define three broad \emph{temporal scopes} to group seminal machine vision action understanding tasks. We visualize action sequence progression in~\Cref{fig:tasks} with a currently (partially) performed action followed by a subsequent action. Tasks that require an action to be \emph{observed in full} are broadly referred to as \textbf{recognition} tasks and infer information such as the action categories or high-level semantics. \textbf{Predictions} about the ongoing actions are made from partial observations of actions \emph{not yet completed}. \textbf{Forecasting} tasks use the currently observed action(s) to reason about future actions \emph{not yet observed}. We discuss relevant previous surveys for each of these three temporal scopes and overview their focus in \Cref{tab:surveys}.

\noindent
\textbf{Recognition}. As seen by the top rows in  \Cref{tab:surveys}, early works on action recognition have primarily focused on motion modeling. \tcite{aggarwal1994articulated} used a taxonomy of rigidness and subsequently \pcite{aggarwal1998nonrigid} introduced subdivisions based on prior knowledge of the object's shape. \tcite{cedras1995motion} and later \pcite{moeslund2001survey} discussed temporal modeling approaches in the context of classification and tracking. As overviewed by \tcite{buxton2003learning}, tracking has also been applied to more complex tasks such as behavior analysis or non-verbal human interactions. 
Subsequent overviews were more task-oriented, focusing on action classification and localization \pcite{weinland2011survey}, behavior understanding \pcite{chaaraoui2012review}, and surveillance applications \pcite{vishwakarma2013survey} emerged based on later advancements. Simultaneously, \tcite{turaga2008machine} and \tcite{poppe2010survey} discussed approaches addressing atomic actions and group activities. \tcite{herath2017going} provided an initial summary of approaches using learned features for action recognition. Following surveys covered adaptations of deep learning approaches for topics such as depth-based motion recognition \pcite{wang2018rgb}, activity recognition \pcite{beddiar2020vision}, human-human interactions \pcite{stergiou2019analyzing}, and pose estimation \pcite{zheng2020deep}. \tcite{sun2022human} reviewed approaches across modalities, combining motion features, audio, and vision. More recently, \tcite{selva2023video} discussed attention-based approaches for video tasks while \tcite{schiappa2023self} focused on self-supervised (SSL) approaches. \tcite{madan2024foundation} presented a comprehensive overview of language-enabled action understanding models, focusing on video foundation models.

\noindent
\textbf{Prediction}. Recent advancements in action recognition have also sparked interest in predictive tasks from partial observations. \tcite{rasouli2020deep} discussed four main domains of predictive models including video, action, trajectory, and motion prediction. \tcite{kong2022human} described recent action recognition and prediction advancements. They focused on applications in domains such as robot vision, surveillance, and driver behavior prediction. The surveys of \tcite{dhiman2019review} and \tcite{ramachandra2020survey} overviewed predictive methods specifically for anomaly detection. As shown in \Cref{tab:surveys} overviews on these tasks are scarce.

\noindent
\textbf{Forecasting}. Action forecasting tasks have become prevalent parts of action understanding research. \tcite{rodin2021predicting} discussed future action anticipation in egocentric videos. \tcite{zhong2023survey} overviewed short and long-term action anticipation methods. \tcite{hu2022online} provided a review of online and anticipation works. Recently, \tcite{plizzari2024outlook} discussed challenges in egocentric videos and presented future directions for multiple tasks including forecasting.
  
Despite their extensive coverage, prior surveys focus on specific aspects of action understanding. As shown in \Cref{tab:surveys}, a critical overview that holistically explores action understanding is currently missing in the literature.

\begin{table*}[t]
    \centering
    \caption{\textbf{Action understanding surveys through the years}. For each survey, we note the year and number of papers covered. We identify the coverage of temporal scopes, including recognition (Rec.), prediction (Pred.), and forecasting (For.). Broad objectives include multimodality (MM), self-supervision (SSL), and multi-view (MV). We further highlight other specific tasks discussed, such as human interactions (HI), long video understanding (LVU). Scopes/objectives/tasks addressed partially within surveys are denoted with (partial), and the main focus is denoted with \ding{52}.
    }
    \resizebox{\textwidth}{!}{
    \begin{tabular}{l c c c c c l c c c l c c}
    \toprule
    \multirow{2}{*}{Author(s)} & 
    \multirow{2}{*}{Year} &
    \multirow{2}{*}{\#Papers} & \multicolumn{3}{c}{Temporal Scope} & $\;$ & \multicolumn{3}{c}{Objectives} & $\;$ & \multicolumn{2}{c}{Tasks} \bstrut\\ \cline{4-6}\cline{8-10}\cline{12-13}
    & & &
    Rec. & Pred. & For. & $\;$ & MM & SSL & MV & $\;$ & HI & LVU  \tstrut\\
    \midrule
    \citet{aggarwal1994articulated} & 1994 & 
    69 &  
    (partially) & 
      & 
      && 
      & 
      & 
      && 
      & \\
    \citet{cedras1995motion} & 1995 &
    76 &  
      (partially) & 
      & 
      && 
      & 
      & 
      && 
      & \\
    \citet{aggarwal1998nonrigid} & 1998 & 
    104 &  
      (partially) & 
      & 
      && 
      & 
      & 
      && 
      & \\ 
    \citet{aggarwal1999human} & 1999 & 
    51 &  
      (partially) & 
      & 
      && 
      & 
      & 
      && 
      & \\
    \citet{moeslund2001survey} & 2001 &
    155 &  
      (partially) & 
      & 
      && 
      & 
      & 
      && 
      & \\
    \citet{buxton2003learning} & 2003 & 
    88 &  
      (partially) & 
      & 
      && 
      & 
      &
      &&
      \ding{52}
      & \\
    \citet{moeslund2006survey} & 2006 & 
    424 &  
      \ding{52} & 
      & 
      && 
      (partially) & 
      & 
      && 
      & \\
    \citet{yilmaz2006object} & 2006 &
    160 &  
      (partially) & 
      (partially) & 
      && 
      & 
      & 
      && 
      (partially) &
      \\
    \citet{turaga2008machine} & 2008 &
    144 &  
      \ding{52} & 
      & 
      && 
      (partially) & 
      & 
      && 
      \ding{52} &
      \\
    \citet{poppe2010survey} & 2010 & 
    180 &  
      \ding{52} & 
      & 
      && 
      & 
      & 
      && 
      & \\
    \citet{weinland2011survey} & 2011 & 
    153 &  
      \ding{52} & 
      & 
      && 
      (partially) & 
      & 
      (partially) &&
      \ding{52} &\\
    \citet{chaaraoui2012review} & 2012 &
    123 &  
      \ding{52} & 
      & 
      && 
      (partially) & 
      & 
      \ding{52} &&
      \ding{52} &\\
    \citet{metaxas2013review} & 2013 &
    188 &  
      & 
      & 
      && 
      & 
      & 
      (partially) && 
      \ding{52} & \\
    \citet{vishwakarma2013survey} & 2013 &
    231 &  
      \ding{52} & 
      & 
      && 
      (partially) & 
      & 
      && 
      & \\
    \citet{herath2017going} & 2017 & 
    161 &  
      \ding{52} & 
      & 
      && 
      & 
      & 
      && 
      & \\
    \citet{wang2018rgb} & 2018 & 
    182 &  
      \ding{52} & 
      (partially) & 
      && 
      (partially) & 
      & 
      (partially) && 
      (partially) & \\
    \citet{dhiman2019review} & 2019 & 
    208 &  
      & 
      & 
      \ding{52} && 
      (partially) & 
      & 
      &&
      (partially) & \\
    \citet{hussain2019different} & 2019 & 
    141 &  
      \ding{52} & 
       & 
      && 
      \ding{52} & 
      & 
      (partially) && 
      \ding{52} & \\
    \citet{stergiou2019analyzing} & 2019 & 
    178 &  
      \ding{52} & 
      & 
      && 
      & 
      & 
      && 
      \ding{52} & \\
    \citet{yao2019review} & 2019 &
    106 &  
      \ding{52} & 
      & 
      && 
      & 
      & 
      && 
      & \\
    \citet{zhang2019comprehensive} & 2019 &
    127 &  
      \ding{52} & 
      & 
      && 
      & 
      & 
      && 
      & \\
    \citet{beddiar2020vision} & 2020 & 
    237 &  
      \ding{52} & 
      & 
      (partially) && 
      & 
      & 
      (partially) && 
      \ding{52} & 
      \\
      \citet{ramachandra2020survey} & 2020 &
    109 &  
      & 
      \ding{52} & 
      && 
      & 
      & 
      && 
      & \\
    \citet{zheng2020deep}& 2020 & 
    317 &  
      & 
      & 
      && 
      \ding{52} & 
      & 
      && 
      & \\
    \citet{rasouli2020deep} & 2020 &
    333 &  
      & 
      \ding{52} & 
      && 
      & 
      & 
      && 
      & \\
    \citet{pareek2021survey} & 2021 & 
    218 &  
      \ding{52} & 
      & 
      && 
      (partially) & 
      & 
      && 
      & \\
    \citet{rodin2021predicting}& 2021 & 
    156 &  
      & 
      & 
      \ding{52} && 
      (partially) & 
      & 
      \ding{52} && 
      & 
      \ding{52} \\
    \citet{song2021human} &
    2021 &
    157 &  
      \ding{52} & 
      & 
      && 
      & 
      & 
      && 
      & \\
    \citet{sun2022human} & 2022 & 
    503 &  
      \ding{52} & 
      & 
      && 
      (partially) & 
      \ding{52} & 
      && 
      & \\
    \citet{kong2022human} & 2022 & 
    337 &  
      \ding{52} & 
      (partially) & 
      && 
      & 
      & 
      \ding{52} && 
      (partially) & \\ 
    \citet{hu2022online} & 2022 & 
    168 &  
      \ding{52} & 
      & 
      \ding{52} && 
      & 
      & 
      && 
      \ding{52} & 
      (partially) \\
    \citet{oprea2022review} & 2022 & 
    211 &  
      \ding{52} & 
      \ding{52} & 
      && 
      & 
      & 
      && 
      & \\
    \citet{schiappa2023self} & 2023 &
    216 &  
      \ding{52} & 
      & 
      && 
      & 
      \ding{52} & 
      \ding{52} && 
      & 
      \ding{52}\\
    \citet{selva2023video} & 2023 &
    209 &  
      \ding{52} & 
      & 
      && 
      & 
      & 
      && 
      & \\
    \citet{wang2023temporal} & 2023 & 229 & \ding{52} & 
      & 
      && 
      (partially) & 
      & 
      (partially) && 
      & \\
    \citet{zhong2023survey} & 2023 & 
    207 &
      & 
      & 
      \ding{52} && 
      \ding{52} & 
      \ding{52} & 
      \ding{52} && 
      & 
      \ding{52} \\  
    \citet{ding2023temporal} & 2023 & 
    168 &  
      \ding{52} & 
      & 
      && 
      & 
      & 
      \ding{52} && 
      & 
      (partially) \\ 
    \citet{tang2023video} & 2023 &
    338 &
      \ding{52} & 
      & 
      && 
      \ding{52} & 
      \ding{52} & 
      (partially) && 
      & 
      \ding{52} \\ 
    \citet{plizzari2024outlook} & 2024&
    367 &
      \ding{52} & 
      & 
      \ding{52} && 
      \ding{52} & 
      & 
      \ding{52} && 
      \ding{52} & 
      \ding{52}  \\
    \citet{madan2024foundation} & 2024&
    367 &
      \ding{52} & 
      & 
      && 
      \ding{52} & 
      \ding{52} & 
      && 
      & 
      \ding{52}  \\
    \citet{lai2024human} & 2024 & 
    202&
    &
    (partially) &
    \ding{52} &&
    \ding{52} &
    &
    \ding{52} &&
    &
    \ding{52} \\
    \midrule
    Stergiou and Poppe (\textbf{this survey}) & 2025 & \textbf{\total{citnum}} 
    & 
    \ding{52} & 
    \ding{52} & 
    \ding{52} && 
    \ding{52} & 
    \ding{52} & 
    \ding{52} && 
    \ding{52} & 
    \ding{52} \\
    \end{tabular}
    }
    \label{tab:surveys}
\end{table*}

This survey fills this void by focusing on advancements across a broad range of action understanding tasks. We do this from a temporal perspective. We survey general approaches for modeling actions in videos over the years in~\Cref{sec:modeling}, and discuss common datasets and benchmarks in~\Cref{sec:datasets}. We then detail recognition tasks in~\Cref{sec:recognition}, predictive tasks in~\Cref{sec:prediction}, and forecasting tasks in~\Cref{sec:forecasting}. Based on the temporal scopes, we then outline the main challenges and provide future directions in~\Cref{sec:directions}. We conclude in~\Cref{sec:conclusion}.
\section{Modeling actions in videos}
\label{sec:modeling}

In this section, we define two general groups of approaches for encoding videos without explicitly relating them to tasks. We start with characterizing key challenges in \Cref{sec:modeling::challenges}. Approaches discussed in~\Cref{sec:modeling::separate} model spatial and temporal information separately, while works overviewed in~\Cref{sec:modeling::joint} use joint spatiotemporal representations. 

\subsection{Challenges in action representation}
\label{sec:modeling::challenges}

The diversity of the video input poses several challenges. \textbf{Intra-class variations} in the visual appearance of actions of the same category across videos can be due to viewpoint, occlusions, background noise, or lighting conditions. The performances and durations of actions can also significantly deviate. Such variations appear across datasets  \pcite{grauman2022ego4d,kay2017kinetics,miech2019howto100m,soomro2012ucf101}. Training/test set instance distribution variance can also significantly impact the performance and overall generalization of the learned semantics. Challenging action instances can be traced to feature representations further from the training set distribution in such cases.

Since action understanding tasks are increasingly semantic, we also face challenges in the diversity and granularity of the target outputs. Interpretation of the visual input, and sometimes the lack of observable information, increasingly requires higher-level understanding. Consequently, the relation between visual input and model output becomes more complex. \textbf{Vocabulary limitations} present challenges as action categories are often finite. The generalization of models to \emph{open-set} or \emph{cross-domain} settings primarily depends on the similarity between seen and unseen instances. Limited \textit{inter-class variation} further affects good representation performance of rare coarse-grained concepts of visually similar actions. This issue is more prevalent for tasks that require fine-grained semantic granularities.

\subsection{Separating visual and temporal information}
\label{sec:modeling::separate}
We first discuss approaches that process visual and temporal information independently. 

\noindent
\textbf{Tracking and template matching}. Early works \pcite{bobick2001recognition} have applied template matching to spatially and temporally localize motions. These approaches relied on view-specific representations of movements, in the form of templates, to capture underlying motion similarity across action instances. Templates have been explored through local patches \pcite{shechtman2005space}, correlation filters \pcite{rodriguez2008action}, and voxels \pcite{ke2007spatio}. Another line of research has considered temporal pattern discovery by directly tracking visual features over time \pcite{cipolla1990dynamic,isard1998condensation,rohr1994towards}. Template approaches have relied on assumptions such as static backgrounds, fixed camera views, and linear motions that limit the exploration of intra-class variability.

\noindent
\textbf{Local descriptors}. Motivated by the observation that actions can be characterized through appearance changes over time, a set of approaches aims to associate per-frame changes from local descriptor features to action categories. Pose primitives \pcite{thurau2008pose}, temporal bins \pcite{nowozin2007discriminative}, pictorial structures \pcite{tran2012part}, and graphical structures of the actions \pcite{ni2014multiple} have been explored as descriptors for local action features. \tcite{mikolajczyk2008action} clustered an ensemble of local features to tree representations and related them to action categories. Other approaches \pcite{gupta2009observing,yao2010modeling} cast action recognition as a two-step structural connectivity task by recognizing parts of objects and understanding actions through pose.
Several methods have extended this notion to individual regions \pcite{ikizler2010object}, poselet clusters \pcite{pishchulin2013strong}, decision trees \pcite{rahmani2014real}, and covariance matrices \pcite{kviatkovsky2014online}.

\noindent
\textbf{Spatial convolutions}. Convolutions can efficiently extract local patterns from visual inputs. An early application of Convolutional Neural Networks (CNNs) to video \pcite{karpathy2014large} temporally fused spatial frame embeddings over pre-defined sets of layers. Others explored the factorization of frame embeddings \pcite{sun2015human}, frame ranking \pcite{fernando2015modeling}, pooling \pcite{fernando2016rank}, salient region focus \pcite{girdhar2017attentional, zong2021motion}, and relation reasoning between neighboring frames \pcite{zhou2018temporal}. \tcite{le2011learning} spatially convolved videos over combinations of the spatial and temporal dimensions. Seminal efforts focused on single volumes to represent motion \pcite{bilen2016dynamic,chung2016signs,iosifidis2012view} or learned the correlation and exclusion between action classes \pcite{hoai2015improving}. \tcite{tran2018closer} proposed convolutional blocks based on spatial (2D) and temporal (1D) kernels to create more efficient video models. \tcite{lin2019tsm} reduced redundancies by shifting features at subsequent frames, while later adaptations also included conditional gates \pcite{sudhakaran2020gate}.

\noindent
\textbf{Temporal recursion}. A parallel line of research has focused on extracting motion patterns with recurrent layers \pcite{ballas2015delving,dwibedi2018temporal,perrett2019ddlstm,yue2015beyond,ullah2017action}, from the static frame features of spatial CNNs. Several works have jointly encoded frame features and learned changes in appearance over time with Convolutional LSTMs \pcite{donahue2015long,srivastava2015unsupervised}. Similarly, for multi-actor action recognition, \tcite{wang2017recurrent} used three individual pathways with LSTMs for person action, group action, and scene recognition.

\noindent
\textbf{Two-stream models}. An alternative group of approaches included a parallel motion-specific stream in spatial CNNs. Two-stream models \pcite{simonyan2014two} encode motion and appearance explicitly with respective optical flow and RGB streams over stacks of frames. Extensions \pcite{feichtenhofer2016convolutional} have fused flow and spatial streams at intermediate layers while other approaches used cross-stream connections \pcite{feichtenhofer2017spatiotemporal}, multiple appearance streams \pcite{tu2018multistream}, recurrent layers \pcite{singh2016multi}, or concatenated appearance and motion volumes \pcite{jain2015modeep,wang2017spatiotemporal} to share information between the streams. \tcite{wang2016temporal} used a step-based approach that segmented videos into individual snippets, processed them in parallel, and fused class scores from each snippet. Improvements in inference speeds of two-stream models have been achieved with the addition of motion vectors \pcite{zhang2016real} or key volume mining \pcite{zhu2016key}. Although such approaches have established a new research direction in modeling videos, the representation of motion with precomputed motion features limits the capabilities of learned backbones \pcite{sevilla2019integration}.

\subsection{Jointly encoding space and time}
\label{sec:modeling::joint}

\looseness-1 Time and appearance can also be encoded jointly. 

\noindent
\textbf{Part-based representations}. SpatioTemporal Interest Points (STIPs) \pcite{laptev2003space} extended spatial interest point detection methods \pcite{forstner1987fast,harris1988combined} to the video domain. \tcite {liu2008learning,oikonomopoulos2005spatiotemporal} explored salient points based on peaks of activity variation. STIP features have been quantized in histograms of codewords \pcite{schuldt2004recognizing}. Several approaches have studied action-relevant temporal locations across viewpoints \pcite{yilmaz2006matching} and view-invariant trajectories \pcite{sheikh2005exploring}. \tcite{dollar2005behavior} proposed modeling periodic motions using sparse distributions of points of interest. This feature extractor prompted subsequent works \pcite{niebles2008unsupervised} with actions classified through a codebook of features.

\noindent
\textbf{Holistic stochastic representations}. Actions have also been modeled based on global information. \tcite{efros2003recognizing} created representations for different body parts and regressed towards representations of pre-classified actions. Subsequent works have explored action descriptors focused on object shapes \pcite{gorelick2006shape,jia2008human}, movements \pcite{sun2009action}, and spatiotemporal salient regions \pcite{wong2007extracting}. They have also extended existing approaches to multiple features and temporal scales \pcite{amer2012sum,liu2008recognizing,zelnik2001event,yang2020temporal}. Later works \pcite{blank2005actions} adapted and generalized holistic descriptors \pcite{gorelick2006shape} by concatenating 2D silhouettes to form space-time shapes corresponding to action performances. \tcite{sadanand2012action} similarly proposed a bank of volumetrically pooled features containing high-level representations of the actions. 

\noindent
\textbf{3D CNNs}. Orthogonal to hand-crafted features, 2D convolutions have been extended in various ways to 3D spatiotemporal kernels to jointly encode space and time \pcite{baccouche2011sequential,ji20123d,taylor2010convolutional,tran2015learning}. Subsequent works have demonstrated the potential of adapting image models to video \pcite{hara2018can}, explored video-specific architectures with spatiotemporal volumes across channels \pcite{chen2018multi}, and tiled 3D kernels \pcite{hegde2018morph}. They have also used channel-separated convolutions \pcite{jiang2019stm,luo2019grouped,tran2019video}, temporal residual connections \pcite{qiu2017learning}, global feature fusion \pcite{qiu2019learning}, resolution reduction \pcite{chen2019drop,stergiou2021multi}, and related appearance to spatiotemporal embeddings \pcite{wang2018appearance,zhou2018mict}. \tcite{carreira2017quo} integrated 3D convolutions into two-stream models for motion-implicit appearance representations in the RGB stream and motion-explicit representations in the optical flow stream. Several works have focused on improving the efficiency of action recognition architectures \pcite{feichtenhofer2020x3d,kondratyuk2021movinets,liu2022convnet}. They have used visual context from the scenes of actions, by either scene-type objectives \pcite{choi2019can}, decoupling scene and motion features \pcite{wang2021enhancing}, multi-domain information concatenation \pcite{kapidis2023multi}, or by fusing motion and scene information \pcite{stergiou2021learn}. To better extract temporal information, \tcite{feichtenhofer2019slowfast} proposed a dual pathway video model with a slow pathway operating over low frame rates for spatial semantics and a fast pathway with a high frame rate for motion. Similarly, \tcite{wang2020self} included a contrastive objective for learning the pace in videos. \tcite{xu2019self} explored temporal reasoning by including clip order prediction as an additional task to improve action recognition. The extension of 3D CNNs to longer sequences by segmenting videos with multiple temporal patches has also been attempted \pcite{ji2020action,hussein2019timeception,varol2017long}.

\noindent
\textbf{Spatiotemporal attention}. Attention is an effective approach for learning feature correspondences over space and time. \tcite{sharma2015action} used visual attention to localize action regions from CNN features with recurrent layers. \tcite{du2017recurrent} attended over spatial features across multiple frames based on their relevance to the action. Similarly, \tcite{chen20182} aggregated and propagated global information by attending over convolution features. \tcite{wang2018non} introduced non-local operations with bi-directional attention blocks over convolutions. Another early application of attention \pcite{girdhar2019video} was based on region proposals and the creation of feature banks \pcite{wu2019long} in longer videos. The introduction of Vision Transformers (ViTs) \pcite{dosovitskiy2020image} that encode visual information through region-based tokenization led to video-based adaptations that explored different spatiotemporal attention configurations \pcite{arnab2021vivit,bertasius2021space}. Others have explored token selection \pcite{bulat2021space,ryoo2021tokenlearner,zha2021shifted}, and the inclusion of contextual information \pcite{kim2021relational}. \tcite{liu2022video} introduced shifted non-overlapping attention windows to share information across patches. \tcite{hu2024tcnet} additionally used attention over motion-aligned input volumes. Feature hierarchies and latent resolution reductions have led to more compute- \pcite{fan2021multiscale,li2022mvitv2} and memory-efficient \pcite{wu2022memvit} architectures. Recent models such as MViT \pcite{yan2022multiview}, Hiera \pcite{ryali2023hiera}, UniFormer \pcite{li2022uniformer}, and MooG \pcite{van2024moving}, improved both the performance and capacity of video models. SSL has also shown great promise with pretext tasks based on contrastive learning \pcite{chen2020simple} or token masking \pcite{he2022masked}. \tcite{xing2023svformer} increased the complexity of the contrastive objective with pseudo labels and token mixing from different inputs. Masked autoencoders have also been extended to video data \pcite{feichtenhofer2022masked,wei2022masked}. Subsequent works have explored adaptive token masking \pcite{bandara2023adamae}, double masking on both the encoder and decoder \pcite{wang2023videomae}, token fusion \pcite{kim2024token}, and teacher-student masked autoencoders \pcite{wang2023masked}.

\noindent
\textbf{Video-language models}. Recently, language semantics from Large Language Models (LLMs) \pcite{brown2020language,touvron2023llama} have been used as a supervisory signal for vision tasks \pcite{li2023blip,liu2024visual,radford2021learning}. Initial efforts \pcite{zellers2021merlot} matched frame-level encodings to corresponding LLM embeddings of captions. Other approaches have optimized image-based encodings over frames by pooling spatial tokens \pcite{yu2022coca}, including cross-modal skip connections \pcite{xu2023mplug}, cross-attending modalities \pcite{alayrac2022flamingo}, and jointly attending visual and text embeddings \pcite{maaz2023video}. As static features provide only an appearance-based view, works have also used spatiotemporal Vision-Language models (VLMs) \pcite{piergiovanni2024mirasol3b} and extended the training objective \pcite{lu2024improving,zhao2024videoprism} to a two-step SSL pre-training with video-to-text alignment and video masking.
\begin{table}
    \centering
    \caption{\textbf{Action understanding datasets}. Works are grouped by year of release (Y). The number of classes, video instances, and actors are denoted with \#Cls, Inst, and Act. The average duration per annotation is denoted as AD. Short descriptions per dataset appear in the Context column.}
    \resizebox{\linewidth}{!}{
    \setlength\tabcolsep{1.0pt}
    \begin{tabular}{l l l l l l}
    \toprule
      Y & Dataset & \#Cls/Inst/Act/AD & Context  \\
      \midrule
      \multirow{4}{*}{\rotatebox{90}{2004-2007}} & KTH \citep{schuldt2004recognizing} & 6/2K/25/2.5s & \makecell[l]{Grayscaled videos of motions} \\
      & Weizmann \citep{gorelick2007actions} & 10/90/8/12s & \makecell[l]{Low-res. atomic motions} \\
      & Coffee\&Cigarettes \citep{laptev2007retrieving} & 2/245/5/5s & \makecell[l]{Smoking/drinking in movies} \\
      & CASIA Action \citep{wang2007human} & 15/1446/24/NA  &  \makecell[l]{Outdoor activities} \\
      \midrule
      \multirow{20}{*}{\rotatebox{90}{2008-2014}} & UCF Sports \citep{rodriguez2008action} & 9/150/$<$100/5s & \makecell[l]{Sports videos} \\
      & Hollywood \citep{laptev2008learning} & 8/475/$<$100/16s & \makecell[l]{Actions in movies} \\
      & UT-interaction \citep{ryoo2009spatio} & 6/90/60/17s & \makecell[l]{Dyadic human interactions} \\
      & CMU-MMAC \citep{deguide2008guide} & 5/182/43/7m & Multi-view recipe preparations \\
      & UCF-11 \citep{liu2009recognizing} & 11/1K/100+/5s & \makecell[l]{Actions in YouTube videos} \\
      & Hollywood2 \citep{marszalek2009actions} & 12/3K/100+/12s & \makecell[l]{Actions from movies} \\
      & TV-HI \citep{patron2010high} & 4/300/100+/3s & \makecell[l]{Interactions in TV shows} \\
      & UCF-50 \citep{reddy2013recognizing} & 50/5K/100+/15s & \makecell[l]{Web-sourced videos} \\
      & Olympic Sports \citep{niebles2010modeling} & 16/800/100+/3s & \makecell[l]{Actions in sports} \\
      & HMDB-51 \citep{kuehne2011hmdb} & 51/7K/100+/3s & \makecell[l]{Actions from movies} \\
      & CCV \citep{jiang2011consumer} & 20/9K/100+/80s & \makecell[l]{Web-sourced videos} \\
      & UCF-101 \citep{soomro2012ucf101} & 101/13K/100+/15s & \makecell[l]{Action with hierarchies} \\
      & CAD-60 \citep{sung2012unstructured} & 12/60/$<$30/45s & \makecell[l]{Atomic actions in RGB-D}  \\
      & MPII \citep{rohrbach2012database} & 65/5.6K/100+/11m & \makecell[l]{Web-source actions}  \\ 
      & ADL \citep{pirsiavash2012detecting} & 32/436/20/1.3s & Videos of daily activities \\
      & 50 Salads \citep{stein2013combining} & 17/899/25/37s & \makecell[l]{Salad making videos} \\ 
      & J-HMDB \citep{jhuang2013towards} & 21/928/100+/1.2s & Videos with joints positions \\
      & CAD-120 \citep{koppula2013learning} & 12/120/$<$60/45s & \makecell[l]{Extension of CAD-60} \\
      & Penn Action \citep{zhang2013actemes} & 15/2.3K/100+/2s & Web-sourced atomic actions \\
      & Sports-1M \citep{karpathy2014large} & 487/1M/1,000+/9s & \makecell[l]{Sports actions/activities} \\
      \midrule
      \multirow{23}{*}{\rotatebox{90}{2015-2018}} & EGTEA Gaze+ \citep{li2015delving} & 106/15K/32/28s & \makecell[l]{Egocentric actions w/ gaze} \\
      & ActivityNet-100 \citep{caba2015activitynet} & 100/5K/100+/2m & \makecell[l]{Untrimmed web videos} \\ 
      & Watch-n-Patch \citep{wu2015watch} & 21/2K/7/30s. & \makecell[l]{Daily activities in RGB-D} \\
       & NTU-RGB-60 \citep{shahroudy2016ntu} & 60/57K/40/2s. & Multi-sensory actions \\
      & ActivityNet-200 \citep{caba2015activitynet} & 200/15K/100+/2m & \makecell[l]{ActivityNet-100 extension} \\
      & YouTube-8M \citep{abu2016youtube} & NA/8M/NA/NA & \makecell[l]{Multi-labelled videos} \\
      & Charades \citep{sigurdsson2016hollywood} & 157/67K/267/30s & \makecell[l]{Daily activities videos} \\
      & ShakeFive2 \citep{van2016spatio} & 5/153/33/7s & \makecell[l]{Interactions with pose data} \\
      & DALY \citep{weinzaepfel2016towards} & 10/510/100+/4m & Untrimmed YouTube videos \\
      & OA \citep{li2016recognition} & 48/480/$<$100/5s & \makecell[l]{Ongoing actions} \\
      & CONVERSE \citep{edwards2016pose} & 10/NA/NA/NA & \makecell[l]{Human interactions} \\
      & TV-Series \citep{de2016online} & 30/6,2K/100+/2s & \makecell[l]{Actions from TV series} \\
      & Volleyball \citep{ibrahim2016hierarchical} & 6/1.4K/$<$100/$<$1s & Group actions in volleyball  \\
      & MSR-VTT \citep{xu2016msr} & 200K/7.1K/1,000+/20s & Video captions \\
      & Okutama Action \citep{barekatain2017okutama} & 12/4.7K/$\sim$400/60s & \makecell[l]{Aerial views of action} \\
      & K-400 \citep{kay2017kinetics} & 400/306K/1,000+/10s & \makecell[l]{Web-sourced short actions} \\
      & Smthng-Smthng v1 \citep{goyal2017something} & 174/109K/100+/4s & \makecell[l]{Human actions with objects} \\
      & MultiTHUMOS \citep{yeung2018every} & 65/39K/100+/3s & \makecell[l]{Densely labeled actions} \\
      & Diving-48 \citep{li2018resound} & 48/18K/NA/3s & \makecell[l]{Diving sequences} \\
      & EK-55 \citep{damen2018scaling} & 2,747/40K/35/3s & \makecell[l]{Egocentric actions in kitchens} \\
      & K-600 \citep{carreira2018short} & 600/495K/100+/10s & \makecell[l]{Extension of K-400} \\
      & VLOG \citep{fouhey2018lifestyle} & 30/122K/10.7K/10s & \makecell[l]{Actions in lifestyle VLOGs} \\
      & AVA \citep{gu2018ava} & 80/430/100+/15m & Localized atomic actions \\
      \midrule
      \multirow{24}{*}{\rotatebox{90}{2019-now}} & NTU-RGB-120 \citep{shahroudy2016ntu} & 120/114K/106/2s. & Multi-sensory actions \\ 
      & Charades-Ego \citep{sigurdsson2018charades} & 156/7.8K/100+/9s & Daily indoor activities \\
      & Smthng-Smthng v2 \citep{goyal2017something} & 174/221K/100+/4s & \makecell[l]{Human actions with objects} \\
      & K-700 \citep{carreira2019short} & 700/650K/1,000+/10s & \makecell[l]{Extension of K-600} \\
      & Moments in Time \citep{monfort2019moments} & 339/1M/1,000+/3s & \makecell[l]{Short dynamic scenes} \\
      & HACS (Clips) \citep{zhao2019hacs} & 200/1.5M/1,000+/2s & \makecell[l]{Action over fixed durations} \\
      & IG65M \citep{ghadiyaram2019large} & NA/65M/NA/NA & \makecell[l]{Actions in Instagram videos} \\
      & Toyota Smarthome \citep{dai2022toyota} & 31/16K/18/21m & Senior home activities \\
      & AViD \citep{piergiovanni2020avid} & 887/450K/1,000+/9s & \makecell[l]{Anonymized videos} \\
      & HVU \citep{diba2020large} & 3K/572K/1,000+/10s & \makecell[l]{Hierarchy of semantics} \\
      & Action-Genome \citep{ji2020action} & 453/10K/100+/1s & Daily home activities \\
      & K-700 (2020) \citep{smaira2020short} & 700/647K/1,000+/10s & \makecell[l]{Update of K-700}  \\
      & FineGym \citep{shao2020finegym} & 530/33K/100+/10m & \makecell[l]{Gymnastics videos}  \\
      & RareAct \citep{miech2020rareact} & 122/7.6K/100+/10s & Unusual actions \\
      & HAA500 \citep{chung2021haa500} & 500/10K/1,000+/2s & \makecell[l]{Atomic actions}  \\
      & MultSports \citep{li2021multisports} & 4/3.2K/100+/21s & Localized sports actions \\
      & MOMA \citep{luo2021moma} & 136/12K/100+/10s & Hierarchical actions \\
      & WebVid-2M \citep{bain2021frozen} & NA/2M/1,000+/4s & Video-image pairs  \\
      & HOMAGE \citep{rai2021home} & 453/5.7K/40/2s & Extension of \citep{ji2020action} \\
      & EK-100 \citep{damen2022rescaling} & 4,053/90K/37/3s & \makecell[l]{Egocentric actions}  \\
      & FineAction \citep{liu2022fineaction} & 106/103K/1,000+/7s & Hierarchies for TAL  \\
      & EGO4D \citep{grauman2022ego4d} & 1000+/9.6K/931/48s & Diverse egocentric videos  \\
      & Assembly-101 \citep{sener2022assembly101} & 1.3K/4.3K/53/2s & Procedural activities \\
      & Ego-Exo-4D \citep{grauman2024ego} & 689/5,035/740/5m & Multimodal multi-view videos \\
      
      \end{tabular}
    }
    \label{tab:action_recognition_datasets}
    \vspace{-1em}
\end{table}

\section{Video datasets comprising human actions}
\label{sec:datasets}

Significant efforts have been made to collect video datasets for various action understanding tasks. We discuss the main challenges associated with dataset collection in \Cref{sec:datasets::challenges}. We then explore two broad dataset types based on target tasks and use cases. The first set includes general-purpose datasets for pre-training and model evaluation. The second set of datasets has been collected to evaluate models on specific modalities or domains. The sets are discussed in \Cref{sec:datasets::general} and \Cref{sec:datasets::specific}, respectively.

\begin{figure*}[t]
    \centering
    \includegraphics[width=\linewidth]{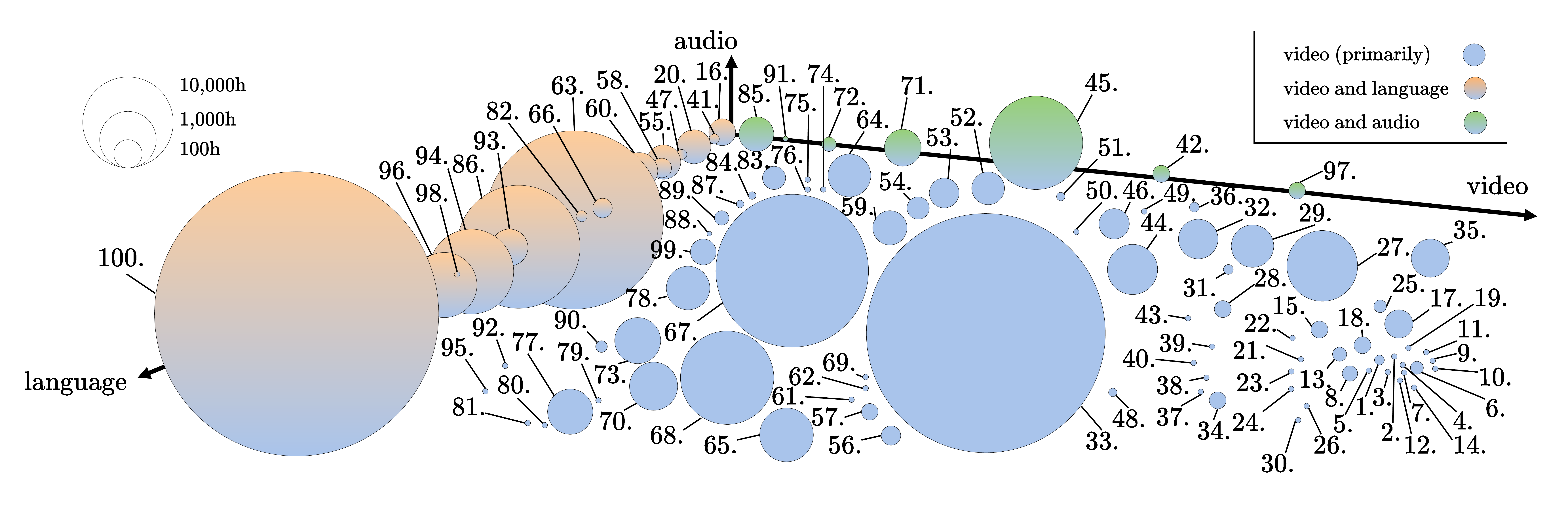}
    \resizebox{\linewidth}{!}{
    \begin{tabular}{llll}
      1. KTH~\citep{schuldt2004recognizing} &  
      2. Weizmann~\citep{gorelick2007actions} &  
      3. Coffee \& Cigarettes~\citep{laptev2007retrieving} & 
      4. CASIA Action~\citep{wang2007human} \\ 
      5. UCF Sports~\citep{rodriguez2008action} & 
      6. Hollywood~\citep{laptev2008learning} & 
      7. UT-Interaction~\citep{ryoo2009spatio} & 
      8. CMU-MMAC~\citep{deguide2008guide} \\
      9. UCF-11~\citep{liu2009recognizing} & 
      10. Hollywood2~\citep{marszalek2009actions} & 
      11. TV-HI~\citep{patron2010high} &
      12. Humaneva~\citet{sigal2010humaneva} \\
      13. UCF-50~\citep{reddy2013recognizing} & 
      14. Olympic Sports~\citep{niebles2010modeling} &
      15. HMDB-51~\citep{kuehne2011hmdb} &
      16. MSVD~\citep{chen2011collecting} \\
      17. CCV~\citep{jiang2011consumer} & 
      18. UCF-101~\citep{soomro2012ucf101} &
      19. CAD-60~\citep{sung2012unstructured} &
      20. MPII~\citep{rohrbach2012database} \\ 
      21. ADL~\citep{pirsiavash2012detecting} &
      22. 50 Salads~\citep{stein2013combining} &
      23. AVENUE~\citep{lu2013abnormal} &
      24. J-HMDB~\citep{jhuang2013towards} \\
      25. CAD-120~\citep{koppula2013learning} &
      26. Penn Action~\citep{zhang2013actemes} &
      27. Sports-1M~\citep{karpathy2014large} & 
      28. EGTEA Gaze+~\citep{li2015delving} \\
      29. ActivityNet-100~\citep{caba2015activitynet} &
      30. Watch-n-Patch~\citep{wu2015watch} &
      31. NTU-RGB-60~\citep{shahroudy2016ntu} &
      32. ActivityNet-200~\citep{caba2015activitynet} \\
      33. YouTube-8M~\citep{abu2016youtube} &
      34. Charades~\citep{sigurdsson2016hollywood} &
      35. ShakeFive2~\citep{van2016spatio} &
      36. DALY~\citep{weinzaepfel2016towards} \\
      37. OA~\citep{li2016recognition} &
      38. CONVERSE~\citep{edwards2016pose} &
      39. TV-servies~\citep{de2016online} & 
      40. Volleyball~\citep{ibrahim2016hierarchical} \\
      41. MSR-VTT~\citep{xu2016msr} &
      42. Greatest Hits~\citep{owens2016visually} &
      43. Okutama Action~\citep{barekatain2017okutama} &
      44. K-400~\citep{kay2017kinetics} \\
      45. AudioSet~\citep{gemmeke2017audio} &
      46. Smthng-Smthng (v1/v2)~\citep{goyal2017something} &
      47. TGIF~\citep{jang2017tgif} &
      48. CMU Panoptic~\citep{joo2017panoptic} \\
      49. MultiTHUMOS~\citep{yeung2018every} &  
      50. RESOUND~\citep{li2018resound} &
      51. EK-55~\citep{damen2018scaling} &
      52. K-600~\citep{carreira2018short} \\
      53. VLOG~\citep{fouhey2018lifestyle} &
      54. AVA~\citep{gu2018ava} &
      55. TVQA~\citep{lei2018tvqa} &
      56. UCF-Crime~\citep{sultani2018real} \\
      57. Charades-Ego~\citep{sigurdsson2018charades} &
      58. YouCook2~\citep{zhou2018towards} &
      59. K-700~\citep{carreira2019short} &
      60. COIN~\citep{tang2019coin} \\
      61. AIST~\citep{tsuchida2019aist} &
      62. Drive\&act~\citep{martin2019drive} &
      63. HowTo100m~\citep{miech2019howto100m}&
      64. Moments in Time~\citep{monfort2019moments} \\
      65. HACS~\citep{zhao2019hacs} &
      66. VATEX~\citep{wang2019vatex} &
      67. IG65M~\citep{ghadiyaram2019large} &
      68. Toyota Smarthome~\citep{dai2022toyota} \\
      69. OOPS~\citep{epstein2020oops} &
      70. AViD~\citep{piergiovanni2020avid} &
      71. VGGSound~\citep{chen2020vggsound} &
      72. LLP~\citep{tian2020unified} \\
      73. HVU~\citep{diba2020large} &
      74. DMP~\citet{ortega2020dmd} &
      75. Action Genome~\citep{ji2020action} &
      76. Countix~\citep{dwibedi2020counting} \\
      77. K-700 (2020)~\citep{smaira2020short} &
      78. FineGym~\citep{shao2020finegym} & 
      79. RareAct~\citep{miech2020rareact} &
      80. Immersive Light Field~\citet{broxton2020immersive} \\
      81. HAA500 \citep{chung2021haa500} &
      82. IKEA ASM~\citep{ben2021ikea} &
      83. MultiSports~\citep{li2021multisports} &
      84. MOMA~\citep{luo2021moma} \\
      85. Spoken Moments~\citep{monfort2021spoken} &
      86. WebVid-2M~\citep{bain2021frozen} &
      87. Home action genome~\citep{rai2021home} & 
      88. DanceTrack~\citep{sun2022dancetrack} \\
      89. EK-101~\citep{damen2022rescaling} &
      90. FineAction~\citep{liu2022fineaction} &
      91. AVSBench~\citep{zhou2022audio} &
      92. Neural 3D Video~\citet{li2022neural} \\
      93. Assembly101~\citep{sener2022assembly101} &
      94. Ego4D \citep{grauman2022ego4d} &
      95. SportsMOT~\citep{cui2023sportsmot} &
      96. Ego-Exo-4D \citep{grauman2024ego} \\ 
      97. EPIC-Sounds \citep{huh2023epic} &
      98. MVBench \citep{li2024mvbench} &
      99. OVR \citep{dwibedi2024ovr} &
      100. Vidchapters-7M \citep{yang2024vidchapters}
    \end{tabular}
    }
    \caption{\textbf{Datasets compared by total dataset duration and primary modality}. Circle sizes correspond to the (approximate) summed duration of all videos in the datasets. Recent datasets (i.e., $>80$) have longer total running times and include additional modalities such as language or audio.}
    \label{fig:dataset_blobs}
\end{figure*}

\subsection{Data collection challenges}
\label{sec:datasets::challenges}

\noindent
\textbf{Ensuring sufficient diversity} of people, scenarios, and activities is important when amassing video datasets at scale. Although in recent years scalability has been achieved by transiting from locally-sourced datasets \pcite{caba2015activitynet,laptev2008learning,shahroudy2016ntu,schuldt2004recognizing,soomro2012ucf101} to multi-team-multi-year efforts \pcite{grauman2022ego4d,li2024mvbench,yang2024vidchapters} dataset diversity remains subjective which can result to semantically overlapping \pcite{khamis2015walking,wray2019learning} or ambiguous \pcite{kim2022action,sigurdsson2017actions} labels. As video understanding is a multifaceted topic relating to vision, robotics, and augmented reality, collecting \textbf{meaningful scenarios and annotations} also presents difficulties. Several datasets have used surveys \pcite{caba2015activitynet,grauman2022ego4d,grauman2024ego,lin2023videoxum} or relied on meta-data from online videos \pcite{chen2023vast,miech2019howto100m,smaira2020short,yang2024vidchapters} as guides for collection. Despite such efforts, sourcing is still bound to preliminary data assumptions \pcite{rahaman2022generalized}, inconsistencies across annotations \pcite{moltisanti2017trespassing}, and difficulties in task definitions \pcite{alwassel2018diagnosing}. The collection pipelines also require scalability. In recency, \textbf{greater automation} in video collection has been achieved with the use of embeddings from vision encoders \pcite{chen2020vggsound,huang2024vbench,zhu2024video}, and LLM-generated descriptions \pcite{fu2024video,li2024videovista,mangalam2023egoschema}.

\subsection{General datasets}
\label{sec:datasets::general}

The past two decades have seen a significant increase in dataset size, leading to more robust baselines. We present widely-adopted benchmarks chronologically in \Cref{tab:action_recognition_datasets}. The primary focus of initial benchmarks \pcite{schuldt2004recognizing,gorelick2007actions} has been the categorization of simple actions such as \emph{walking} and \emph{hand waving}. Subsequent datasets predominantly comprised videos from either TV shows/movies \pcite{laptev2007retrieving,laptev2008learning,marszalek2009actions,patron2010high,kuehne2011hmdb} or sports footage \pcite{rodriguez2008action,liu2009recognizing,reddy2013recognizing,niebles2010modeling}. Important steps towards establishing large-scale datasets for the video domain were made with the introduction of Sports-1M \pcite{karpathy2014large}, YouTube-8M \pcite{abu2016youtube}, and Kinetics \pcite{carreira2017quo} that include web-sourced videos of a diverse range of actions. Evident from their sizes shown in \Cref{fig:dataset_blobs}, these datasets paved the way as general benchmarks for models that can subsequently be adapted to smaller, more niche datasets such as UCF-101 \pcite{soomro2012ucf101} and ActivityNet \pcite{caba2015activitynet}. Despite their size and use in multiple downstream tasks, there is still room to address specific action understanding tasks or modalities supplementary to vision. Domains such as 
egocentric vision, human-object interaction recognition, and hierarchical action understanding have gained popularity, prompting the creation of domain-specific datasets. EGTEA Gaze+ \pcite{li2015delving}, EPIC KITCHENS \pcite{damen2022rescaling}, and later EGO4D \pcite{grauman2022ego4d} have been the main benchmarks for egocentric vision. Something-Something \pcite{goyal2017something} and Charades \pcite{sigurdsson2016hollywood} have been predominantly used as benchmarks for object-based actions with a greater focus on temporal information. Datasets such as Diving-48 \pcite{li2018resound} and FineGym \pcite{shao2020finegym} incorporate semantic hierarchies in their annotations. More recent datasets have focused on tasks related to action recognition including instruction learning \pcite{alayrac2016unsupervised,bansal2022my,ben2021ikea,liui2024kea,ohkawa2023assemblyhands,sener2022assembly101,tang2019coin}, action phase alignment \pcite{sermanet2017unsupervised}, repeating action counting \pcite{dwibedi2020counting,dwibedi2024ovr,hu2022transrac,runia2018real,zhang2020context}, action completion prediction \pcite{epstein2020oops}, driver behavior recognition \pcite{martin2019drive,ortega2020dmd}, anomaly detection \pcite{acsintoae2022ubnormal,liu2018future,lu2013abnormal,sultani2018real,wu2020not}, hand-object interactions \pcite{chao2021dexycb,garcia2018first,hampali2020honnotate,kwon2021h2o,moon2020interhand2,mueller2017real}, and object state change detection in actions \pcite{souvcek2022look}.

\subsection{Domain- and modality-specific datasets}
\label{sec:datasets::specific}

Apart from general-purpose datasets, several benchmarks have been designed to evaluate model capabilities of specific aspects of action understanding. We overview of benchmarks in three groups: based on the holistic understanding of scenes from multiple viewpoints, and with supplementary modalities such as language and audio.

\noindent
\textbf{Multi-view}. Initial efforts to compile multi-view videos included a small number of subjects \pcite{sigal2010humaneva} or synthetic data \pcite{ionescu2013human3}. High-quality multi-view videos depend highly on the hardware and setup \pcite{wang2023learning}. CMU panoptic \pcite{joo2017panoptic} captured group interactions within a dome with 480 cameras. Interactions included social settings, games, dancing, and musical performances. ZJU-Mocap \pcite{peng2021neural} comprised dynamic videos of human motions from 20 cameras. The Immersive Light Field dataset \pcite{broxton2020immersive} contains videos with 6 degrees of freedom from a camera rig consisting of 46 action cameras. Multi-view datasets are collected for a variety of target tasks, including dance sequence reconstruction \pcite{tsuchida2019aist}, and the dynamic synthesis of indoor spaces in which actions take place \pcite{dai2022toyota,tschernezki2024epic}.

\noindent
\textbf{3D vision}. 3D human action reconstruction initially relied on synthetic data \pcite{anguelov2005scape,bronstein2010shrec} or lacked detailed ground-truth annotations \pcite{koppula2013learning,li2010action,shahroudy2016ntu,sung2012unstructured}. (D)FAUST \pcite{bogo2014faust,bogo2017dynamic} introduced a full pipeline for capturing high-resolution deformations. The collected 3D models of 300 meshes from static scans later increased to 40K meshes from dynamic scans. With the same multi-camera setting, DYNA \pcite{pons2015dyna} collected meshes by aligning 3D scans to template meshes using only geometric information. Other 3D data collection efforts have been specific to action synthesis in indoor \pcite{li2022neural} and outdoor \pcite{lin2021deep,yoon2020novel} settings, depth-based procedural learning \pcite{ben2021ikea,sener2022assembly101}, human-object interaction tracking \pcite{bhatnagar2022behave}, novel view synthesis \pcite{jiang2022neuman,peng2021neural}, language captioning for poses \pcite{delmas2022posescript}, or representing human-relevant attributes such as clothes and hair dynamics \pcite{black2023bedlam}. Large-scale collections that unify existing datasets also exist. \tcite{guo2020action2motion} created a dataset for 3D human motions by revamping \tcite{joo2017panoptic,shahroudy2016ntu,zou20203d}. Similarly,  \tcite{mahmood2019amass} combined a corpus of 15 archival marker-based mocap datasets while \tcite{lin2023motion} created a superset of datasets by sourcing videos \pcite{cai2022humman,chung2021haa500,taheri2020grab,tsuchida2019aist,zhang2022egobody} relevant to LLM motion text prompts. Recently, space-time-depth datasets and benchmarks have also been introduced for egocentric data for tracking human-object interactions \pcite{liu2022hoi4d,perrett2025hd}, mistake detection in procedural tasks \pcite{wang2023holoassist}, multi-task augmented reality \pcite{grauman2024ego,lv2024aria,pan2023aria}, and surface estimation and reconstruction \pcite{straub2024efm3d}.

\noindent
\textbf{Video-language}. In recent years, language has been integrated into vision methods as a natural extension to represent high-level semantics. Commonly, learning to map textual concepts and visual representations in a shared embedding space has been a widely adopted strategy by many video tasks \pcite{amrani2021noise,gabeur2020multi,liu2019use,miech2020end}. Initial video-language datasets \pcite{chen2011collecting,xu2016msr} were based on short video snippets and short textual descriptions of actions. More recent efforts also provide multilingual descriptions \pcite{wang2019vatex}. Video question-answering is a popular language-based task \pcite{jang2017tgif,lei2018tvqa,li2024mvbench,oncescu2021queryd,rawal2024cinepile,xiao2021next}. The order of instructions has been of great interest in longer videos since the introduction of HowTo100M \pcite{miech2019howto100m} and YouCook2 \pcite{zhou2018towards}. Benchmarks have also been proposed for other long-form tasks such as moment retrieval \pcite{rohrbach2015dataset,song2024moviechat,yang2024vidchapters}, frame extraction \pcite{li2024llava}, multimodal open-ended question answering \pcite{fu2024video,ying2024mmt}, and long-term reasoning \pcite{chandrasegaran2024hourvideo,fei2024video,mangalam2023egoschema}.

\noindent
\textbf{Audio and vision}. Human perception often relies on the inclusion of audio for understanding actions, especially in conditions where appearance may lead to ambiguous predictions. Audioset \pcite{gemmeke2017audio} is the largest audio-visual action dataset containing 2.1M clips across a long-tail distribution of 527 classes. VGG-Sound \pcite{chen2020vggsound} is another common benchmark with a uniform distribution of 200K videos over 300 classes. Datasets have also been collected for specific tasks such as audio-visual semantic segmentation \pcite{zhou2022audio}, audio-visual video parsing \pcite{tian2020unified}, material sound and action classification \pcite{huh2023epic,owens2016visually}, and video captioning \pcite{monfort2021spoken}.

The introduction of both general and task-specific datasets has improved the exploration of video tasks and standardized evaluation protocols. Main lines of research and challenges of these tasks are overviewed next.
\section{Recognizing observed actions}
\label{sec:recognition}

The recognition of actions in videos is a fundamental computer vision research theme. Relevant tasks focus on different aspects of the actions observed in full. We start by discussing approaches for optimizing model inputs in \Cref{sec:recognition::inputs}. We then overview popular temporal-based recognition tasks in \Cref{sec:recognition::temporal}. Tasks based on the semantic relationships between language and video are discussed in \Cref{sec:recognition::language}, whereas audio-visual and other multimodal approaches appear in \Cref{sec:recognition::audio}. 

\subsection{Video reduction methods}
\label{sec:recognition::inputs}

\begin{figure}[t]
     \centering
     \begin{subfigure}[b]{0.49\linewidth}
         \centering
         \includegraphics[width=\linewidth]{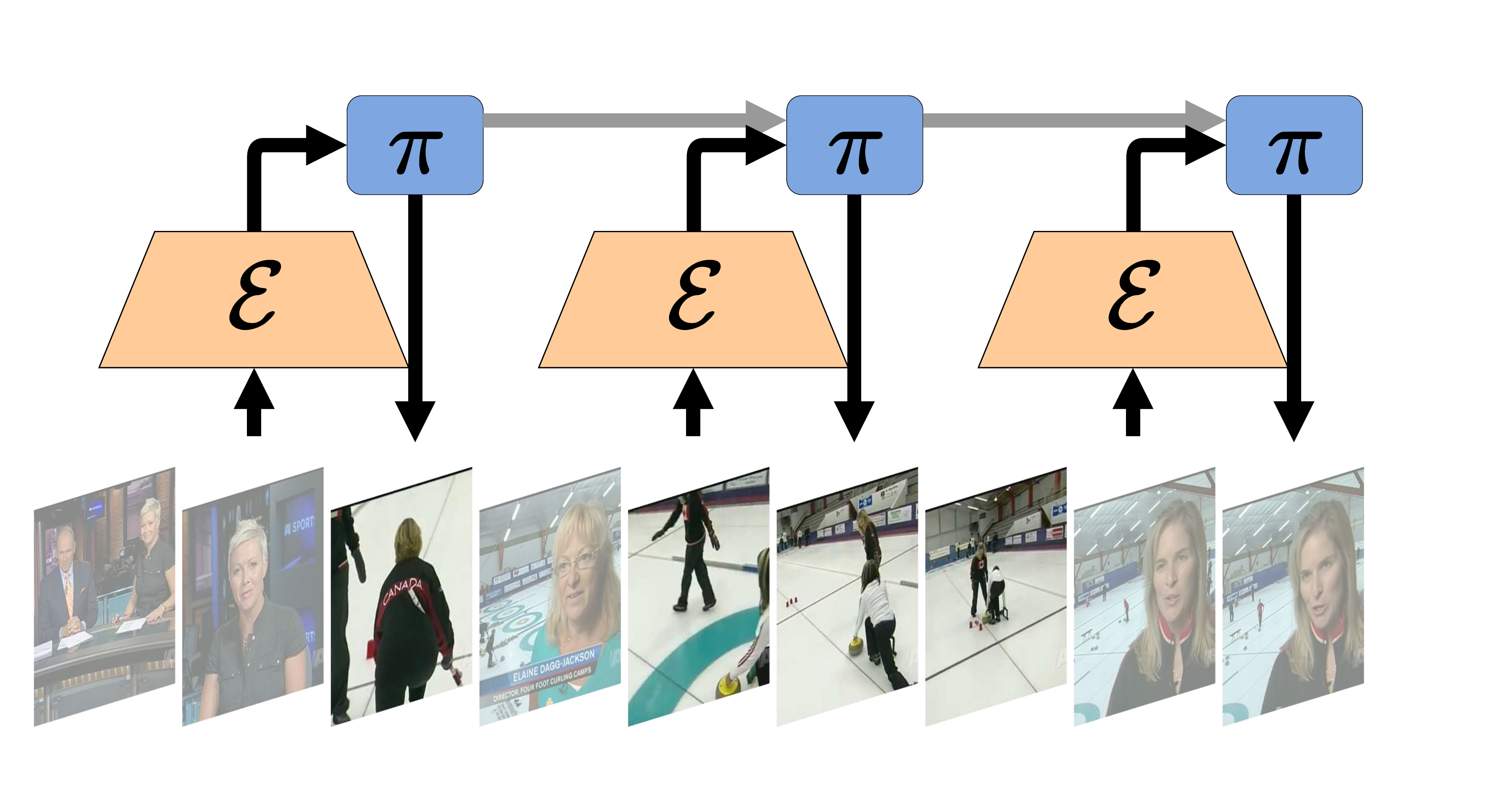}
         \caption{\textbf{Frame Sampling}}
         \label{fig:redundancies_reduction::sampling}
     \end{subfigure}
     \hfill
     \begin{subfigure}[b]{0.49\linewidth}
         \centering
         \includegraphics[width=\linewidth]{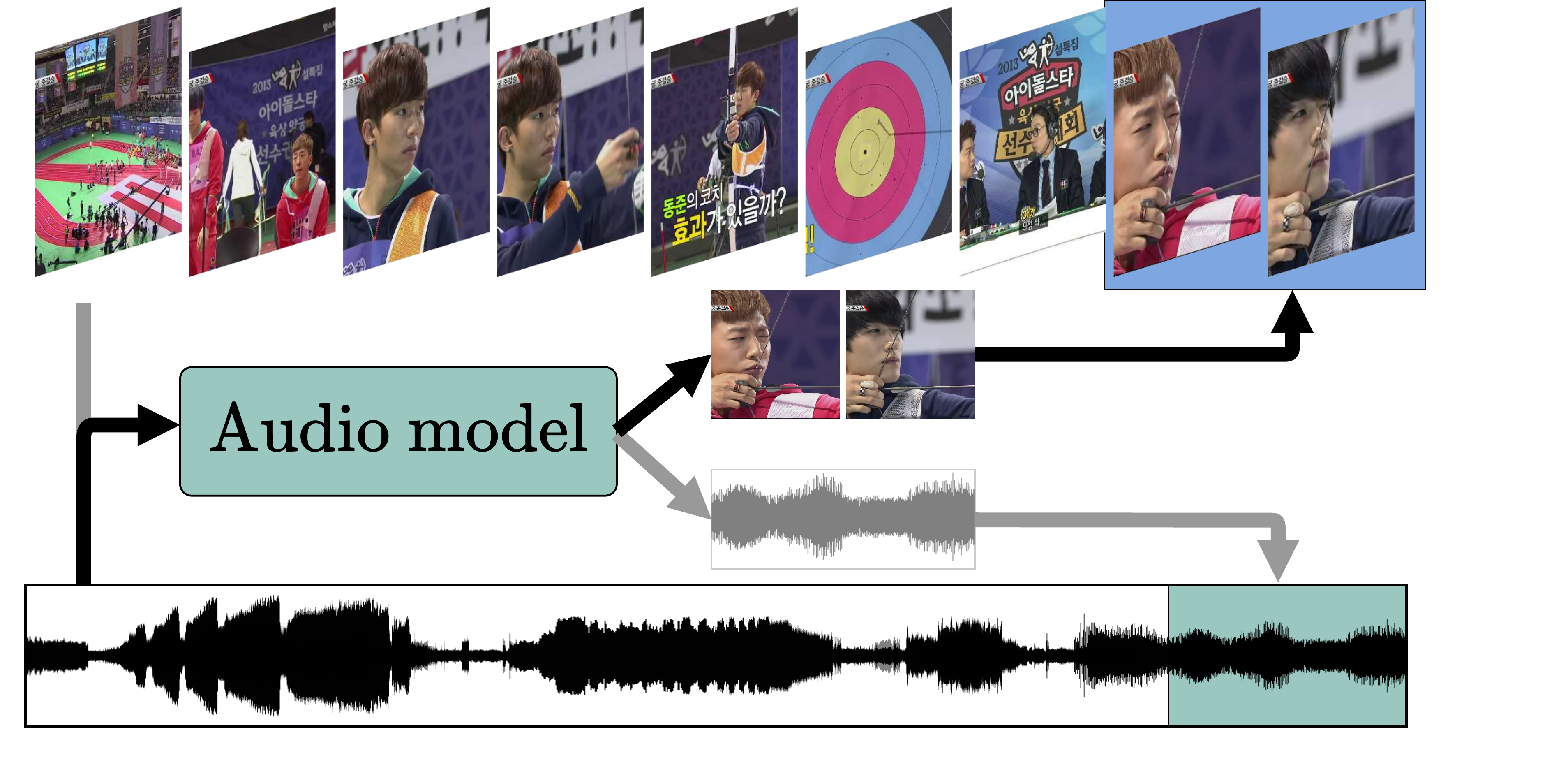}
         \caption{\textbf{Audio previewing}}
         \label{fig:redundancies_reduction::preview}
     \end{subfigure}
     \\
     \begin{subfigure}[b]{0.49\linewidth}
         \centering
         \includegraphics[width=\linewidth]{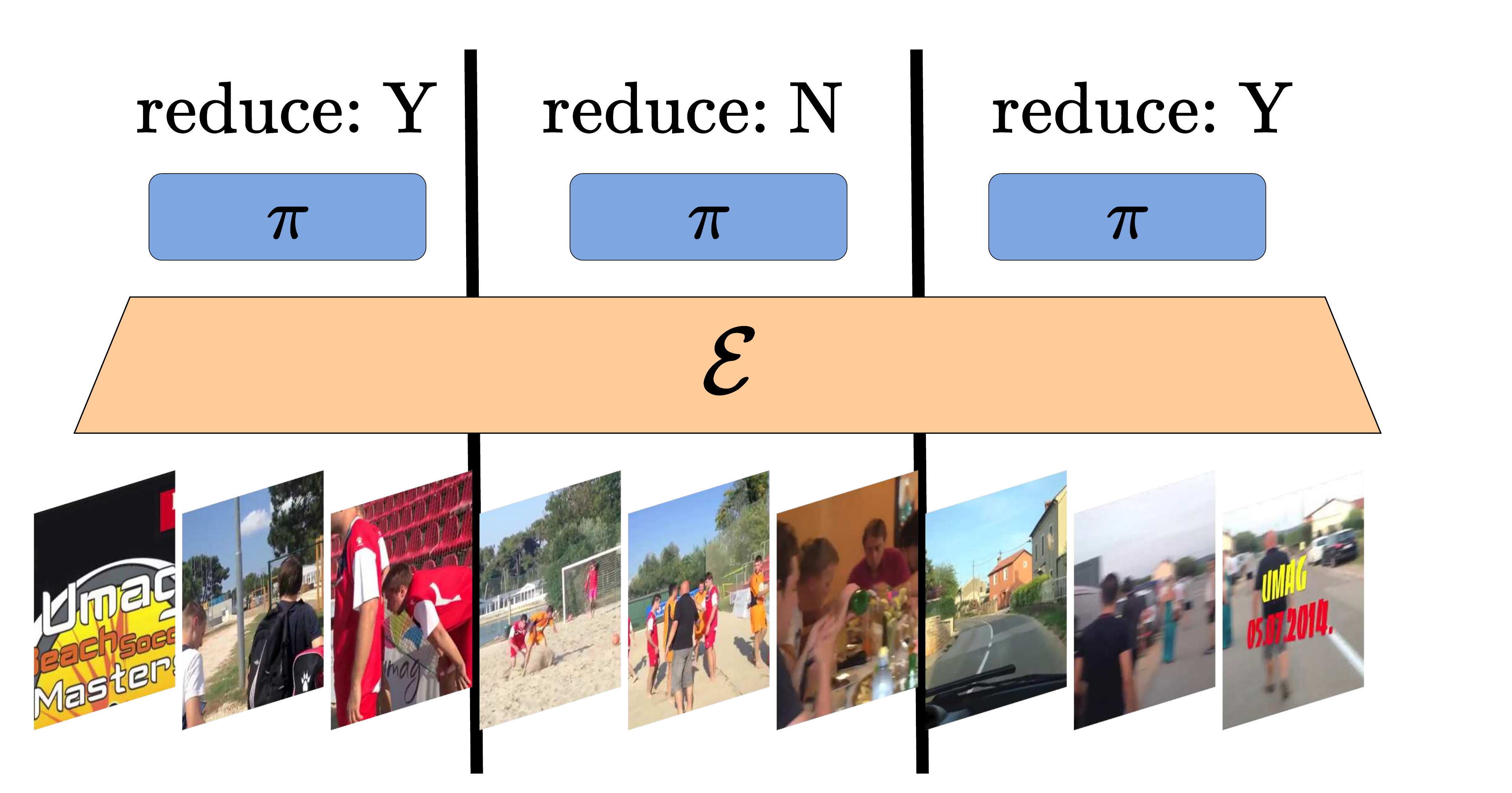}
         \caption{\textbf{Video input permuting}}
         \label{fig:redundancies_reduction::permute}
     \end{subfigure}
     \hfill
     \begin{subfigure}[b]{0.49\linewidth}
         \centering
         \includegraphics[width=\linewidth]{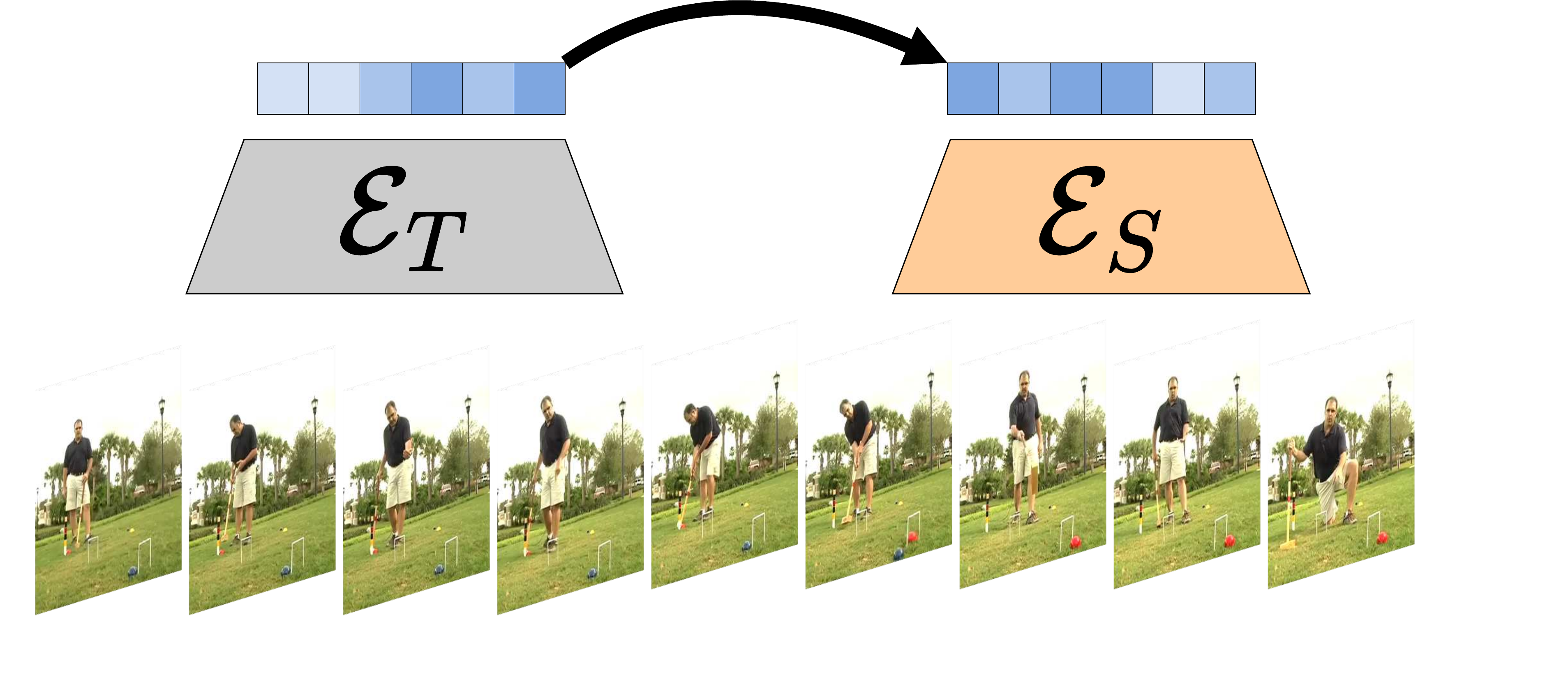}
         \caption{\textbf{Knowledge transfer}}
         \label{fig:redundancies_reduction::transfer}
     \end{subfigure}
        \caption{\textbf{Redundancy reduction methods} include (a) selection of task-specific salient frames, (b) use of supplementary modalities such as audio to preview relevant regions to sample from, (c) input permutations to compress irrelevant frames and segments, and (d) using embeddings from a teacher model as targets.}
        \label{fig:redundancies_reduction}
\end{figure}

Video inputs typically consist of tens to hundreds of highly visually similar frames. The uniform use of all frames can lead to an unsustainable computational burden. However, humans process stimuli selectively \pcite{eagleman2010does}. 
Several recognition approaches, as shown in \Cref{fig:redundancies_reduction}, aim to reduce compute and improve memory utilization by considering inputs selectively.

\subsubsection{Challenges}
\label{sec:recognition::challenges} 

Reducing frame-level redundancies in videos requires a high-level understanding of each temporal segment's relevance. The distinction and selection of relevant segments directly impact information loss. Long and complex scenes present significant challenges to video reduction methods. This \textit{uneven context inclusion} requires more efficient utilization of the model's capacity.

\subsubsection{Approaches}
\label{sec:recognition::approaches}

Among the most common approaches for reducing redundancies is \emph{frame sampling}. Works on frame sampling rely on policy networks that select frames based on the action's complexity \pcite{ghodrati2021frameexit,yeung2016end}, video context correspondence \pcite{wu2019adaframe}, or changes in the target class' probability \pcite{korbar2019scsampler}. \tcite{wang2021adaptive} used a recurrent network to localize action-relevant regions. Subsequent extensions targeted early stopping \pcite{wang2022adafocus} and related local and global features to determine action-relevant patches \pcite{wang2022adafocusv3}. \tcite{xia2022nsnet} used pseudo labels obtained by computing the embedding distance to class centroids to distinguish individual frames as salient and non-salient. Other approaches have used reward functions based on predictions from the selected frames \pcite{wu2020dynamic}, combined frame-level and video-level predictions \pcite{gowda2021smart}, optimized towards balancing accuracy and number of frames used \pcite{wu2019liteeval}, or removed tokens in transformer architectures \pcite{wu2024haltingvt}. 

A related set of approaches has extended unimodal frame sampling with \emph{audio previewing}. \tcite{gao2020listen} used both frame and audio features with a recurrent network to predict the next informative moment in the video. The video resolution used by the model was determined based on discovered informative parts in the audio stream. Similarly, \tcite{nugroho2023audio} used a saliency loss to localize the informative audio segments from which the corresponding video frames can be sampled. 

Although coarse frame sampling can be beneficial in short videos, selecting a limited number of frames in longer videos with a broader context can result in information loss. Another line of research thus studies redundancy reduction through \emph{video input permutations}. These works change frame resolutions based on classifier confidence \pcite{meng2020ar} or quantize frames at different precision \pcite{abati2023resq,sun2021dynamic}. \tcite{zhang2022look} used a two-branch approach for lightweight computations over large, less-relevant segments, and assigned more compute for segments with relevant context similar to \pcite{feichtenhofer2019slowfast}.  

Recent efforts have also used \emph{knowledge distillation} to improve the training efficiency of video model pipelines. \tcite{ma2022rethinking} reduced computations by learning to match student network features from videos with reduced resolution to the full-resolution features from a teacher network. \tcite{kim2021efficient} extended this approach by using cross-attention to learn the correspondence between teacher and student features. Distillation approaches have also used non-vision teacher models. \tcite{lei2021less} bound language embeddings to sparsely sampled clips from long videos while \tcite{xia2022temporal} used embeddings from textual event-object relations to discover salient frames. \tcite{tan2023egodistill} proposed a reconstruction approach for interpolating egocentric video features using embeddings from partial frames and the camera motion for the unobserved frames.

\subsubsection{Future outlooks}
\label{sec:recognition::outlooks}

Context-aware models can significantly improve both processing times and performance. Although most video reduction methods are primarily evaluated on classification or detection, more recent action understanding models have been optimized on multi-task and multi-domain objectives. This makes the discovery of relevant frames more difficult. For example, suppressing background frames is sub-optimal for episodic memory in which specific locations or attributes of objects not directly relevant to a current action need to be inferred. We expect that future redundancy reduction methods will focus more on preserving general scene information rather than ensuring that semantics of objects in the scene are not lost. This can be potentially achieved by discovering informative frames for diverse tasks or distilling scene context in low-memory representations, \eg language embeddings.

\begin{figure*}[t]
    \centering
    \begin{overpic}[width=\linewidth, trim={0 17cm 0 5cm},clip]{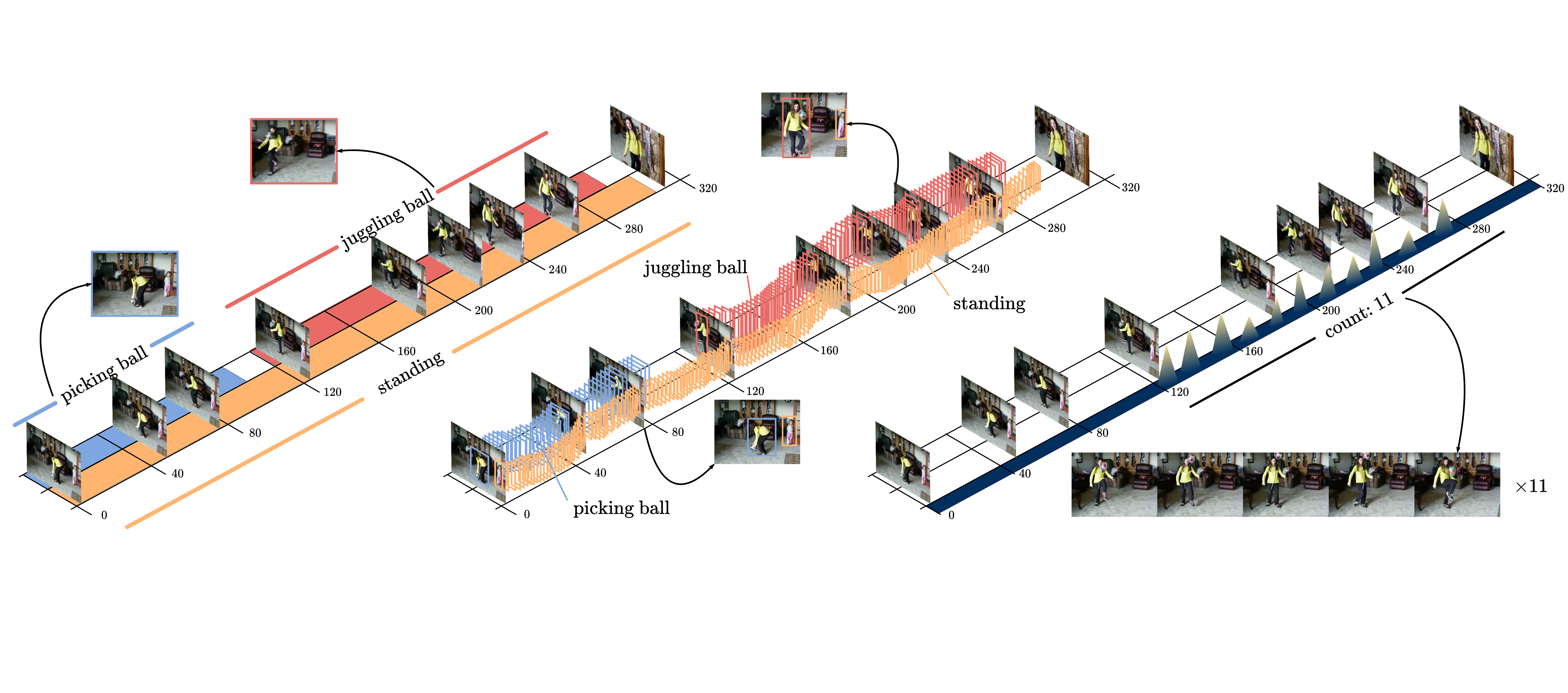}
    \put (15,0) {(a) TAL}
    \put (43,0) {(b) STAD}
    \put (71,0) {(c) VRC}
    
    \end{overpic}
    \caption{\textbf{Visualization of temporal-based tasks}. (a) Temporal Action Localization (TAL) discovers the start and end times of individual actions. In contrast, (b) Spatio-Temporal Action Detection (STAD) is more complex as it requires temporally and spatially localizing actions with bounding boxes for actors and objects over time. Distinctively, (c) Video Repetition Counting (VRC) is not based on action labels and instead requires counting repetitions of actions or motions in an open-set setting. Video source from \tcite{kay2017kinetics}.}
    \label{fig:loc_det_count}
\end{figure*}

\subsection{Temporal-based tasks}
\label{sec:recognition::temporal}

The perception of actions across time is a complex capability of human cognition. Understanding the timing of events is crucial for developing motor memory \pcite{eagleman2010does} for actions such as moving, speaking, and determining the causality of perceived temporal patterns.  
The importance of processing temporal information efficiently by computer vision systems has been shown through both standard performance metrics and semantic benchmarks \pcite{albanie2020end,stergiou2023leaping}. We provide a visualization of temporal-based tasks in \Cref{fig:loc_det_count} with the main challenges and task details discussed below.

\subsubsection{Challenges}
\label{sec:recognition::temporal:::challenges}

For most tasks, action categories are inferred directly without an explicit notion of their complexity based on \textbf{levels of abstraction}, \eg, atomic movements, composite motions, singular actions, or general activities. Although some datasets include action hierarchies \pcite{shao2020finegym,li2018resound}, these relationships are only used by a handful of existing works \pcite{mettes2020searching}. Different levels of abstraction typically have different temporal ranges, which require different approaches to process visual inputs. Moreover, there can be substantial temporal variations across class instances. This leads to larger temporal distances between discriminative information in videos, requiring proper extraction and modeling of long-range dependencies. We discuss solutions to cope with these variations in this section.

\subsubsection{Temporal localization} 
\label{sec:recognition::temporal:::localization}

A well-established video task is the discovery and classification of the actions performed alongside their temporal segments. Temporal Action Localization (TAL) aims to infer the action categories alongside the start and end times of the corresponding locations in untrimmed videos.

Early attempts have used Improved Dense Trajectories \pcite{wang2013action} and Fisher vectors \pcite{oneata2013action} to model the temporal dynamics of local points in scenes. \tcite{shou2016temporal} was one of the first to approach TAL with a joint action proposal and classification objective with a regional CNN. This joint optimization has been adapted with spatial- and temporal-only networks \pcite{lin2018bsn,paul2018w,wang2017untrimmednets}, regional proposal selection \pcite{chao2018rethinking,xu2017r}, and intra-proposal relationships with graph convolutions \pcite{zeng2019graph} in subsequent works. \tcite{shou2017cdc} predicted granularities at a frame level by transposing the temporal resolution of pretrained video encoders. More recent approaches for TAL can be categorized into three broad categories. 

\noindent
\textbf{One-stage}. Most similar to the aforementioned methods, one-stage approaches localize and classify actions in a single step by using hierarchical embeddings from feature pyramids \pcite{lin2021learning,liu2020progressive,shi2023tridet,zhang2022actionformer} or relating relevant video segments \pcite{shou2018autoloc,yang2020localizing}. Recently, \tcite{yan2023unloc} integrated scene semantic context for TAL with the inclusion of vision-language encoders. The start and end frames of an action can often be ambiguous. A set of approaches have relaxed deterministic start-end times objectives either by weighing the training loss with the importance of each frame \pcite{shao2023action} or by
learning distributions of possible start and end times \pcite{moltisanti2019action}. These approaches aimed to introduce a variance prior to the typical duration of the action. 

\noindent
\textbf{Two-stage}. Two-stage approaches disentangle the optimization into separate parts. \tcite{zhai2020two} combined proposals from spatial- and temporal-only streams into a fused final prediction. \tcite{chen2022dcan} used long- and short-range temporal information to refine the confidence of the generated action proposals. \tcite{huang2019decoupling} decoupled classification and localization to two objectives during training with two separate models that use cross-modal connections for exchanging information. Some works have focused on maximizing the embedding difference between representations of frames from action segments and non-relevant frames, either with positive and negative instances \pcite{luo2020weakly,zhang2021cola} or by a scoring function \pcite{rizve2023pivotal}. Methods have also improved the features of backbone encoders with TAL-relevant pretext tasks \pcite{zhang2022unsupervised}. Several two-stage methods process videos holistically \pcite{alwassel2021tsp,he2022asm,liu2021weakly,qing2021temporal}. \tcite{alwassel2021tsp} encoded local features over a sliding window using aggregated features from multiple local encoders. The model was optimized with a dual objective for discovering action regions and classifying every segment. Graphs have also been adopted for TAL with \tcite{bai2020boundary} using generated candidate proposals for the graph's start and end edges and their connected nodes. \tcite{zhao2021video} created a graph of hierarchical features over multiple temporal resolutions. More recent approaches \pcite{nag2023difftad} have formulated proposal prediction as a denoising task with noisy action proposals as input to a diffusion model conditioned on the video. Recent approaches have also used pretrained LLMs for TAL with specific Contrastive Language-Image pretraining (CLIP) query embeddings \pcite{ju2022prompting}, video representation masks adapted to the CLIP text encoder space \pcite{nag2022zero}, and CLIP text and visual features that are correlated to foreground masks learned from the video \pcite{phan2024zeetad}.

\noindent
\textbf{DEtection TRansformers (DETR)}. A recent set of methods is based on adapting image-based DETR \pcite{carion2020end} to TAL. These approaches rely on transformer encoder-decoders to create regional proposals optimized with bipartite matching. \tcite{tan2021relaxed} adapted DETR to video with a matching scheme of multiple positive action proposals to address sparsity in temporal annotations. Subsequent works have optimized the detection pipeline by either including dense residual connections \pcite{zhao2023re2tal} or caching short-term features \pcite{cheng2022tallformer,hong2022spotting}. 
\tcite{liu2024end} increased the model capacity by training intermediate adapters propagating information to the decoder from intermediate frozen encoder layers. Approaches have also explored training recipes with sparsely updating model layers \pcite{cheng2022stochastic}, vision-language pretraining distillation \pcite{ju2023distilling}, proposal hierarchies \pcite{wu2023newsnet},
and end-to-end TAL encoder-decoder optimization \pcite{liu2022empirical}. Works aiming to reduce training requirements have also been based on language embeddings with tuned visual-language projectors \pcite{liberatori2024test} and by start/end queries \pcite{aklilu2024zero}.

\subsubsection{Spatiotemporal detection} 
\label{sec:recognition::temporal:::detection}

SpatioTemporal Action Detection (STAD) is related to TAL but aims to jointly localize actions temporally and spatially detect action-relevant actors and objects. The main challenge of STAD methods is consistently linking detections and temporal action proposals across frames. Similar to TAL, two general directions can be used to overview relevant literature.

\noindent
\textbf{Two stages}. Building upon the advancements of image-based object detectors \pcite{girshick2014rich,girshick2015fast}, the majority of STAD approaches first detect objects and then temporally localize actions by tracking object candidates~\pcite{jain2014action,weinzaepfel2015learning}, ROI-pooling RGB and flow features \pcite{peng2016multi}, refining proposals iteratively \pcite{soomro2015action}, aligning source and target domain features~\pcite{agarwal2020unsupervised}, or using the general action level in the video as context \pcite{mettes2016spot}. \tcite{li2018recurrent} built upon prior two-stage detection works and incorporated recurrent proposals to include temporal context. Other approaches that focus on temporal information \tcite{singh2017online} used the arrow of time with different portions of the video detected at each step. Later benchmarks included longer videos to focus on activity-related tasks \pcite{gu2018ava}, enabling the greater exploration of context with feature banks \pcite{feng2021relation,pan2021actor,tang2020asynchronous,wang2018videos,wu2019long,wu2022memvit} and supplementary object information \pcite{arnab2021unified,hou2017tube,zhang2019structured}. Additional information such as keyframe saliency maps~\pcite{li2020actions,ulutan2020actor}, hands and poses~\pcite{faure2023holistic}, actor-object relations~\pcite{sun2018actor}, and SSL~\pcite{wang2023videomae} have also been explored. \tcite{alwassel2018diagnosing} analyzed the benefits of two-stage approaches and showed that they are primarily performant in handling temporal context. However, they also note that a significant limitation of two-stage approaches is that features are computed from backbones trained over auxiliary video tasks, potentially missing specific discriminative information.

\noindent
\textbf{Single-stage}. Drawing inspiration from single-stage object detection methods~\pcite{carion2020end,redmon2016you,liu2016ssd}, single-stage STAD approaches use an end-to-end trained unified framework for joint localization and detection~\pcite{chen2021watch,girdhar2019video,zhu2024dual}. \tcite{ntinou2024multiscale} extended the bipartite matching loss from \tcite{carion2020end} to spatio-temporal tokens. Other approaches used adaptive feature sampling~\pcite{wu2023stmixer}, conditionally modeled visual features based on motion~\pcite{zhao2019dance}, and contrasted different views~\pcite{kumar2022end}. Directly predicting tubelets has also been adopted by recent approaches \pcite{gritsenko2024end,kalogeiton2017action,song2019tacnet,yang2019step,zhao2022tuber}. \tcite{kalogeiton2017action} stacked embeddings from a backbone applied over a sliding window and regressed both classes and tubelets over the entire video. \tcite{zhao2022tuber} used an encoder-decoder to generate tubelet queries and cross-attended them to visual features. \tcite{gritsenko2024end} generated candidate tubelets from condensed query representations cross-attended by features from each frame. Beyond STAD, tubelets have also been used as a self-similarity pretraining objective~\pcite{thoker2023tubelet} to enforce correspondence of videos from different domains but with similar local motions. 

\subsubsection{Repetition counting}
\label{sec:recognition::temporal:::vrc}

Video Repetition Counting (VRC) aims to count the number of action repetitions. In contrast to TAL and STAD, VRC is an open-set task and does not require action categories.

\newcolumntype{g}{>{\columncolor{LightGrey}}l}

\begin{table*}[t]
    \centering
    \caption{\textbf{Self-Supervised Learning (SSL) methods and video task adaptations}. Three learning paradigms are used to group pretext tasks.
    }
    \resizebox{\linewidth}{!}{
    \setlength\tabcolsep{1.0pt}
    \begin{tabular}{c l l c l l }
    \toprule
      Learning & & Task & & Method & Video adaptations~$\textcolor{red}{^1}$  \\
      \midrule
      \multirow{5}{*}{Context-based} & \multirow{5}{*}{       
    \begin{tikzpicture}
    \draw    (0,.5) .. controls (.5,.5) and (.5,1.2) .. (1,1.2) ;
    \draw    (0,.5) .. controls (.5,.5) and (.5,-.2) .. (1,-.2) ;
    \draw    (0,.5) -- (1,.5) ;
    \end{tikzpicture}
      }&  Arrow of Time & & & \makecell[l]{
      \citet{benaim2020speednet},
      \citet{dwibedi2018temporal},
      \citet{destro2024cyclecl},
      \citet{donahue2024learning}, \\
      \citet{salehi2023time},
      \citet{wei2018learning},
      \citet{wu2021contrastive}
      }  \tstrut \bstrut \\
      & & \multicolumn{1}{g}{Jigsaw} & \multicolumn{1}{g}{} & \multicolumn{1}{g}{} & \multicolumn{1}{g}{\tabularCenterstack{l}{
      \citet{kim2019self},
      \citet{lee2017unsupervised},
      \citet{liu2024solving},
      \citet{misra2016shuffle},
      \citet{wang2022video}, \\
      \citet{xu2019self}
      }} \tstrut \bstrut \\
      & & Colorization & & & \makecell[l]{
      \citet{ali2023task},
      \citet{dhiman2023corf},
      \citet{jabri2020space},
      \citet{liu2024temporally},
      \citet{vondrick2018tracking},\\
      \citet{wu2020memory},
      \citet{zhang2023temporal}
      } \tstrut \bstrut \\
      \multirow{16}{*}{Contrastive} & \multirow{16}{*}{       
    \begin{tikzpicture}
    \draw    (0,0.5) .. controls (.5,0.5) and (.5,2.) .. (1,2.) ;
    \draw    (0,.5) .. controls (.5,0.5) and (.5,.3) .. (1.25,.3) ;
    \draw    (0,.5) .. controls (.5,.5) and (.5,-.8) .. (1,-.8) ;
    \draw    (0,.5) .. controls (.5,.5) and (.5,-1.75) .. (1,-1.75) ;
    \end{tikzpicture}
      }&\multirow{8}{*}{Negative Samples} & \multirow{8}{*}{       
    \begin{tikzpicture}
    \draw    (0,-.1) .. controls (.5,-.1) and (.5,1) .. (1,1) ;
    \draw    (0,-.1) .. controls (.5,-.1) and (.5,.35) .. (1,.35) ;
    \draw    (0,-.1) .. controls (.5,-.1) and (.5,-.3) .. (1,-.3) ;
    \draw    (0,-.1) .. controls (.5,-.1) and (.5,-1) .. (1,-1) ;
    \end{tikzpicture}
      } & \multicolumn{1}{g}{MoCo \citep{he2020momentum}} & \multicolumn{1}{g}{\tabularCenterstack{l}{
      \citet{feichtenhofer2021large},
      \citet{han2020memory},
      \citet{kuang2021video}, 
      \citet{liu2021hit},\\
      \citet{ma2021active},
      \citet{pan2021videomoco},
      \citet{qian2021spatiotemporal},
      \citet{xu2021rethinking},
      \citet{yao2021seco} }} \tstrut \bstrut \\
      && && SimCLR \citep{chen2020simple} & \makecell[l]{
      \citet{badamdorj2022contrastive},
      \citet{chen2021rspnet},
      \citet{han2020self},
      \citet{jenni2021time}, \\
      \citet{sun2021composable},
      \citet{wang2020self},
      \citet{yang2020video},
      \citet{zhang2021video}
      } \tstrut \bstrut \\
      && && \multicolumn{1}{g}{CPC \citep{oord2018representation}} & \multicolumn{1}{g}{\tabularCenterstack{l}{
      \citet{akbari2021vatt},
      \citet{bagad2023test}
      \citet{dave2022tclr},
      \citet{li2021motion},
      \citet{miech2020end}, \\
      \citet{park2022probabilistic},
      \citet{parthasarathy2023self},
      \citet{yang2021taco}}} \tstrut \bstrut \\
      && && DIM \citep{hjelm2018learning} & \makecell[l]{
      \citet{bai2022salient},
      \citet{cai2022heterogeneous},
      \citet{feng2023mutual},
      \citet{gordon2020watching}, \\
      \citet{hjelm2020representation},
      \citet{nan2021interventional},
      \citet{sameni2023spatio},
      \citet{sun2019learning}
      } \tstrut \bstrut \\
      && Clustering & \multirow{1}{*}{ 
    \begin{tikzpicture}
    \draw   (0,0.6) -- (1,.6) ;
    \end{tikzpicture}
      } & \multicolumn{1}{g}{SwAV \citep{caron2020unsupervised}} & \multicolumn{1}{g}{
      \tabularCenterstack{l}{
      \citet{coskun2022goca},
      \citet{diba2021vi2clr},
      \citet{long2023cross},
      \citet{toering2022self},\\
      \citet{wei2022inter},
      \citet{yan2020clusterfit}
      }
      } \\
      && \multirow{4}{*}{Self-Distillation} & \multirow{4}{*}{ 
    \begin{tikzpicture}
    \draw    (0,.5) .. controls (.5,.5) and (.5,.85) .. (1,.85) ;
    \draw    (0,.5) .. controls (.5,.5) and (.5,.1) .. (1,.1) ;
    \end{tikzpicture}
      }& BYOL \citep{grill2020bootstrap} & \makecell[l]{
      \citet{escontrela2023video},
      \citet{liu2022funnynet},
      \citet{morales2022leveraging},
      \citet{recasens2021broaden},\\
      \citet{ranasinghe2022self},
      \citet{sarkar2023uncovering},
      \citet{xiong2021self},
      \citet{zhang2022contrastive}
      } \\
       && && \multicolumn{1}{g}{DINO \citep{caron2021emerging}} & \multicolumn{1}{g}{\tabularCenterstack{l}{
       \citet{ding2024betrayed},
       \citet{fan2023unsupervised},
       \citet{huang2024vbench},
       \citet{huang2024uvis},\\
       \citet{ponimatkin2023simple},
       \citet{teeti2023temporal},
       \citet{wang2024videocutler}
       }} \\
       && \multirow{2}{*}{Decorrelation} &\multirow{1}{*}{ 
    \begin{tikzpicture}
    \draw    (0,.66) .. controls (.5,.66) and (.5,.85) .. (1,.85) ;
    \draw    (0,.66) .. controls (.5,.66) and (.5,.45) .. (1,.45) ;
    \end{tikzpicture}
      } & Barlow Twins \citep{zbontar2021barlow} & \multirow{1}{*}{
      \citet{da2022unsupervised},
      \citet{peh2024learning},
      \citet{zhang2022contrastive},
      \citet{zhou2023self}
      } \\
       && && \multicolumn{1}{g}{VICReg \citep{bardes2021vicreg}} & \multicolumn{1}{g}{
       \multirow{1}{*}{
       \citet{bardes2023mc},
       \citet{sun2023unified},
       \citet{yang2023contrastive},
       \citet{yu2024evolve}
       }
       } \\
       \multirow{7}{*}{Masking} &
       \multirow{7}{*}{       
    \begin{tikzpicture}
    \draw    (0,1.8) .. controls (.5,1.8) and (.5,2.7) .. (1,2.7) ;
    \draw    (0,1.8) .. controls (.5,1.8) and (.5,1.8) .. (1,1.8) ;
    \draw    (0,1.8) .. controls (.5,1.8) and (.5,1.25) .. (1,1.25) ;
    \end{tikzpicture}}
    & \multirow{2}{*}{low-level targets} & \multirow{1}{*}{ 
    \begin{tikzpicture}
    \draw    (0,.66) .. controls (.5,.66) and (.5,.85) .. (1,.85) ;
    \draw    (0,.66) .. controls (.5,.66) and (.5,.45) .. (1,.45) ;
    \end{tikzpicture}
      }& ViT \citep{dosovitskiy2020image} & \makecell[l]{
      \citet{girdhar2023imagebind},
      \citet{lin2022frozen},
      \citet{piergiovanni2023rethinking}
      } \\
       && && \multicolumn{1}{g}{MAE \citep{he2022masked}} & \multicolumn{1}{g}{\tabularCenterstack{l}{
       \citet{feichtenhofer2022masked}, \citet{girdhar2023omnimae}, \citet{huang2023mgmae}, \citet{huang2023mavil}, \\ \citet{ryali2023hiera},  \citet{tong2022videomae}, \citet{wang2023videomae}, \citet{wu2023dropmae}}} \\
       && high-level targets & \multirow{1}{*}{ 
    \begin{tikzpicture}
    \draw    (0,.85) .. controls (.5,.85) and (.5,.85) .. (1,.85) ;
    \end{tikzpicture}
      }& BEiT \citep{bao2021beit} & \makecell[l]{
      \citet{cheng2023vindlu},
      \citet{fu2021violet},
      \citet{li2023svitt},
      \citet{tan2021vimpac},
      \citet{wang2022bevt}
      } \\
       && \multirow{2}{*}{Teacher-based} &\multirow{1}{*}{ 
    \begin{tikzpicture}
    \draw    (0,.66) .. controls (.5,.66) and (.5,.85) .. (1,.85) ;
    \draw    (0,.66) .. controls (.5,.66) and (.5,.45) .. (1,.45) ;
    \end{tikzpicture}
      }& \multicolumn{1}{g}{data2vec \citep{baevski2022data2vec}} & \multicolumn{1}{g}{
      \tabularCenterstack{l}{
      \citet{li2023unmasked},
      \citet{lian2023av}
      }
      } \\
       && && MaskFeat \citep{wei2022masked} & 
       \makecell[l]{
       \citet{feichtenhofer2022masked},
       \citet{mizrahi20234m},
       \citet{lin2023smaug},
       \citet{pei2024videomac} \\
       \citet{stergiou2024holistic}, 
       \citet{wang2023masked},
       \citet{woo2023towards},
       \citet{zhao2024asymmetric}
       }  \\
      \end{tabular}
    }
    \label{tab:SSL_tasks}
    \vspace{-1em}
\end{table*}

Early works on signal periodicity \pcite{thangali2005periodic} have decomposed signal repetition with a Fourier analysis \pcite{branzan2008generic,briassouli2007extraction,ousman2008segmentation,ross2000robust,pogalin2008visual}. Signal-based works have also used the direction of motion flow over time \pcite{runia2018real} to count repetitions. Another set of methods approached VRC as a classification task over a finite set of maximum repetitions. \tcite{lu2004repetitive} used dynamic parameters based on the Frobenius norm to classify changes corresponding to action end times.
\tcite{zhang2021repetitive} fused audio and video representations while \tcite{zhang2020context} used multiple cycles to refine the repetition count prediction. \tcite{li2024efficient} extracted action query features and classified the queries by their repetitions. In contrast to defining repetition counts as classes, \tcite{dwibedi2020counting} adopted a temporal self-similarity matrix \pcite{benabdelkader2004gait,junejo2010view,korner2013temporal} to discover repetition periodicity. Subsequent methods have investigated embedding similarity matrices at multiple scales \pcite{bacharidis2023repetition,hu2022transrac}, triplet contrastive losses \pcite{destro2024cyclecl}, and graph representations \pcite{panagiotakis2018unsupervised}. Because embeddings of adjacent frames are highly similar, several recent methods have aimed to limit the discovery of correspondences in repetitions to poses \pcite{ferreira2021deep,yao2023poserac}, specific frames \pcite{li2024repetitive,zhao2024skim}, visual exemplars \pcite{sinha2024every}, or language descriptions \pcite{dwibedi2024ovr}. VRC remains a challenging task given the open-set nature and the lack of robust baselines in recent large-scale datasets \pcite{dwibedi2024ovr}.

\subsubsection{Future outlooks}
\label{sec:recognition::temporal:::outlooks}

Despite the great progress, using unified systems to generalize across tasks remains challenging. For example, STAD methods \pcite{dai2021pdan,tirupattur2021modeling} benchmarked on TAL perform lower than TAL-based models, as their joint objective of localizing both \emph{when} and \emph{where} actions are performed is significantly more challenging to optimize. Similarly, despite the task similarities between TAL and VRC, standard TAL methods do not generalize to VRC as action interruptions and out-of-distribution categories cannot be effectively segmented \pcite{hu2022transrac,sinha2024every}. The recent introduction of unified VLMs for multiple video tasks, e.g., \pcite{wang2024internvideo2} and their use as a feature extractor in subsequent works \pcite{chen2024video}, has shown a promising direction through the use of SSL. Training recipes typically include multiple stages of contrastive and masking pretext self-supervision objectives to allow the generalization of the model to multiple tasks. Training on SSL pretext tasks is a prominent scheme for many video-based models, as shown in \Cref{tab:SSL_tasks}. Context-based approaches rely on inherited spatiotemporal structural relationships in videos. Contrastive objectives are based on instance discrimination tasks, while masking tasks learn representation structures through completion. A possible direction of future research can be the unification of downstream objectives through relevant context, contrastive, and masking pretext tasks based on \emph{the arrow of time}, relationships between task-specific embeddings, or clustering embeddings of semantically similar tasks.

\subsection{Language semantics in videos}
\label{sec:recognition::language}

LLMs have achieved great success in Natural Language Processing (NLP) and have consequently been adapted for action understanding tasks. The relationships between learned context-rich semantic space and visual world attributes are useful for tasks such as caption generation~\pcite{seo2022end,sun2019videobert,wang2024omnivid}, inferring scene information~\pcite{anderson2018vision,cheng2024egothink}, understanding the general context in highlight detection~\pcite{lei2021detecting}, and instructional video learning~\pcite{miech2020end}. Beyond their direct applicability to language-based tasks, they can incorporate vision encoders~\pcite{ashutosh2023hiervl,fu2021violet,kahatapitiya2024victr,song2024moviechat,xu2021videoclip,zellers2021merlot} learning general and semantically-rich representations that can then be used as feature extractors in downstream tasks. However, notable challenges persist despite the popularity of vision-language semantic similarity pretraining for distilling context into video models. 

\subsubsection{Challenges}
\label{sec:recognition::language:::challenges}

\begin{figure}[t]
    \centering
    \includegraphics[width=\linewidth]{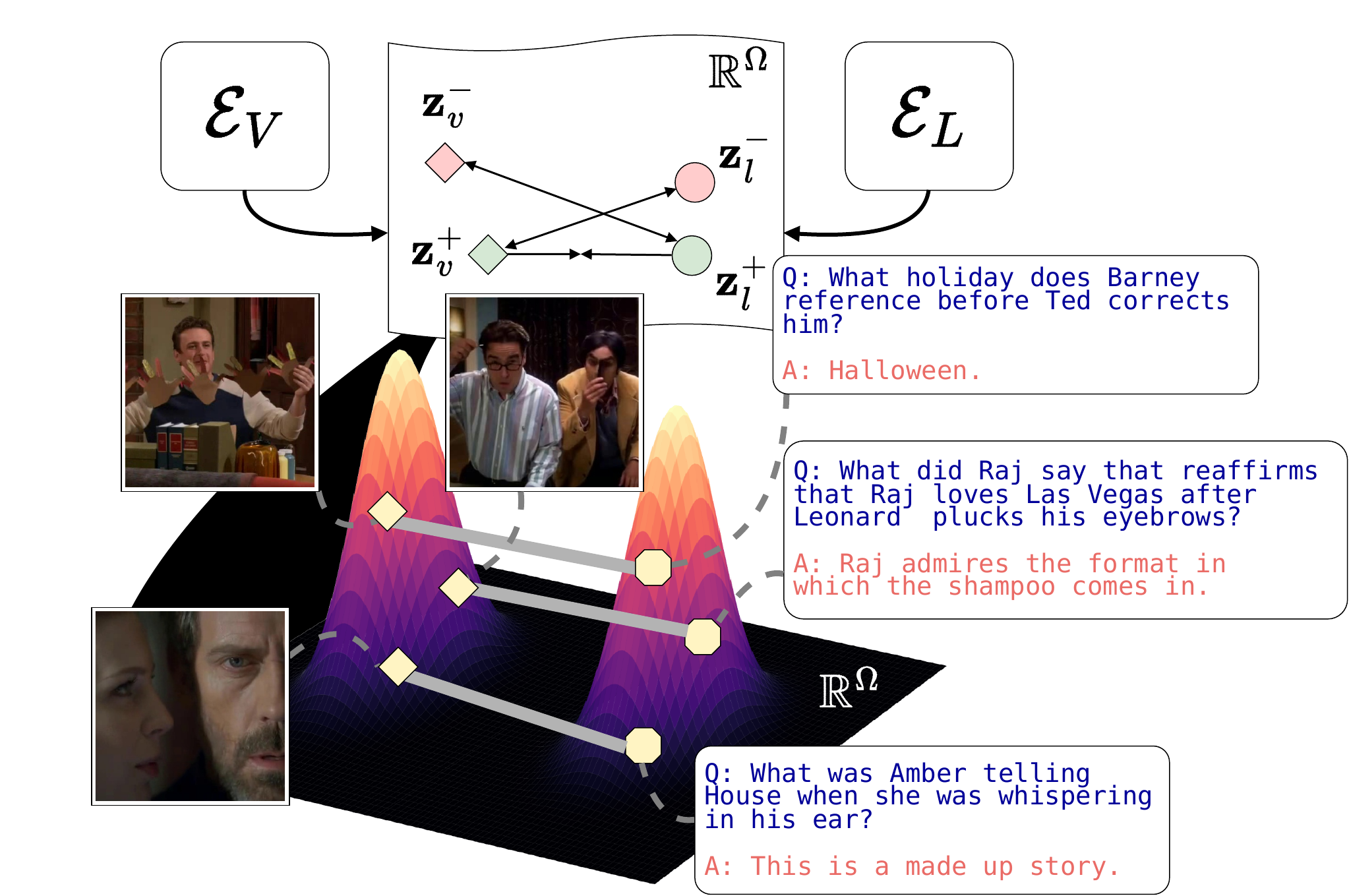}
    \caption{\textbf{VLM modality gap}. Given video encoder $\mathcal{E}_V$ and text encoder $\mathcal{E}_L$, video and text are embedded to $\mathbf{z}_v^{+}$ and $\mathbf{z}_l^{+}$ in a joint embedding space $\mathbb{R}^{\Omega}$. VLM objectives align both $\mathbf{z}_v^{+}$ and $\mathbf{z}_l^{+}$. Contrastive approaches~\pcite{chen2020simple,oord2018representation,xu2021videoclip} additionally maximize the distance between negative vision-language pairs: ($\mathbf{z}_v^{+}$, $\mathbf{z}_l^{-}$) and ($\mathbf{z}_v^{-}$, $\mathbf{z}_l^{+}$). Despite high-level semantic similarity, relevant modality-specific information that is not transferable across modalities can lead to a \emph{modality gap} over embeddings. Videos sourced from \tcite{lei2018tvqa}.}
    \label{fig:mod_gap}
    \vspace{-1em}
\end{figure}

Visual and language information can provide partly complementary perspectives of a video. However, as information from each modality is often heterogeneous, specific representations may not be directly matched through cross-modal correspondence, \eg, due to occluded objects or fine-grained visual details about the performance of the action. Such discrepancies can arise based on domain knowledge specificity or distribution patterns of the available data~\pcite{liang2024foundations}. \textbf{Modality gap}~\pcite{liang2022mind}, shown in \Cref{fig:mod_gap}, is a phenomenon that arises in VLM training in which embeddings of each modality are represented in distinct low-variance regions in the embedding space. In VLMs trained with cross-modal information maximization~\pcite{bain2021frozen,lei2021less,li2020hero,li2022align,zhu2020actbert} this effect becomes stronger with the enforcement of strong coordinate restrictions based on positive and negative cross-modal pairs. This is also relevant to difficulties in the \textbf{cross-modal context alignment} over local elements for tasks with available ground-truth pairs and global representations for tasks without vision-language pairs. In both cases, aligning language and vision context information at either the word/object level or over groups of instances in the embedding space provides a significant challenge in optimization. For generative tasks, this can also lead to difficulties in \textbf{modality-specific generation}. Generating semantic-rich data based on relationships from auxiliary modalities with ambiguous correspondence can impact conditional, stochastic, or auto-regressive generation.

\begin{figure*}
    \centering
    \begin{overpic}[width=\linewidth,keepaspectratio]{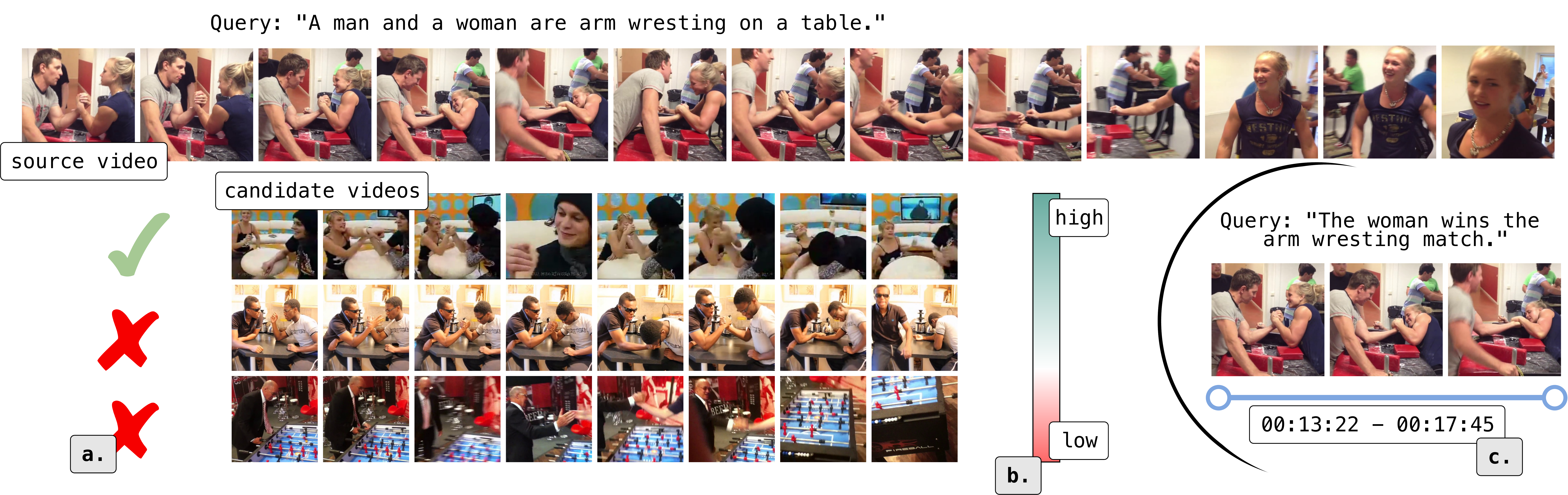}
    \end{overpic}
    \caption{\textbf{Video retrieval tasks}. (a) Instance-based retrieval returns only a \emph{single video} corresponding to a search query.(b) Semantics-based retrieval returns \emph{a ranking score} corresponding to each video's relevance to the search query. (c) Temporal Sentence Grounding (TSG) receives \emph{video segments from queries} and returns the start and end time per segment. Videos sourced from \tcite{xu2016msr}.}
    \label{fig:retreival_tasks}
\end{figure*}

\subsubsection{Vision-language retrieval} 
\label{sec:recognition::language:::vr}

Video retrieval sources relevant videos from a dataset based on an input query in natural language. As shown in \Cref{fig:retreival_tasks}, research works can be categorized into instance- and semantic-based.

Image methods have explored cross-view ranking~\pcite{wang2016learning}, language to visual attention~\pcite{torabi2016learning}, or visual features as embedding targets for language encodings~\pcite{dong2018predicting}. Early adaptation of visual-language approaches to videos have used image-text-video triplets \pcite{otani2016learning}, and related parts of speech to objects and actions \pcite{gabeur2020multi,xu2015jointly}. 

\noindent
\textbf{Instance-based} approaches use a binary score function to rank correspondences. This formulation assumes only a single relevant caption/video for each video/caption.  Refinements to this objective have been made through visual-language binding with the inclusion of parts-of-speech in target captions \pcite{wray2019fine} and dual object-text and action-text models \pcite{liu2019use,mithun2018learning}. A number of methods have studied vision-language pretraining approaches \pcite{ge2022bridging,lin2022egocentric,xue2022advancing}. \tcite{ge2022bridging} related verbs and nouns to questions and video segments. \tcite{xue2022advancing} studied the correspondences between keyframes and all video frames, subsequently contrasting the keyframe-fused video features to language embeddings.

\noindent
\textbf{Semantics-based}. A more challenging task is to retrieve images based on shared semantics to query images~\pcite{gordo2017beyond}. Semantic-based approaches primarily use triplet losses that contrastively regress between text queries and corresponding positive and negative visual inputs. These methods are based on the similarity between every (video, and caption)
pair. Video retrieval works have studied this through either a contrastive objective based on a support set of videos with similar action categories \pcite{patrick2020support} or a semantic
similarity scoring function for videos from the same category \pcite{wray2021semantic}. Recently, \tcite{kim2024you} have utilized prior knowledge in retrieving text features based on embedding correspondences to similar visual features. \tcite{chun2021probabilistic} proposed probabilistic representations of visual features to accommodate multi-query relevance. Similarly, \tcite{li2023progressive} created an object-phrase and event-phrase prototype-matching framework to enforce relations between high-level concepts across modalities. \tcite{hao2024uncertainty} employed an uncertainty estimate based on the Wasserstein distance between source and target domains of text-vision pairs.

\noindent
\textbf{Temporal sentence grounding (TSG)}. TSG \pcite{regneri2013grounding} localizes moments in videos based on provided natural language queries. Compared to retrieving entire videos, TSG only retrieves relevant segments from a video based on queries. The task closely relates to TAL as also shown by the overlapping works \pcite{gao2017tall} jointly exploring the two tasks. However, in contrast to TAL, TSG requires both natural language reasoning between query and answer, and language-vision reasoning with query-video and answer-video relevance. Following \tcite{gao2017tall}, a broad formulation of TSG includes a visual-language semantic alignment between videos and sentences with a regression loss used for temporal sentence localization.

Early approaches have explored region proposals \pcite{chen2018temporally,qu2020fine,liu2018cross}, ranking \pcite{escorcia2019temporal}, distance-based joint vision-language embeddings \pcite{anne2017localizing,rohrbach2016grounding}, and cross-modal graph representations \pcite{liu2022memory,zhang2019man}. As
the association of visual and text features can be performed at multiple levels of abstraction, subsequent methods have learned correspondences over multiple proposals \pcite{xu2019multilevel}, local and global information \pcite{jiang2019cross,mun2020local}, and word/sentence-level cues \pcite{hao2022query}. \tcite{zhang2021natural} used language-guided highlighting by cross-attending text features to multi-resolution video features. Another important aspect of discovering associations between the two modalities is their conditionality, as visual aspects should depend on the descriptions. Approaches have explored step-wise fusion of language key and value tokens 
\pcite{cao2021pursuit}, localizing relevant video features based on text embeddings \pcite{yang2022tubedetr}, matching video segments and text features contrastively \pcite{flanagan2023learning}, and enforcing similarity between sequential tokens \pcite{qian2024momentor}. \tcite{ge2019mac} used both instance-based vision-language embeddings and general category representations to calculate an actionness score and location offset. Other approaches have fused context from global and local temporal resolutions \pcite{liu2021context}, grounded cues from anchor frames and boundary proposals \pcite{wang2020temporally}, related unimodal and cross-modal representations \pcite{nan2021interventional}, and adopted instance-relevant positional information \pcite{gu2024context}. \tcite{goletto2024amego} explored TSG for language hand-object interaction queries.

\begin{table}[t]
    \centering
    \caption{\textbf{Video captioning papers grouped by target task and overall approach}. Tasks are grouped by the generation of single or dense captions and the specialization to coherency with visual storytelling. Approach denotes architectural and model choices.}
    \resizebox{\linewidth}{!}{
    \setlength\tabcolsep{1.0pt}
    \begin{tabular}{c c l}
    \toprule
      Task & Approach & Works \\
      \midrule
      \multirow{4}{*}{\makecell[c]{Single video \\ captioning}} & \cellcolor{LightGrey}{CNN+LSTM} & \multicolumn{1}{g}{\tabularCenterstack{l}{
      \citet{aafaq2019spatio},
      \citet{chen2017generating},\\
      \citet{gan2017semantic},
      \citet{pan2017video},\\
      \citet{wang2018reconstruction},
      \citet{zheng2020syntax}
      }}\\
      & Trnsf-based & \makecell[l]{
      \citet{lin2022swinbert},
      \citet{shen2023accurate},\\
      \citet{yan2023prompt}
      } \\
      &\cellcolor{LightGrey}{VLM} & \cellcolor{LightGrey}{ \citet{chen2024panda}, \citet{seo2022end}}  \\
      \midrule
      \multirow{3}{*}{\makecell[c]{Dense video\\ captioning}} & \makecell[c]{Region \\ proposals} & \makecell[l]{
      \citet{deng2021sketch},
      \citet{iashin2020better},\\
      \citet{iashin2020multi}
      \citet{krishna2017dense},\\
      \citet{li2018jointly},
      \citet{mun2019streamlined},\\
      \citet{shi2019dense},
      \citet{wang2018bidirectional},\\
      \citet{zhou2018end}
      } \\
      & \cellcolor{LightGrey}{MIL} & \multicolumn{1}{g}{\tabularCenterstack{l}{
      \citet{chen2021towards},
      \citet{shen2017weakly}
      }} \\
      \midrule
      \makecell[c]{Visual\\ storytelling} & VLM & \makecell[l]{ 
      \citet{li2019video},
      \citet{yu2021transitional},\\
      \citet{xiao2022hierarchical} ,
      \citet{han2023autoad},\\
      \citep{han2023autoadii},
      \citep{han2024autoadiii}
      }
      \end{tabular}
    }
    \label{tab:captioning_methods}
    \vspace{-1em}
\end{table}

\subsubsection{Video Captioning} 
\label{sec:recognition::language:::vc}

A long-standing challenge in computer vision is the generation of high-level descriptions in language. In contrast to retrieval tasks that depend on a fixed vocabulary, captioning is a generative task. Starting from matching a small corpus of words to objects in images \pcite{barnard2001learning,barnard2003matching}, current works in the image domain are capable of generating diverse and detailed image descriptions \pcite{mokady2021clipcap,alayrac2022flamingo}. Video captioning includes further challenges as the appearance of objects and the context of scenes change throughout the video. Given the temporal extent of videos, captioning tasks can be divided into two categories (\Cref{tab:captioning_methods}).

\noindent
\textbf{Single video captioning}. A large number of works have studied the direct extension of image captioning to video with single captions.  
Given a video clip and a corresponding caption, a general formulation of a single video captioning objective would be the minimization of the log-likelihood of the caption conditioned on the video.

Initial efforts \pcite{guadarrama2013youtube2text} used semantic hierarchies with word selection through decision tree nodes. Following methods \pcite{aafaq2019spatio,chen2017generating,gan2017semantic,pan2017video,wang2018reconstruction} used encoder-decoder architectures that combined CNNs' visual feature extractors and recurrent architectures (RNNs or LSTMs) to generate textual descriptions. To focus on object semantics, \tcite{aafaq2019spatio,zheng2020syntax} included object detector embeddings. The improved context size of transformers has enabled more recent approaches to explore spatio-temporal dynamics in videos and cross-modal relationships. Transformer approaches have included language supervision over hierarchies \pcite{ye2022hierarchical} and token masking \pcite{lin2022swinbert,shen2023accurate,yan2023prompt}. \tcite{seo2022end} used a VLM with the video encoder and language decoder trained jointly on a reconstruction loss. Recently, \tcite{chen2024panda} explored knowledge distillation from multiple VLM models to generate captions. \tcite{majumder2024viewpoint} jointly learned a viewpoint ranking model for video captioning in multi-view settings.

\begin{figure}[t!]
    \centering
    \includegraphics[width=\linewidth]{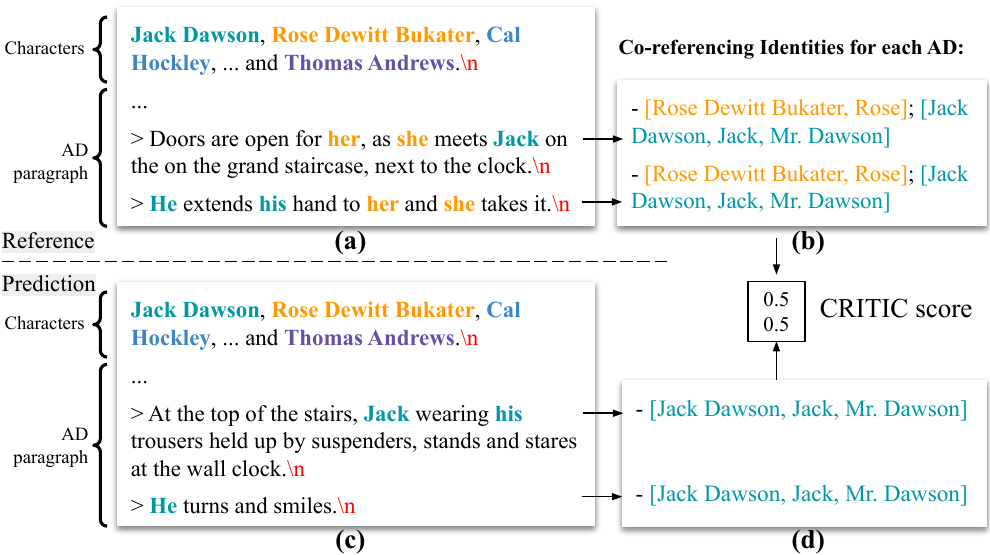}
    \caption{\textbf{CRITIC metric for visual storytelling}. Identities are obtained from character lists and descriptions fed to a co-referencing model. CRITIC \pcite{han2024autoadiii} is calculated as the IoU between predicted and reference identities.}
    \label{fig:CRITIC}
\end{figure}

\noindent
\textbf{Dense video captioning}. Dense video captioning approaches generate multiple captions and temporally ground them to corresponding video segments. This is a significantly more challenging task as distinct video segments need to be localized to generate corresponding captions. Early works on event localization \pcite{krishna2017dense,li2018jointly,shi2019dense,wang2018bidirectional,zhou2018end} were based on proposal modules from extracted video features. To learn vision-language correspondence explicitly for regions of interest, \tcite{zhou2018end} used proposals as masks for visual and language embeddings. Other approaches improved proposal generation by using their sequential occurrence as a prior \pcite{mun2019streamlined}, deployed feature clipping based on proposals \pcite{iashin2020better,iashin2020multi}, refined general captions for each proposal \pcite{deng2021sketch}, and used unique CLIP properties to generate distinct captions \pcite{perrett2024s}. Overall, proposal-based methods are optimized on a loss that relates the proposal interval to the ground truth segment and a captioning loss. As ground truth proposals require exhaustive annotation efforts, more recent works have focused on proposal-free approaches. \tcite{shen2017weakly} used Multi-Instance Learning (MIL) in which word instances are assigned to bags. They used a binary objective to separate positive bags in which at least one instance corresponds to a target word and negative bags in which no instance contains the target word. MIL has been a building block in subsequent weakly-supervised approaches \pcite{chen2021towards}. Further works \pcite{yang2023vid2seq,ren2024timechat} have also used sequence-to-sequence modeling with learnable time tokens for visual-language relations. \tcite{mavroudi2023learning} combined instruction learning to model the sequentially of video captioning. \tcite{islam2024video} used a two-stage autoregressive approach that first generates dense captions for short clips and then cross-attends them to visual features over longer segments to generate longer captions. \tcite{zhou2024streaming} aimed at efficiency improvements by compressing frame-instance visual features to clusters.

\noindent
\textbf{Visual storytelling}. A recently introduced challenging task that is gaining interest is the generation of coherent sentences for sequential videos \pcite{li2019video}. To bridge cross-modal semantics, \tcite{yu2021transitional} used a coherence loss for past, present, and future frames and contrastively pulled visual and language embeddings closer. A similar contrastive objective was used by \pcite{xiao2022hierarchical} alongside masking part of the visual input. \tcite{han2023autoad} trained a mapping module to project joint CLIP visual features, audio descriptions, and subtitles to an LLM input space to generate captions. Following efforts proposed additional refinements in the pipeline by injecting visual and caption embeddings over multiple LLM layers \pcite{han2023autoadii}, using exemplars \pcite{han2024autoadiii}, and using character-based prompting \pcite{xie2024autoad}. The CRITIC metric, shown in \Cref{fig:CRITIC}, was recently introduced by \tcite{han2024autoadiii} to measure the conceptual alignment of the generated sentences.

\subsubsection{Video Question Answering (VideoQA)}
\label{sec:recognition::language:::vqa}

A widely-used benchmark for VLM models is the utilization of visual context to answer natural language questions \pcite{antol2015vqa,goyal2017making}. In contrast to video captioning, it requires understanding parts of objects and the temporal extent of relevant answers. Depending on the task setting, answers can be obtained from multi-choice QA or as a global answer in open-end QA. 

\begin{figure}[t!]
    \centering
    \begin{subfigure}[t]{0.5\linewidth}
        \centering
        \includegraphics[width=\textwidth]{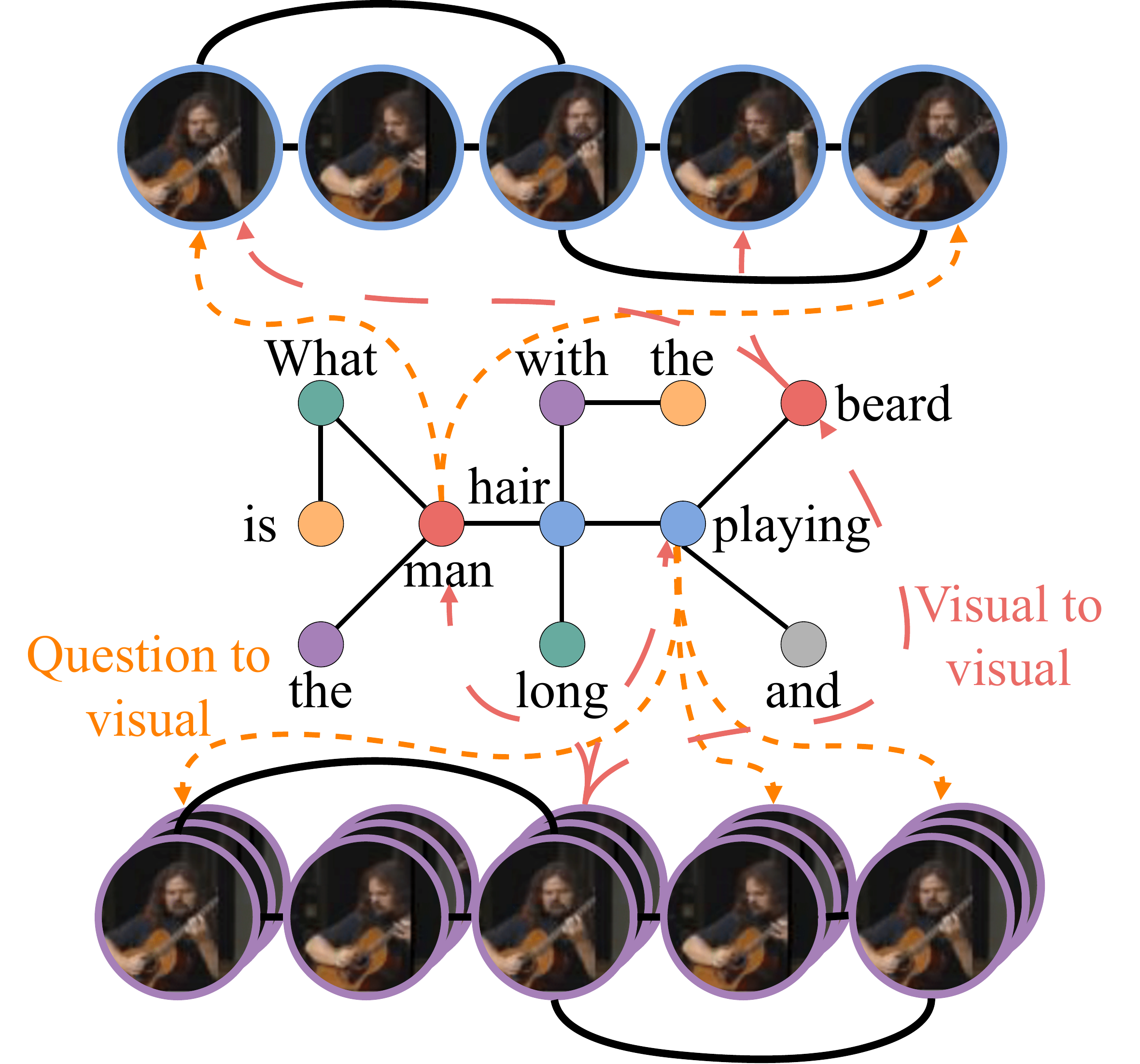}
        \caption{Graph-based}
    \end{subfigure}%
    ~ 
    \begin{subfigure}[t]{0.5\linewidth}
        \centering
        \includegraphics[width=\textwidth]{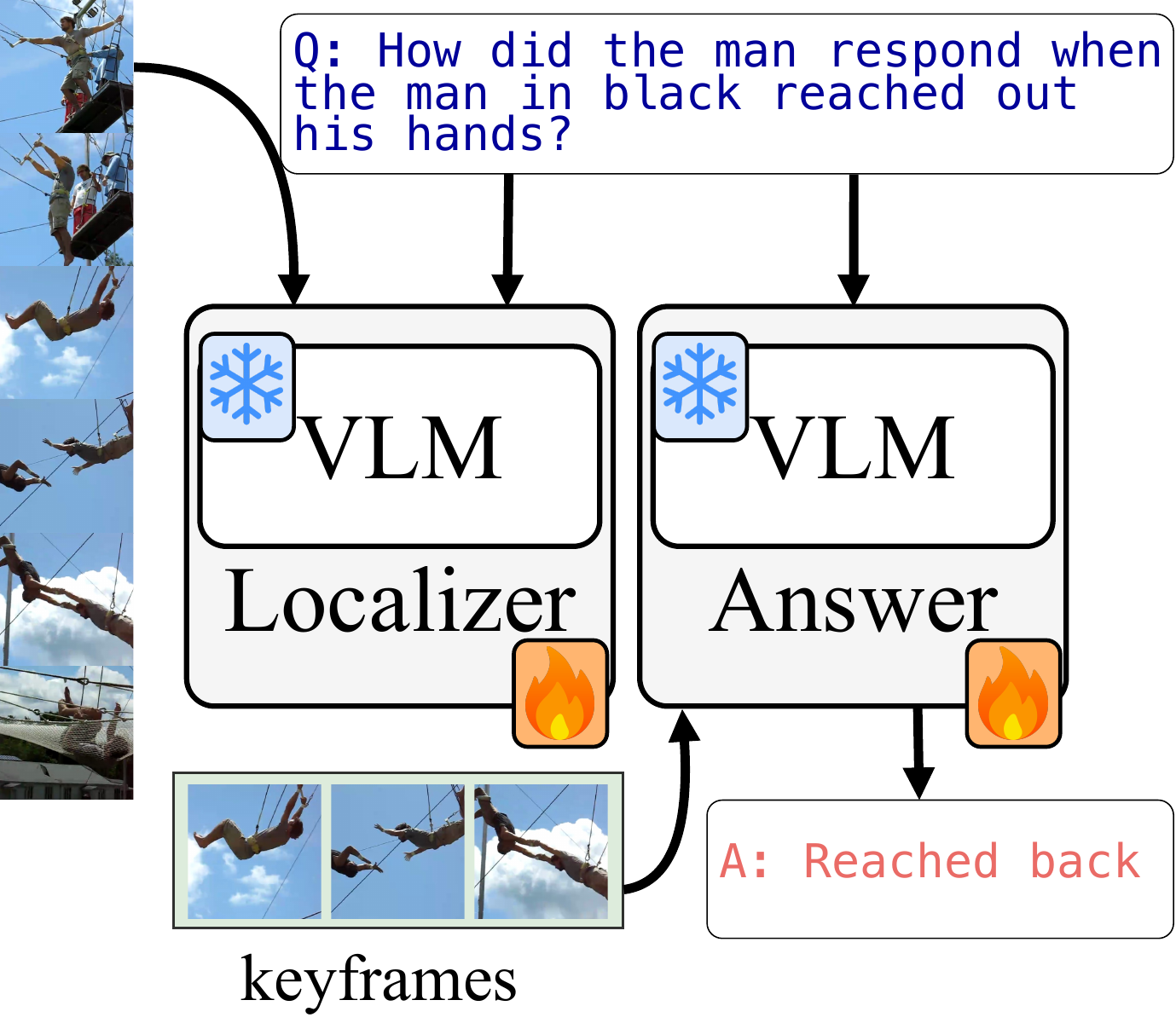}
        \caption{Memory-based}
    \end{subfigure}
    \caption{\textbf{VideoQA approaches}. The graph-based approach in (a) is based on the method from \tcite{park2021bridge}. The memory-based approach with a two-stage VLM in (b) is based on \tcite{yu2023self}. Videos sourced from \tcite{xiao2021next}.}
    \label{fig:videoqa}
\end{figure}

In a multi-choice QA setting, given a video and a question, the goal is to learn a mapping that returns an answer from a set of possible answers. In open-end QA settings, the answer is instead generated from a model conditioned on the video and question. VQA methods can be divided into two broad groups (\Cref{fig:videoqa}).

\noindent
\textbf{Scene-graphs}. Early VideoQA approaches were based on either graph representations \pcite{jiang2020reasoning,tu2014joint} or on each modality's heterogeneity. \tcite{huang2020location} used object and location-based graph embeddings to relate visual and text features with a cross-modal similarity matrix. Graph representations have also been created from hierarchies of objects and their interactions~\pcite{dang2021hierarchical} as well as over multiple frames~\pcite{liu2021hair}. Alternative approaches defined scales from multiple graph convolution resolutions to relate cross-scale interactions \pcite{guo2021multi} or from subgraphs to capture static and dynamic scene objects  \pcite{cherian20222}. \tcite{park2021bridge} created appearance, motion, and question graphs, learning conditionality by propagating nodes across graphs. Graph representations have also been learned contrastively \pcite{xiao2023contrastive} from positive and negative pairs of video snippets and answers.

\noindent
\textbf{Multimodal memory}. Another set of methods aims to memorize relations between visual and text features across time. Initial efforts integrated additional memory modules in LSTMs \pcite{jang2017tgif,xu2017video,zeng2017leveraging}. Attention-based approaches \pcite{ye2017video} combined modality-specific memory modules~\pcite{fan2019heterogeneous} and memory-sharing modules to cross-attend motion and appearance \pcite{gao2018motion,li2019beyond}. Several works \pcite{gao2023mist,li2023discovering,yang2022zero,xue2023egocentric} have used a single model with concatenated language and vision tokens to predict answers to queries. Recent approaches have adapted large VLMs for VideoQA. \tcite{yu2023self} used a two-stage dual-VLM to first localize video segments based on the video and question and then used only the selected frames and question to generate the answer. Similarly, \tcite{min2024morevqa} used a list of generated VLM captions describing scenes in videos as input to an LLM. The question was then passed as a prompt to discover the most relevant answer. However, recent efforts \pcite{xiao2024can} have also revealed that VLM-based approaches may produce answers based on spurious language correlations and not the visual context.

\subsubsection{Future outlooks}
\label{sec:recognition::language:::outlooks}

Advancements in VLMs have enabled the recognition of actions based on their correspondence to a large lexical corpus. Building upon this correspondence, retrieval, captioning, and question-answering models have moved beyond single-instance structural representations and toward the discovery of abstract cross-modal semantics. The increased model capacity provides opportunities for future lines of research.

Most VLMs strongly rely on linguistic associations that may not be relevant in vision instances \pcite{rahmanzadehgervi2024vision}. A possible alternative is to develop unified multimodal models that tokenize and encode video frames and images in the same manner with positional embeddings also encoding temporal relationships. Initial efforts by \tcite{jang2023unifying} and \tcite{jin2024integration} have been promising. Another direction includes a better exploration of the objectives used. The majority of works train models on objectives \pcite{chen2020simple,he2020momentum,oord2018representation} or using downstream task adapters \pcite{hu2021lora} which can enforce properties such as feature suppression \pcite{chen2021intriguing} and pretext granularity \pcite{cole2022does} despite aiming to maximize correspondence. Crafting better alignment objectives for cross-model representations and semantic relevance can benefit future VLM approaches.

\subsection{Multimodal recognition} 
\label{sec:recognition::audio}

The recognition of actions or activities has been predominantly studied in the vision domain. In contrast, the auditory recognition of actions from sounds emitted by objects or actors and their interactions is more sparsely researched. This task presents distinct challenges as the sounds emitted by different objects or actions can be similar. 

Time-frequency spectrograms have been a popular format for representing audio events in videos. Initial audio-based models have been built following image-based object recognition \pcite{gong2021psla} or video classification \pcite{kazakos2021slow} CNNs. Attention-based audio methods have used convolutional features \pcite{gulati2020conformer,kong2020panns} or image-pretrained encoders \pcite{koutini2021efficient} to attend over spectrogram patches. Approaches have also explored patch masking \pcite{baade2022mae,huang2022masked}, focused on salient sounds \pcite{stergiou2023play}, and adapted \pcite{liu2022learning_the} or compressed \pcite{feng2024coarse} spectrogram resolutions. More recently, the use of audio has gained attention in multimodal systems as it can provide supplementary information to both visual features and language context.

\subsubsection{Challenges}
\label{sec:recognition::audio:::challenges}

The use of multiple modalities introduces several challenges. \textbf{Learning cross-modal dynamics} is a fundamental challenge of multimodal models as it aims to preserve heterogeneous properties of modalities while maintaining interconnectivity between modalities \pcite{liang2022foundations}. Fused embedding spaces \pcite{girdhar2023imagebind,girdhar2022omnivore,piergiovanni2023rethinking,zhu2024languagebind} effectively reduce modality-specific information and instead rely on learning a high level of abstraction, with lower heterogeneity and higher interconnectivity. In contrast, modality-specific embedding spaces \pcite{gong2022uavm,gong2023contrastive,chen2024soundingactions,recasens2021broaden} are learned through cross-modal associations and rely on effectively transferring distribution across modalities, causing higher heterogeneity and lower interconnectivity. These paradigms are affected by \textbf{domain-specific noise topologies}. Based on a given task or data distribution, the discriminability of each modality differs. Noise naturally occurs based on environment settings, \eg visual features are more relevant in daylight than in night videos. It can also be observed with instance-based occlusions or sensory imperfections. Such topologies are important when developing reasoning structures over modalities \pcite{gat2021perceptual}. \textbf{Input representation reasoning} can be defined as combining knowledge from the data and the structure of the objective. Compositional relationships between modalities can be established through concept hierarchies, temporal correspondence, or interactive states. Commonly, such structures are not available beforehand and are instead learned in an unsupervised manner.

\begin{figure*}
    \centering
    \begin{overpic}[width=\linewidth,keepaspectratio]{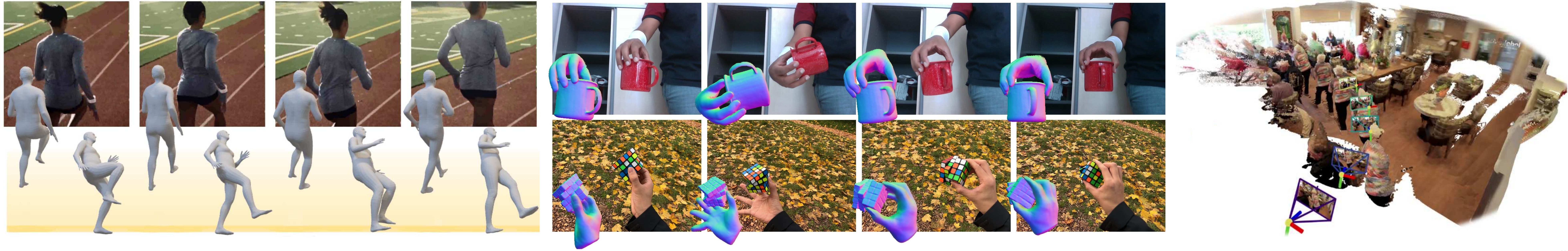}
        \put(2,-2){(a) \textbf{3D pose and shape regression}}
        \put(50,-2){(b) \textbf{HOI}}
        \put(75,-2){(c) \textbf{Dynamic scene rendering}}
    \end{overpic}
    \vspace{0.05em}
    \caption{\textbf{4D video understanding tasks}. (a) 3D human pose and shape regression takes as input monocular videos and produces expressive 3D representations. (b) Human/hand-object interactions predict aspects of human-object interactions, such as the contact area or the grasp. (c) Dynamic scene rendering estimates the per-timestep geometry of scenes from sets of images. Figures sourced from \citet{dwivedi2024tokenhmr,fan2024hold,zhang2025monst3r}.}
    \label{fig:4d_tasks}
\end{figure*}

\subsubsection{Audio-visual models} 
\label{sec:recognition::audio:::avmodels}

As video and audio signals differ significantly, works have used two-step models to infer predictions. Two-step approaches extract video and audio embeddings first and then fuse either modality-specific predictions \pcite{fayek2020large}, embeddings from multiple modalities \pcite{xiao2020audiovisual}, or they jointly attend vision and audio features for the final prediction \pcite{gong2022uavm}. More recently, architectures have tokenized and attended audio and vision jointly with multimodal learnable tokens \pcite{nagrani2021attention}, cross-modal attention \pcite{jaegle2021perceiver}, and modality gating \pcite{xue2023dynamic}. To account for models trained on unimodal tasks, \tcite{lin2023vision} proposed cross-modal adapters to combine unimodal embeddings in multimodal tasks. Exploring the relevant audio and visual features with self-supervision has also been a learning paradigm of significant interest. Common embedding spaces can be useful for discovering correspondences in both unimodal and cross-modal retrieval \pcite{arandjelovic2018objects,wu2021exploring}, multimodal clustering \pcite{hu2019deep}, and sound source separation \pcite{hu2022mix,mo2023unified,zhao2018sound}. Token reconstruction through masking has also been used as a self-supervised pretraining task with a variety of training schemes, including concatenating masked tokens \pcite{gong2023contrastive}, multi-view masking per modality \pcite{huang2023mavil}, fusing a mixture of per-modality masked tokens \pcite{guo2024crossmae}, combining modality-specific masked and unmasked embeddings \pcite{georgescu2023audiovisual}, and using multiple masking ratios with siamese networks \pcite{lin2024siamese}. 

Variations in the relevance of visual or auditory signals depend on instances. A promising direction for integrating this into optimization is gradient blending \pcite{wang2020makes} which recalibrates per-modality losses. Other works explored multi-audio to single-visual scene correspondence with contrastive learning. This was done by utilizing joint semantic similarity in both modalities \pcite{morgado2021audio}, using active sampling to diversify negative sample selection
\pcite{ma2021active}, and by counterfactual audio and video pairs to enforce a relationship between multi-audio to single visual scenes \pcite{singh2024looking}. Enforced audio and vision steams similarities can also be used to train models on incremental tasks \pcite{pian2023audio}.

\subsubsection{Gaze and vision models} 
\label{sec:recognition::audio:::gazemodels}

Gaze can be used as a saliency cue to direct attention or processing priority towards target focal areas \pcite{itti2002model}. Several egocentric activity datasets \pcite{huang2024egoexolearn,li2018eye,pan2023aria} record gaze as an additional modality. Based on gaze inputs, gaze fixations have been used to recognize objects being manipulated \pcite{land2001ways}, and discover ways of interacting with them \pcite{damen2024youdo}. Observing gaze patterns, including fixations and saccades, in conjunction with scene appearance has also been used to predict future gaze behavior \pcite{huang2018predicting}.

Because gaze and action are tightly coupled \pcite{vickersadvances}, subsequent research has focused on action recognition with the additional availability of gaze information. For example, \tcite{fathi2012learning} probabilistically modeled the joint distribution of gaze target, scene objects, and action label. \tcite{min2021integrating} modeled gaze fixations as latent variables for activity classification. \tcite{xu2015gazeenabled} used gaze to identify relevant actions in long videos, to summarize egocentric videos. More recently, gaze patterns have been used to identify deviations from expected executions of procedural activities \pcite{mazzamuto2025gazing}.

Gaze has also been used to train image and video models with gaze-free inference. \tcite{liu2021goal} learned to attend discriminative local features for zero-shot object identification. In general, the additional availability of gaze has shown improvements in training across egocentric video understanding tasks \pcite{kapidis2023multi}. 

Several works have also addressed the opposite process of estimating salient regions likely to be looked at given an image or video. While initial works introduced bottom-up algorithms \pcite{borji2012state}, later research increasingly considered higher-level information in predicting gaze targets \pcite{judd2009learning,torralba2006contextual}. In egocentric videos, it was shown that object presence and, in particular, object manipulation strongly directed eye gaze \pcite{tavakoli2019digging}. This notion has led to fusion approaches that first process either global scene context and local saliency independently \pcite{lai2024intheeye}, or static and dynamic information as separate branches of a two-stream model \pcite{lu2019deepattention}.

Other methods focused on action- or task-dependent gaze estimation \pcite{huang2018predicting}. \tcite{li2018eye} jointly determined gaze targets and the person's actions. \tcite{huang2020mutual} jointly modeled gaze-conditioned action recognition and action-conditioned gaze estimation in a single network. \tcite{chong2020detection} recognized eye contact in egocentric videos. Models that predict gaze over subsequent frames have also been introduced \pcite{zhang2017deepfuture}. Recently, auditory information was included to improve gaze predictions \pcite{lai2024listen}.

Estimation of eye gaze in third-person perspectives additionally considers a person's body and head orientation \pcite{chong2020detecting,marinjimenez2021laeonet} to estimate the focus of their gaze. \tcite{recasens2017following} extend the gaze prediction to targets that appear in subsequent frames by learning correspondences between views. Increasingly, gaze predictions in third-person videos consider multiple persons to include social context \pcite{tafasca2024sharingan}.

\subsubsection{4D-vision methods} 
\label{sec:recognition::audio:::4dmethods}

While the majority of the research has addressed 2D video and time as the third dimension, increasingly scenes are modeled in 3D, leading to the space-time-depth (4D) reconstruction of humans, objects, and scenes. We provide a visualization of such tasks in \Cref{fig:4d_tasks} and discuss the main approaches and objectives per task below.

\noindent
\textbf{Pose and shape regression}. Initial efforts for human pose and shape regression from monocular images primarily captured the overall body shape and pose \pcite{allen2003space,allen2006learning,loper2015smpl}. Additionally, finer details were modeled with either MANO \pcite{romero2017embodied} for hands or the Frank model \pcite{joo2018total} for faces. Unified parametric frameworks were later introduced to produce pose and shape details through keypoints \pcite{hassan2019resolving,kolotouros2019learning,pavlakos2019expressive,xu2020ghum} or silhouettes \pcite{kanazawa2018end,omran2018neural}. Attention-based methods have also been used with either pseudo-ground-truths to align body-to-image projections \pcite{joo2021exemplar,li2022cliff,moon2022neuralannot} or with probabilistic representations \pcite{li2024coin,sengupta2023humaniflow,stathopoulos2024score,zhang2023probabilistic}. Recent methods have used quantized human poses as a prior \pcite{dwivedi2024tokenhmr,fiche2024vq}.

\noindent
\textbf{3D human and object interaction (HOI)}. HOI tasks seek to capture human and object relations in 3D. They include estimations of dense human-object contact \pcite{huang2024intercap,jiang2023full,nam2024joint,yang2024lemon,xie2022chore} through 2D image-based semantics and geometric correlations. Other approaches learned object affordances \pcite{mo2021where2act,zhai2024background} by mapping them to their shapes, human-object interactions, and spatial relations \pcite{liu2023contactgen,xu2023interdiff,xu2025interdreamer}. Extensions to these tasks have also included human-human and human-scene interactions \pcite{fieraru2020three,huang2024closely,yin2023hi4d}, self-contact \pcite{fieraru2021learning,muller2021self}, and predicting human and scene layouts \pcite{huang2022capturing,zhang2020perceiving}. 

In hand-object interactions, methods used pre-defined object templates to estimate poses from 3D control points
\pcite{hampali2020honnotate,tekin2019h+} and temporally-consistent sparse labels \pcite{hasson2020leveraging,liu2021semi}. Other template-based methods have explored
object-centric grasp prediction from frames \pcite{corona2020ganhand,hasson2019learning}, contact prediction from hand and object meshes \pcite{grady2021contactopt,zhu2023get}. \tcite{zhang2024graspxl} used generative models to learn possible HOI sequences conditioned on sets of motion objectives and hand-object states. Template-free approaches have also gained interest for reconstructing novel objects in in-the-wild settings. The majority of these methods employ dual-branch models \pcite{leng2023dynamic,tse2022collaborative,xu2023h2onet} that directly predict pose, shape, and camera from frames \pcite{dong2024hamba,pavlakos2024reconstructing}. Recently, \tcite{fan2024hold} jointly reconstructed hands and objects in monocular videos by refining initial structure-from-motion estimates through a per-frame-texture and shape regression objective followed by hand/object pose constraint fine-tuning.

\noindent
\textbf{Dynamic scene rendering}. Rendering tasks estimate 3D scene structures from sets of 2D images for static scenes, and frames for dynamic scenes. Although well-established approaches such as structure-from-motion \pcite{schonberger2016structure,teed2021droid}, LSD-SLAM \pcite{engel2014lsd}, and ORB-SLAM \pcite{mur2017orb} have been promising for static scenes, dynamic scene rendering remains an active challenge. Recent self-supervised methods have jointly estimated depth, camera pose,
and residual motion, with motion segmentation \pcite{gordon2019depth,godard2019digging,kopf2021robust,zhang2022structure}. Another set of methods focuses on 4D dynamic scene reconstruction by space-time optimization of 3D Gaussians \pcite{chu2025dreamscene4d,lei2024mosca,liu2025modgs} to synthesize novel views over both space and time. As the joint learning of geometry and motion can be difficult to learn end-to-end, \tcite{wang2024dust3r} introduced a point-map scene geometry approach in which, given a pair of images, per-image pixels are mapped to discrete 3D locations and then accumulated to a global point cloud. Model pre-training was conducted over cross-view tasks \pcite{weinzaepfel2023croco}. This approach has prompted further extensions through assigning pointmaps to single points in time \pcite{zhang2025monst3r}, coupling intermediate predictions \pcite{li2024megasam}, and including a depth estimation model \pcite{lu2024align3r}. Other approaches have aimed to improve speed \pcite{liang2024feed}, and jointly reconstruct scenes and recover human meshes \pcite{liu2025joint}.

\subsubsection{Multimodal models}
\label{sec:recognition::audio:::mmmodels}

Video is a natural source of multimodal data. Apart from visual information, audio or textual descriptions can be used in tandem to provide additional signals for actions and events at different granularities. Multimodal learning has shown improvements in the generalizability of unsupervised models \pcite{ngiam2011multimodal} over varying tasks \pcite{paredes2012exploiting}. An initial effort by \tcite{kaiser2017one} aimed to create unified multimodal representations with modality-specific encoders and modality-binding decoders. A similar mixture-of-expects approach was also presented by \tcite{munro2020multi} for unsupervised domain adaptation with a dual cross-domain source-target loss over modality pairs. \tcite{dai2022one} used sparse activations to train portions of a unified model on specific modalities and tasks. Multimodal transformers have introduced joint encoder paradigms. \tcite{akbari2021vatt} used modality-specific heads to project outputs from a joint audio-text-video encoder trained with a contrastive loss over modality pairs from \tcite{miech2020end}. Mixtures of modality-specific encoders and multimodal head/decoders have also been trained with masked tokens \pcite{zellers2022merlot}, cross-modal attention blocks in the encoder \pcite{recasens2023zorro}, ensembles of unimodal teachers \pcite{radevski2023multimodal}, and audio-vision projectors on top of LLM heads \pcite{zhang2023video}. \tcite{zhang2024multimodal} fused features from different modalities to a Multimodal head at different training steps capturing cross-modal associations iteratively during training. \tcite{srivastava2024omnivec} included meta tokens to represent modality dimensions and channels to embed modality-specific features in a common space. This was further adjusted \pcite{srivastava2024omnivec2} to also cross-attend joint-embedded features and unimodal features.

\subsubsection{Future outlooks}
\label{sec:recognition::audio:::outlooks}

Most current multimodal models rely on the availability of all modalities at the start. New models are re-trained when additional modalities are added. A promising direction would be to design adaptive models that integrate unseen modalities more efficiently \pcite{ma2022multimodal}. This has been explored by recent methods by transferring seen to unseen modality distributions \pcite{wang2023distribution}, cross-attending over unseen modalities \pcite{recasens2023zorro}, aligning unimodal and multimodal features in training \pcite{zhang2023learning}, using modality-specific adapters \pcite{lin2023vision}, and predicting missing modality features with learnable tokens \pcite{kim2024missing}. Learning adaptable models that can process inputs in new modalities at inference time not only benefits performance for specific tasks but also enables advancement in more general tasks such as online learning \pcite{bottou1998online}, incremental learning \pcite{schlimmer1986case,utgoff1989incremental}, and federated learning \pcite{konevcny2016federated}.

\begin{figure}[t]
    \centering
    \includegraphics[width=\linewidth]{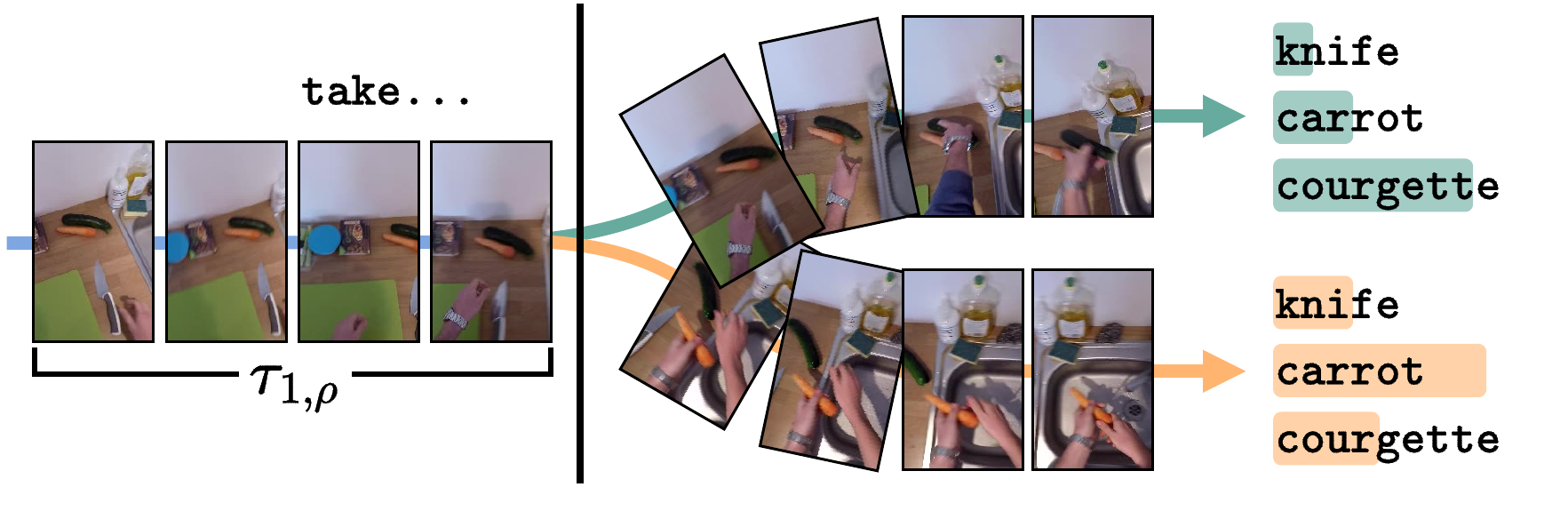}
    \caption{\textbf{Early Action Prediction (EAP)}. Only the observable part of a video $\tau_{1,\rho}$ is used to predict the current action. EAP is challenging as the immediate future is often unpredictable especially when distinguishing between fine-grained actions, e.g., \textit{taking courgette} or \textit{taking carrot}. Video from \tcite{damen2022rescaling}.}
    \label{fig:EAP_overview}
\end{figure}

\section{Predictions in ongoing actions}
\label{sec:prediction}

A critical aspect of video models regardless of the downstream task has been their ability to capture temporal information \pcite{huang2018makes}. Learning temporal patterns can enable models to predict what is happening in a video, without seeing the full action or activity. We start by defining methods that provide semantic predictions on the action categories from partial observations in \Cref{sec:prediction::EAP}. We then discuss approaches that generate unobserved frames in \Cref{sec:prediction::VFP}. States of actions or objects can change at different times during the execution of an action. We explore groups of tasks that predict the states of objects and actions in \Cref{sec:prediction::states}.

\subsection{Early action prediction}
\label{sec:prediction::EAP}

Early Action Prediction (EAP) assumes predictions made based on the observable part of an \emph{ongoing} action being performed $\tau_{1,\rho}$ as shown in \Cref{fig:EAP_overview}. 
Several different lines of research have been explored to address relationships between partial action observation and high-level semantics.

\subsubsection{Challenges}
\label{sec:prediction::EAP:::challenges}

One of the main challenges that arise from partially observed videos is the \textbf{procedural proximity} in the execution of actions. As shown in \Cref{fig:EAP_overview}, there can be an overlap in the steps taken to perform similar activities. For the example shown, the difference between \textit{take courgette} and \textit{take carrot} is subtle, without significant motion variations. Instead, the distinction only becomes visually apparent at the end of the action. This challenge relates to the more general problem of \textbf{abstracted views} in which deterministic information relevant to the task is not always available. Although such fine-grained predictions may not be available, it is still possible to identify general categories for the actions being performed, e.g., \textit{take} $<$object$>$. An \textbf{adjustability requirement} is thus introduced as part of EAP to address the intrinsic uncertainty in partial observation.

\begin{figure}[t]
    \centering
    \begin{overpic}[width=\linewidth]{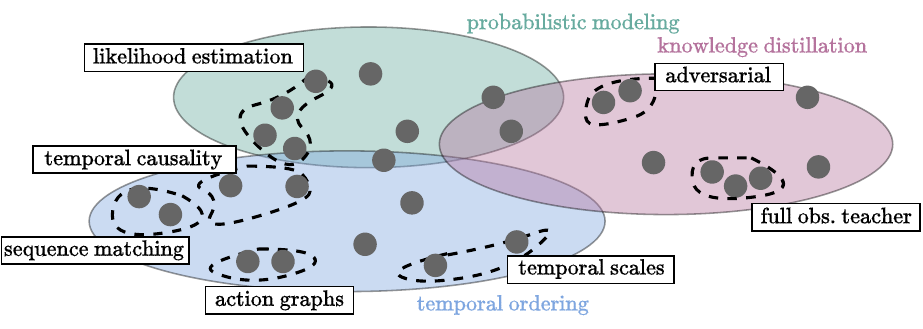}
    \put (36,23.9) {1.}
    \put (25.5,25.8) {2.}
    \put (22,20.2) {3.}
    \put (34.5,19) {4.}
    \put (37.5,29) {5.}
    \put (40.5,23) {6.}
    \put (51,27) {7.}
    \put (8,14) {8.}
    \put (14,11) {9.}
    \put (42.5,17.7) {10.}
    \put (42.5,8.5) {11.}
    \put (38,13.5) {12.}
    \put (23,12.5) {13.}
    \put (30,8.8) {14.}
    \put (20.5,7.7) {15.}
    \put (37.5,11) {16.}
    \put (53,10.8) {17.}
    \put (31.5,12.4) {18.}
    \put (74,18.8) {19.}
    \put (75.4,11.5) {20.}
    \put (81,18) {21.}
    \put (64,16.5) {22.}
    \put (59,25) {23.}
    \put (65.5,27.8) {24.}
    \put (88.5,19) {25.}
    \put (87.5,26.7) {26.}
    \put (55.4,18.5) {27.}
    \end{overpic}
    \resizebox{\linewidth}{!}{
    \begin{tabular}{lll}
         1. \citet{cao2013recognize} &
         2. \citet{hoai2014max} & 
         3. \citet{li2012modeling} \\
         4. \citet{li2014prediction} &
         5. \citet{ryoo2011human} &
         6. \citet{suris2021learning} \\
         7. \citet{chen2022ambiguousness} &
         8. \citet{misra2016shuffle} &
         9. \citet{zhou2015temporal} \\
         10. \citet{xu2015activity} &
         11. \citet{kong2014discriminative} &
         12. \citet{kong2018action} \\
         13. \citet{zhao2019spatiotemporal} &
         14. \citet{wu2021spatial} &
         15. \citet{wu2021anticipating} \\
         16. \citet{wang2023magi} &
         17. \citet{stergiou2023wisdom} &
         18. \citet{rangrej2023glitr} \\
         19. \citet{cai2019action} &
         20. \citet{fernando2021anticipating} &
         21. \citet{wang2019progressive} \\
         22. \citet{hou2020confidence} &
         23. \citet{xu2019prediction} &
         24. \citet{zheng2023egocentric} \\
         25. \citet{xu2023dynamic} &
         26. \citet{foo2022era} &
         27. \citet{hu2018early} \\
    \end{tabular}
    }
    \caption{\textbf{EAP methods} grouped by approach. The three main clusters are colored. Smaller subgroups are denoted with dashed lines. The positioning of the works represents an abstract proximity of the research idea to other seminal works.}
    \label{fig:eap_methods}
    \vspace{-1em}
\end{figure}

\subsubsection{Approaches}
\label{sec:prediction::EAP:::approaches}

Three main groups of approaches can be identified for EAP. We visualize these groups in \Cref{fig:eap_methods}.

\noindent
\textbf{Probabilistic modeling}. A large portion of the EAP literature has originally been based on probabilistic modeling of action classification from partial observations \pcite{cao2013recognize,hoai2014max,li2012modeling,li2014prediction,ryoo2011human}. \tcite{ryoo2011human} used a bag of words based on feature distributions. This division into segments has been relevant in subsequent approaches that used sparse coding \pcite{cao2013recognize}, max-margin \pcite{hoai2014max}, and scoring functions \pcite{li2012modeling,li2014prediction} to infer the action likelihood. More recent probabilistic approaches include the use of hyperbolic representations \pcite{suris2021learning} for hierarchical predictions of actions. The ambiguity of future predictions has also been explored with the generation and subsequent selection of multiple future representations \pcite{chen2022ambiguousness}.

\noindent
\textbf{Temporal ordering}. A different line of works explored EAP based on the temporal evolution of the action. The arrow of time \pcite{pickup2014seeing} can provide a strong signal to associate the procedural understanding of actions with high-level categorical semantics \pcite{misra2016shuffle,zhou2015temporal}. \tcite{xu2015activity} formulated EAP with an auto-completion objective, matching candidate futures to a partial action observation query. The predictability of partial observations can be difficult in instances where there are visual similarities in the performance of actions. To address this, approaches have either used multiple temporal scales \pcite{kong2014discriminative}, created key-value memories of representations \pcite{kong2018action}, or propagated the features' residuals over time \pcite{zhao2019spatiotemporal}. More recent approaches have used temporal graph representations \pcite{wu2021spatial,wu2021anticipating}, contrastive learning over partial observations of the same action \pcite{wang2023magi}, or aggregated attention over temporal scales \pcite{stergiou2023wisdom}, and relevant space-time regions \pcite{rangrej2023glitr}.

\noindent
\textbf{Knowledge distillation}. Transferring class knowledge \pcite{park2019relational} from models trained on the full videos can be an effective technique for refining predictions from partial observations. \tcite{cai2019action}, \tcite{fernando2021anticipating}, and \tcite{wang2019progressive} used learned representations of the full observations as targets to optimize for partial observations. Further methods \pcite{hou2020confidence} have refined this approach with the inclusion of motion sequentiality to learn soft targets and regress model predictions. In a similar effort, \pcite{xu2019prediction} and \pcite{zheng2023egocentric} integrated an adversarial objective for generating representations for the non-observable parts. Similarly, \tcite{xu2023dynamic} learned to reconstruct representations of full observations with a masked autoencoder \pcite{he2022masked}. Other works have fine-tuned expert heads for each action category \pcite{foo2022era} or learned by focusing on videos with distinct visual features \pcite{hu2018early}.

\subsubsection{Future outlooks}
\label{sec:prediction::EAP:::outlooks}

Although EAP remains a challenging task to be explored further, some future directions can be envisioned. First, current EAP evaluation protocols are based on fixed-length temporal occlusions of parts of videos. However, this offline evaluation varies significantly from the intended \textbf{real-time use} of these systems in which singular models are deployed in video streams. EAP methods should instead be built and evaluated in real-time settings in which factors such as latency and inference speeds are crucial.

A second direction of future research is the exploration of \textbf{multi-person action prediction} for group activities. This is a significantly more complex task as it not only requires predicting the intentions of individuals but also general group goals. Such approaches will also have direct application to more general fields such as robotics, security, and augmented reality.

\subsection{Frame-level prediction}
\label{sec:prediction::VFP}

Related to EAP, Video Frame Prediction (VFP) aims to reconstruct future frames of ongoing actions from partial observations. Although high-level semantics such as the level of semantic abstraction to describe the observed action are not learned, VFP still requires relating the consequentiality of motions and intended action to the reconstruction of subsequent frames.

\subsubsection{Challenges}
\label{sec:prediction::VFP:::challenges}

The \textbf{metric-based evaluation} of VFP approaches is done deterministically as VFP aims to predict raw pixel values of future frames. Metrics such as Peak-Signal-to-Noise Ratio (PSNR) and Structural Similarity Index Measure (SSIM) have been widely adopted to quantify VFP performance. However, they do not evaluate the correctness of high-level scene dynamics or the consistency of the scene. A number of image-based statistics adjusted to video \pcite{czolbe2020loss,ding2020image,zhang2018unreasonable} and video-specific statistics \pcite{hou2022perceptual,li2019quality} have targeted such shortcomings by including comparisons in embedding representations between predicted and ground truth frames. The usability of these metrics still leaves room for exploring the robustness of quality validation approaches further.

Similar to EAP, \textbf{future scene dynamics predictions are stochastic}, with varying levels of complexity. Most VFP objectives are based on the changes in the pixel distributions between frames without explicit definitions of optimization criteria for a comprehensive understanding of physics or object structures within scenes. This hinders the prediction capabilities over longer temporal windows and complex scenes with fast motions \pcite{ming2024survey}.

\subsubsection{VFP methods}
\label{sec:prediction::VFP:::methods}

We identify three groups of VFP approaches.

\noindent
\textbf{Sequential adversarial predictions}. A significant portion of VFP works has been based on sequential frame generation \pcite{castrejon2019improved,chaabane2020looking,chang2021mau,chang2022strpm,chen2017learning,guen2020disentangling,hwang2019adversarial,jin2020exploring,liang2017dual,villegas2018hierarchical,wang2018predrnn++,wu2021motionrnn}. These approaches use recursion to generate representations or predictions in an autoregressive manner. One line of methods \pcite{chen2017learning,jin2017video} focused on the correspondence of objects between frames to guide the generation of the next frames. \tcite{castrejon2019improved} used similar adversarial guidance by fusing context information from previous frames. Additional supervisory signals included motion flow \pcite{liang2017dual}, partial differential equations \pcite{guen2020disentangling}, and embeddings over multiple temporal resolutions \pcite{gao2022simvp}. Another line of approach \pcite{chang2021mau,villegas2018hierarchical,wang2018predrnn++} has included long-term memory connections to discover causalities from frames over longer temporal windows. \tcite{park2021vid} incorporated time dynamics for VFP with the inclusion of Ordinary Differentiable Equations (ODE). \tcite{davtyan2023efficient} used ODE with the previous frame as the initial condition and integrated the vector field from Flow Matching \pcite{lipman2022flow} to predict the next frame.

\noindent
\textbf{Parallel multi-frame synthesis}. In contrast to the sequential reconstruction of future frames, approaches have also generated multiple future frames in a single step. One of the first efforts for multi-frame prediction \pcite{liu2017video} used a multi-frame per-pixel optical flow vector with further adaptations including multiple scales \pcite{hu2023dynamic}. Attention-based architectures have also been used for parallelization of frame prediction by introducing encodings of context for frame prediction attended over temporal patches \pcite{tan2023temporal,ye2023unified}, conditioning the generation based on short-term representation variations \pcite{hu2023dynamic,smith2024convolutional}, using multiple motion and appearance scales \pcite{zhong2023mmvp}, and reducing inference speeds \pcite{ye2022vptr,tang2024vmrnn}. As an extension to spatiotemporal attention, \tcite{nie2024triplet} used a triplet module to attend across all dimensions of the video sequentially.

\noindent
\textbf{Probabilistic generation}. A final group of approaches studied the reconstruction of future frames probabilistically. \tcite{babaeizadeh2018stochastic} and \tcite{denton2018stochastic} learned a probabilistic variational model on the stochasticity of the video to generate frame predictions. \tcite{wang2020probabilistic} models the perceptual uncertainty in future frames with a Bayesian framework with different weights assigned to future prediction candidates. Diffusion-based models \pcite{dhariwal2021diffusion,ho2020denoising,rombach2022high} have been applied to a multitude of generative approaches for VFP \pcite{gu2023seer,hoppe2024diffusion,shrivastava2024video,voleti2022mcvd,ye2024stdiff,zhang2024extdm}. These methods gradually transform a complex distribution into unstructured noise and learn to progressively recover the original distribution from noise.

\subsubsection{Future outlooks}
\label{sec:prediction::VFP:::outlooks}

The scarcity of \textbf{high-resolution datasets} remains a limiting factor for the performance of VFP models. More varying data distributions in terms of motions in scenes, sharpness, and blur can enable current approaches to learn across diverse video conditions. Recently, efforts \pcite{stergiou2024lavib,xue2022advancing} have aimed to provide high-resolution videos for several tasks. Future VFP approaches can benefit from training on these datasets. 

A point of improvement for future works is the \textbf{inclusion of world knowledge} to enable predictions based on abstractions of the scene dynamics. The recently introduced term \textit{Stochastic Inverse Problems} defines a broad family of problems relating to predictions from partial observations \pcite{spielberg2023differentiable,tewari2023diffusion}. These approaches aim to permeate knowledge of physical underlying processes throughout training.

\begin{figure*}[t]
\centering
\begin{subfigure}[b]{\textwidth}
\centering
\includegraphics[width=\textwidth]{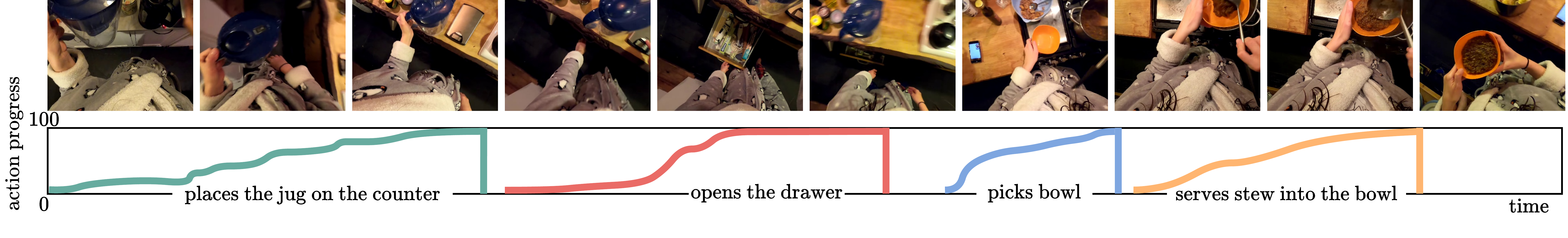}
\caption{\textbf{Action Progress Prediction (APP)}. Given a video stream of a procedural task, estimate the progress of each ongoing action by inferring the time it will take to complete the action performed. Video sourced from \tcite{grauman2024ego}. \vspace{1em}}
\label{fig:states::progress}
\end{subfigure}
\begin{minipage}{0.49\textwidth}
\begin{subfigure}{\linewidth}
\includegraphics[width=\linewidth]{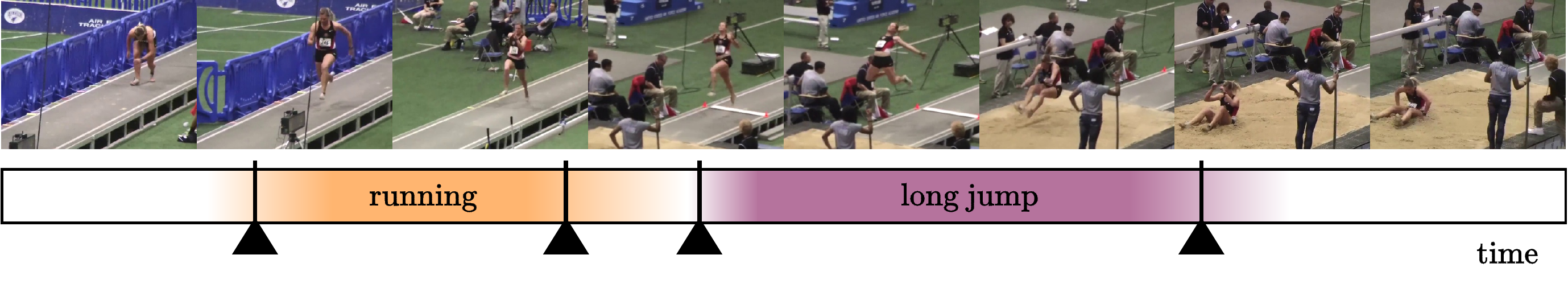}
\caption{\textbf{Event Boundary Detection (EBD)}. Detect the start and end times of ongoing events in video streams. Video sourced from \tcite{carreira2017quo}. \vspace{1em}}
\label{fig:states::boundary}
\end{subfigure}
\hfill
\addtocounter{subfigure}{1}
\begin{subfigure}{\linewidth}
\includegraphics[width=\linewidth]{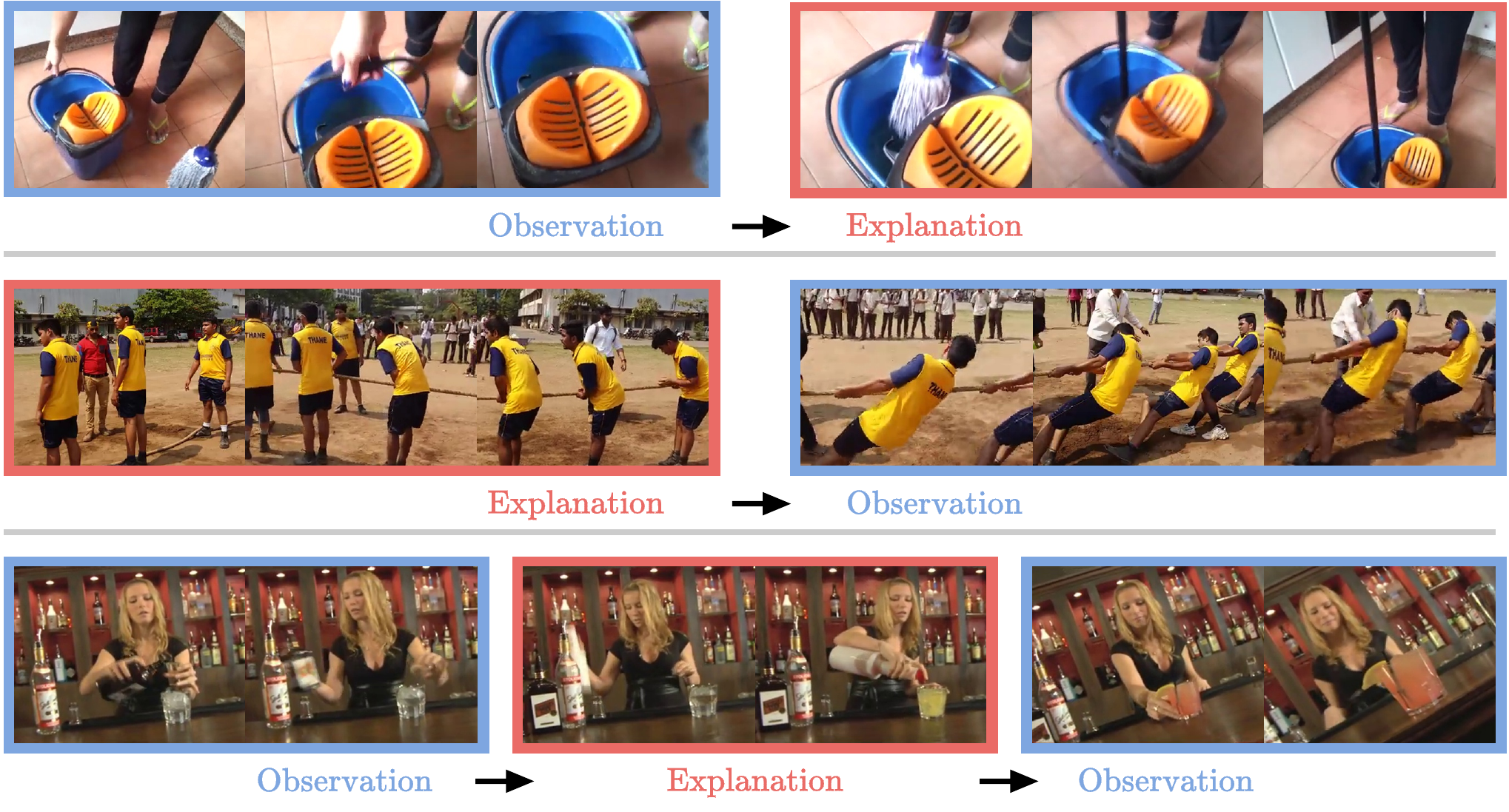}
\caption{\textbf{Visual Abductive Reasoning (VAR)}. Given the observable part of the video in \textcolor{babyblue}{blue}, infer a likely explanation in \textcolor{fadedred}{red} for what follows before, after, or during the observation. The task requires a high-level understanding of the action or activity performed. Video sourced from \tcite{liang2022visual}. \vspace{1em}}
\label{fig:states::reasoning}
\end{subfigure}
\end{minipage}
\hfill
\begin{minipage}{0.49\textwidth} 
\addtocounter{subfigure}{-1} 
\begin{subfigure}{\linewidth}
\includegraphics[width=\linewidth]{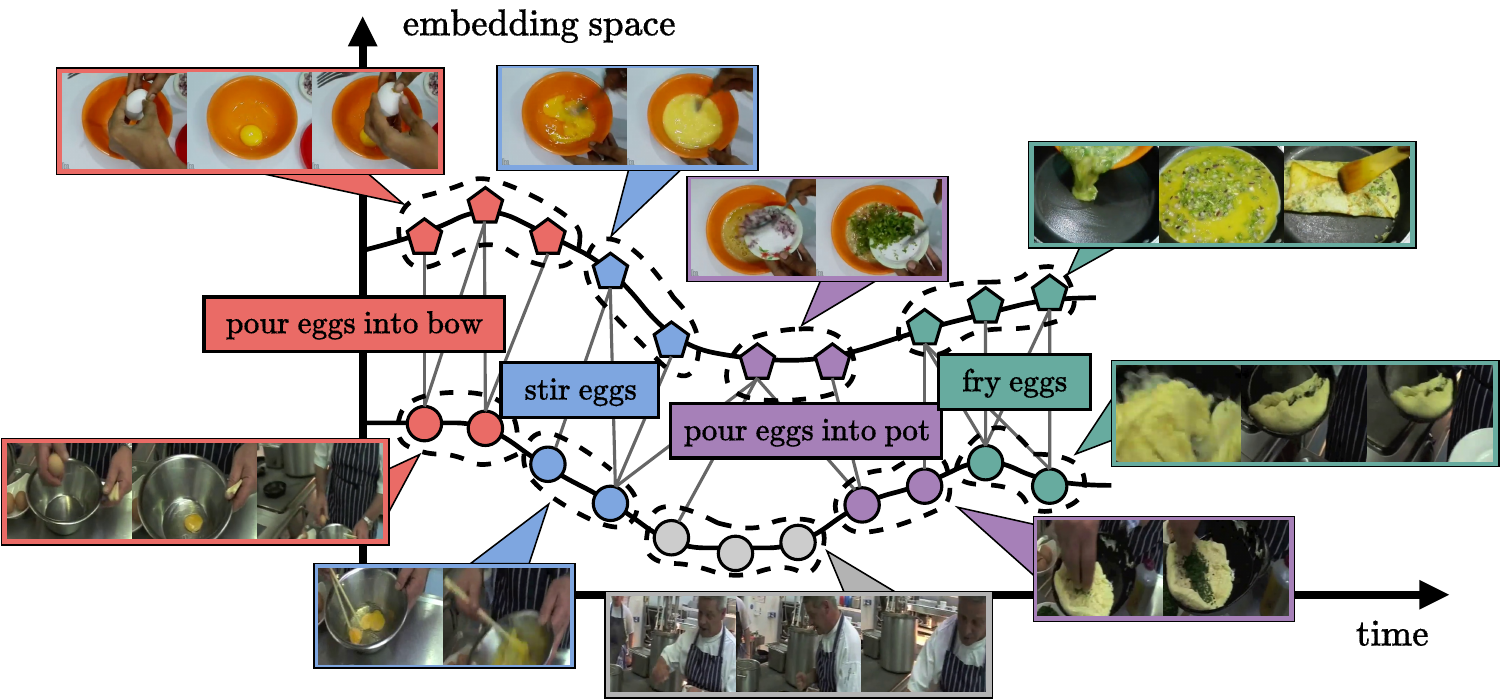}
\caption{\textbf{Video Alignment (VA)}. Find correspondences across video instances with the same action performed and align them such that the execution of the action is synchronized. Video sourced from \tcite{tang2019coin}. \vspace{.5em}}
\label{fig:states::align}
\end{subfigure}
\hfill
\begin{subfigure}{\linewidth}
\includegraphics[width=\linewidth]{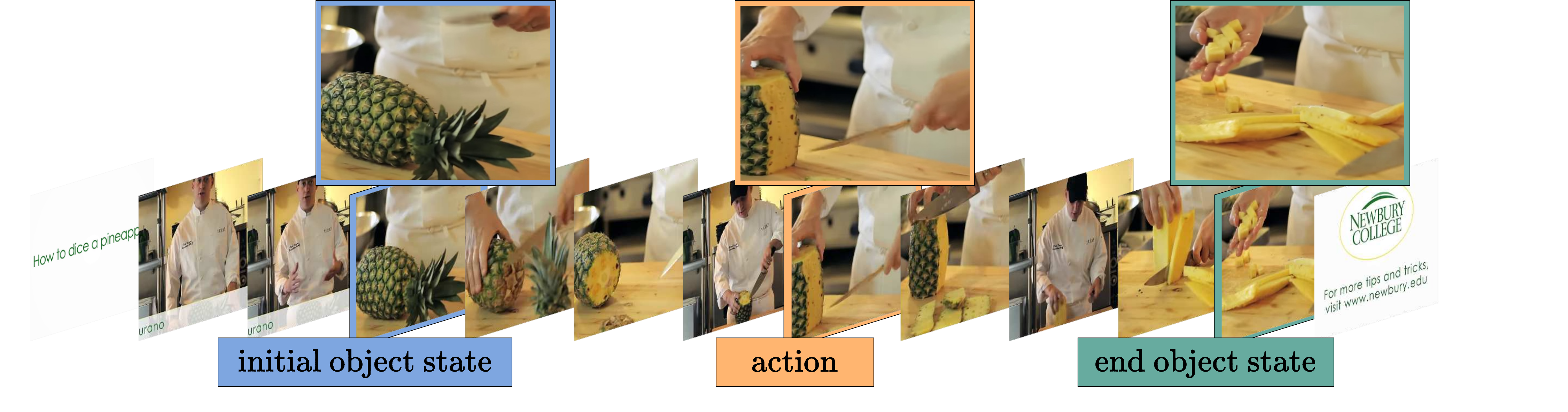}
\caption{\textbf{Object State Change Detection (OSCD)}. State modifying actions such as \textit{cutting} progressively change the visual appearance of objects from an initial state in \textcolor{babyblue}{blue} to a final post-action execution state in \textcolor{pastelteal}{teal}. OSCD identifies the times that these changes occur. Video sourced from \tcite{souvcek2022look}. \vspace{0.5em}}
\label{fig:states::oscd}
\end{subfigure}
\end{minipage} 
\begin{subfigure}{\linewidth}
\includegraphics[width=\linewidth]{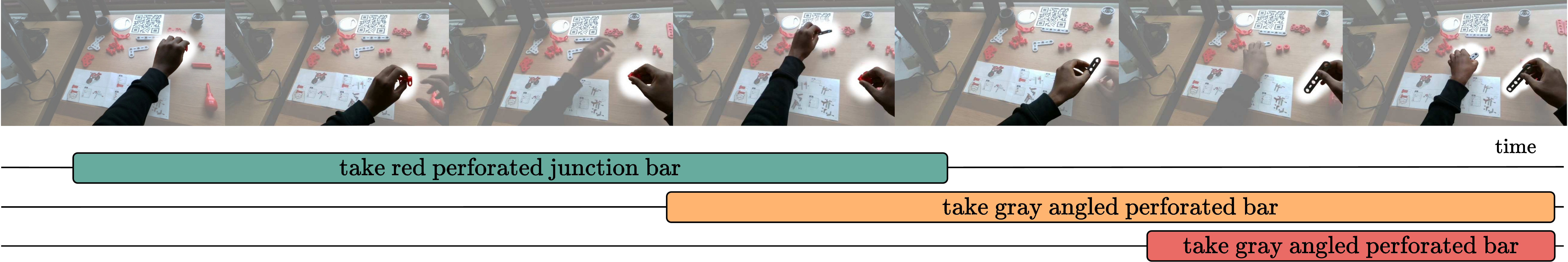}
\caption{\textbf{Active Object Detection (AOD)}. Given a video in which a person interacts with multiple objects, AOD detects the object the person is currently using. Video sourced from \pcite{ragusa2021meccano}.}
\label{fig:states::aod}
\end{subfigure}
\caption{\textbf{Tasks relating to object and action state change}. Each of the presented tasks can involve additional objectives.}
\label{fig:state_changes}
\end{figure*}

\subsection{State changes}
\label{sec:prediction::states}

Another set of action prediction tasks includes modeling the state changes in the environment, actions, objects, or execution speeds. Tracking, inferring, and reasoning in these tasks come with new sets of changes. An overview of the tasks' objectives is visualized in~\Cref{fig:state_changes}.

\subsubsection{Challenges}
\label{sec:prediction::states::challenges}

Understanding the sequentiality in videos is a central part of human perception. The translation of this to computer vision tasks remains challenging \pcite{de2023there}. A central challenge relates to the \textbf{object state variability} as actions affect the visual appearance of objects in terms of shape, visibility, or perspective. Another set of challenges concerns actions with \textbf{non-rigid temporal boundaries}. Actions may not be easily distinguished from backgrounds or their execution may overlap with other actions. This presents \textbf{ambiguities in the action progress} as the completion of an action, or part of it, might also involve another action. Finally, the typically weak relation between visual input and the high-level semantic interpretation thereof complicates the training of robust, general models.

\subsubsection{State-based tasks}
\label{sec:prediction::states:::state_tasks}

We discuss six main tasks based on object and action state changes in their objectives. These include tracking the progress of actions (\textbf{action progress prediction}), defining start-end times (\textbf{action progress prediction}), temporally aligning action phases (\textbf{video alignment}), inferring action goals (\textbf{visual abductive reasoning}), tracking object appearance changes (\textbf{object state change detection}), and localizing relevant objects (\textbf{active object detection}).

\noindent
\textbf{Action Progress Prediction (APP)}. Actions can be understood by procedural sets of motions performed towards an intended goal as shown in \Cref{fig:states::progress}. \tcite{vaina1991object} suggested that understanding the state and progress of the action at different times can provide a holistic understanding of the intent and objective. In machine vision, an initial APP approach \pcite{fathi2013modeling} used local descriptions to model per-frame state changes. \pcite{kataoka2016recognition} used a descriptor to discover transitional actions within activity sequences. \tcite{xiong2017pursuit} introduced a score function to distinguish between actions based on learned distinctive parts. \tcite{becattini2020done} used actor and scene context information as an additional supervisory signal for APP. \tcite{price2022unweavenet} expressed the progress of multiple actions through threads of activities that can overlap, a common situation in long procedural videos. \tcite{shen2024progress} causally attended videos to define a task graph for APP over each action. More recently, generative approaches \pcite{damen2024genhowto} using conditional control \pcite{zhang2023adding} and procedural knowledge \pcite{ashutosh2023video,zhou2023procedure} have been used to generate keyframes of changes. 

Another line of research works \pcite{heidarivincheh2016beyond,heidarivincheh2018action} is aimed at localizing the moments that actions are completed. The speed of action completion or state changes in actions has also been studied in the context of skill determination \pcite{doughty2018s} or their semantic correspondence to textual adverbs \pcite{doughty2020action,doughty2022you,moltisanti2023learning}. Scoring approaches \pcite{tang2020uncertainty} have been used to study the procedural execution of actions in the context of quality assessment. Adjacent tasks such as video captioning and action classification have also been integrated into multi-task settings \pcite{parmar2019and}.

\noindent
\textbf{Event Boundary Detection (EBD)}. Different from the related well-studied task of action localization, EBD \pcite{shou2021generic} localizes event changes in videos \emph{regardless of the action classes}, shown in \Cref{fig:states::boundary}. \tcite{aakur2019perceptual} proposed a self-supervised objective in which their model is initially trained to reconstruct subsequently observed features. \pcite{shou2021generic} used a self-similarity metric to determine event boundaries by relating encoded frame features. Further, EBD approaches \pcite{mounir2023streamer} have studied hierarchies of video events. Recently, \tcite{eyzaguirre2024streaming}  
explored the detection of event starts from natural language queries. 

\noindent
\textbf{Video Alignment (VA)}. As the performance of individual parts of actions can vary, video alignment, shown in \Cref{fig:states::align}, aims to temporally match key moments in the execution of the same action across videos. Initial efforts, motivated by temporal coherence \pcite{goroshin2015unsupervised,fernando2017self,zhang2023modeling}, have studied VA based on Canonical Correlation Analysis (CCA) \pcite{andrew2013deep} or by contrastively creating joint representations from multiple viewpoints \pcite{sermanet2018time}. Dynamic Time Warping \pcite{sakoe1978dynamic} is an algorithm that aligns variable length signals, and it has been adopted for VA \pcite{chang2019d3tw,dvornik2021drop,hadji2021representation}. A more recent self-supervision objective \pcite{dwibedi2018temporal} for VA is to train a video model to project per-frame embeddings in pairs of target videos by matching embeddings of one video to the nearest neighbor embeddings of the other. This approach was extended in subsequent works with the inclusion of context from the entire video \pcite{haresh2021learning}, anchor frames to align redundant frames \pcite{liu2022learning}, embeddings from text \pcite{epstein2021learning}, and regularizers based on the correspondence to repetitions of the same action \pcite{donahue2024learning}.

\noindent
\textbf{Visual Abductive Reasoning (VAR)}. A key element in action understanding is recovering the intended goal. High-level reasoning of events has initially been considered in hierarchies with rule-based approaches \pcite{hakeem2004ontology} or the conditionality of action occurrences across different levels \pcite{albanese2010pads}. \tcite{pei2011parsing} detected atomic actions with graph representations to decompose complex events. VAR \pcite{liang2022visual}, shown in \Cref{fig:states::reasoning}, is the vision-language task that uses characteristics of partial observations as a premise and requires formulating an explanation. Other VAR works have modeled intention by contrastively learning visual and language context \pcite{li2023intentqa}, modeling timelines for news story understanding \pcite{liu2023video}, and forecasting actions by multimodal inputs \pcite{zhu2023personality}. Evaluation of VAR models has also been studied in counterfactual vision-language pairs \pcite{park2022exposing} similar to text-only tasks \pcite{ippolito2019unsupervised,huang2020inset}.

\noindent
\textbf{Object State Change Detection (OSCD)}. Many actions alter the appearance or state of objects. OSCD approaches associate visual changes to changes in the states of objects in the scene, as shown in \Cref{fig:states::oscd}. Efforts \pcite{alayrac2017joint,damen2014you,liu2017jointly,zhuo2019explainable} have initially focused on state modifications that do not involve significant appearance changes, \eg, \texttt{open/close door} or \texttt{fill/empty cup}. \tcite{hong2021transformation} proposed a reasoning-based approach defining a triplet of complexities for single- and multi-step transformations with additional viewpoint changes. Other reasoning-based approaches include the use of language \pcite{xue2024learning} and visual exemplars of start and end states \pcite{souvcek2022look}. OSCD has also been studied in combination with other tasks including cross-state object segmentation \pcite{yu2023video}, cross-action relevance \pcite{alayrac2024multi}, or inspired by state-disentanglement for images \pcite{gouidis2023leveraging,nagarajan2018attributes,saini2022disentangling}, generating start and end states by given context and scene prompts \pcite{damen2024genhowto,saini2023chop}.

\noindent
\textbf{Active Object Detection (AOD)}. Actions can include multiple objects during their execution. Overviewed in \Cref{fig:states::aod}, AOD localizes the objects relevant to the currently performed atomic action with bounding boxes. This task has recently gained interest as scenes can often be cluttered \pcite{ragusa2021meccano} or a varying number of objects can be used for a single action \pcite{miech2019howto100m}. \tcite{nagarajan2019grounded} specifically focused on localizing the human-object interaction areas defining focal points of importance during the execution of actions. \pcite{fu2021sequential} introduced a voting module over potential bounding boxes corresponding to the active object. \tcite{kim2021hotr} used a parallelized model to detect instances and subsequently hand-object interactions. \tcite{yang2024active} used scene context from text to define plausible interactions with target objects for AOD.

\subsubsection{Future outlooks}
\label{sec:prediction::states:::outlooks}

Similar to recent works in robotics that create goal-based policies \pcite{kununi2024uni,wang2023manipulate}, state-understanding vision models require a holistic understanding of actions given a limited \textbf{availability of arbitrary object states}. Conceptually, the execution and changes in objects can be similar for different actions, \eg, \texttt{mixing a cake mix} and \texttt{whisking eggs}. Enforcing better learning objectives in order to improve this semantic correspondence can create more generalizable models with a better understanding of the physical world. In addition, the inclusion of cues from \textbf{supplementary modalities}, such as audio, can improve tasks that require relating parts of videos, as correspondences should be discoverable beyond the visual domain. This also presents the potential for creating general-purpose models based on \textbf{abductive reasoning} pretext objectives that can then be applied to various downstream tasks.

\subsection{Anomaly detection}
\label{sec:prediction::VAD}

Video Anomaly Detection (VAD) is the task of detecting unexpected actions or events in videos that deviate from predictable behaviors. Anomalies are detected through either explicitly classifying a pre-set number of actions in \textbf{close-set} settings or learning robust representations of the expected actions in \textbf{open-set} settings.

\subsubsection{Challenges}
\label{sec:prediction::VAD:::challenges}

VAD depends on strict binary annotations of normal and abnormal sequences, with most evaluation benchmarks including \textbf{limited definitions}. Even in the open-set settings, robust definitions are required for the target normal sequences. This prevents the creation of models that can estimate correctly unseen normal sequences based on their visual proximity to other actions. Although recent-continual-learning approaches have also been proposed for VAD \pcite{bugarin2024unveiling}, their applicability remains sparse. Another significant challenge for VAD models relates to their applicability. With their intended use in continuously operating surveillance systems, current approaches only \textbf{partially use temporal context} to infer predictions. Only a small number of approaches currently study the long-term effects of actions. Most datasets also only include modest temporal resolutions, limiting the exploration of context over longer time segments.

\subsubsection{Detecting anomalies in videos}
\label{sec:prediction::VAD:::methods}

We identify two main approaches for the detection of anomalies in videos.

\noindent
\textbf{Close-set}. Anomalies can be discovered by \emph{close-set} tasks that aim to model both normal and abnormal sequences. \tcite{sultani2018real} used Multiple Instance Ranking \pcite{dietterich1997solving} to define positive groups that include videos with at least a single abnormal segment and negative groups of normal videos. The objective is to maximize the score between the assigned positive and negative groups. Subsequent efforts have built upon MIL with learned features \pcite{dubey20193d}, or pseudo labels \pcite{feng2021mist}. Approaches have also aimed to improve upon MIL's reliance on the dominant negative instances. \tcite{zhang2019temporal} integrated inner-group sampling, \tcite{pu2023learning,zhu2019motion} used temporal weighting, and \tcite{tian2021weakly} maximized the separability between normal and anomalous representations. With a similar goal, the use of multiple temporal pretext tasks \pcite{almarri2024multi,georgescu2021anomaly} and temporal scales \pcite{li2022scale} have also been explored. \tcite{chen2023mgfn} used a contrastive objective between representations of normal and abnormal videos. Clustering approaches have focused on modeling sparsity \pcite{lu2013abnormal}, enforced high distribution variance between normal and abnormal video representations \pcite{li2021deep}, combed dense/spare clusters for normal/abnormal segments \pcite{zaheer2020claws}, and used pseudo labels for anomalous segments \pcite{zaheer2020self}. Another set of methods \pcite{zhong2019graph,purwanto2021dance} has included graph networks to sequentially detect abnormal segments. More recent methods have distinguished between normal and anomalous states with the use of additional modalities such as audio \pcite{wu2020not} and language context \pcite{yang2024text,zanella2024harnessing}.

\begin{figure*}[!ht]
    \centering
    \includegraphics[width=\linewidth,trim={0cm 0 0cm 0},clip]{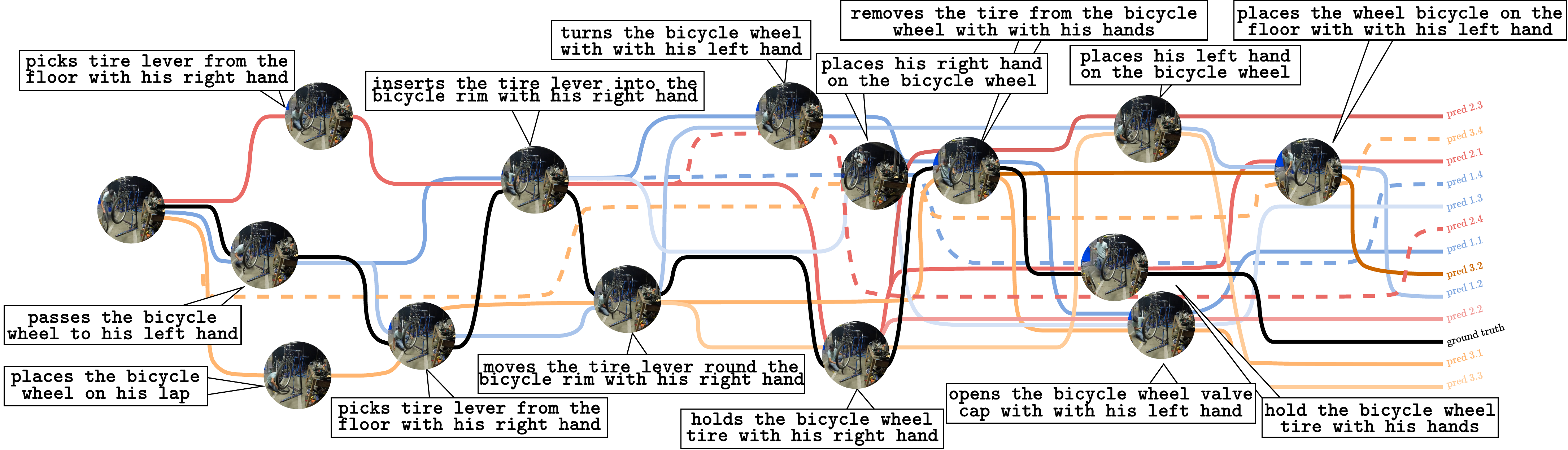}
    \caption{\textbf{Forecasting future actions}. Starting from the observed action, anticipation approaches infer the sequence of probable next actions. Predictions are shown in a narrative chart format similar to \tcite{randal2009movie}.
    Example selected from \tcite{grauman2024ego}.} 
    \label{fig:narration_chart}
\end{figure*}

\noindent
\textbf{Open-set}. As close-set solutions can only model abnormalities in labeled data, models cannot effectively generalize to distributions different than those seen during training. This issue has been studied by \tcite{zhao2011online} and \tcite{luo2017revisit} as a sparse-coding \pcite{lee2006efficient} problem in which the model is trained to reconstruct only plausible normal sequences. Abnormalities are then inferred by large reconstruction loss offsets. Temporal regularity can also be modeled with autoencoders as a reconstruction task \pcite{hasan2016learning}. To deal with the scarcity of abnormal sequences during training, a number of autoencoder (AE)-based approaches use pseudo representations to improve the embedding space \pcite{astrid2021synthetic,astrid2021learning}. \tcite{park2020learning} learned prototypes of normal sequences which can then be used to contrast query videos. Other works have constrained the representation space of normal sequences by optimizing piecewise linear decision boundaries \pcite{wang2019gods}. Two-steam AE frameworks \pcite{cho2022unsupervised,nguyen2019anomaly} have been used to separately reconstruct the appearance and motion characteristics of normal sequences. Generative approaches \pcite{micorek2024mulde} have recently focused on the latent space with Gaussian Mixture Models (GMM) and inferred an anomaly score across all noise levels. \tcite{fioresi2023ted} has explored cross-frame mutual information minimization in tandem with a generative objective for privacy preservation. Other privacy-aimed approaches have studied trajectory-based video anomaly detection. \tcite{morais2019learning} used an RNN to track skeletal points and regressed future locations. The objective of the model was to learn a fixed interpolation for normal sequences with abnormalities captured by the large reconstruction error. Subsequent approaches have used a similar objective to train graph networks \pcite{markovitz2020graph}, probabilistic models \pcite{flaborea2023multimodal}, and masked autoencoders \pcite{stergiou2024holistic}.

\subsubsection{Future outlooks}
\label{sec:prediction::VAD:::outlooks}

Although great progress has been made, most VAD methods still require adjusting the definitions of normal sequences to update predictions. A direction for future works can be a unified framework that can efficiently adapt predictions through \textbf{life-long learning}. Recent works \pcite{yang2024text,zanella2024harnessing} fused frame-wise language descriptions from LLMs without requiring re-training the visual model. This can be an effective strategy for upcoming VAD works. Most VAD approaches are also based on the accuracy of the produced summaries with video-language misalignment directly impacting performance. Potential improvements may explore \textbf{in-context learning}. \tcite{zhao2024can} showed that progressively increasing task difficulty through prompts can aid the generalization ability of models to unseen visual scenes.

\section{Future forecasting}
\label{sec:forecasting}

The future is often uncertain. As shown in \Cref{fig:narration_chart}, a sequence of steps can lead to multiple possible scenarios. Anticipation models are trained on objectives that require the discovery of domain-specific knowledge to address future anticipation challenges. We discuss methods for anticipating categorical semantics in \Cref{sec:forecasting::anticipation}. We then explore approaches for generating future actions in videos in \Cref{sec:forecasting::generation}.

\subsection{Action anticipation}
\label{sec:forecasting::anticipation}

Action Anticipation (AA) uses current action(s) performed at $\tau_1$ to forecast \emph{proceeding actions} at $\tau_2$. In contrast to the partial observations for EAP, anticipation tasks only rely on the expected sequence with which actions can be performed. Early works \pcite{kitani2012activity,kuehne2014language,koppula2015anticipating} have used graphs to model the sequential nature of actions over time. However, to address the long-range dependency limitations of graph-based approaches, works have focused on the procedural execution of actions \pcite{abu2018will,furnari2019would,ke2019time}, the motion transition intensity between actions \pcite{huang2014action}, as well as gaze and hand information \pcite{shen2018egocentric}, while defining future-action objectives with multiple predictions \pcite{furnari2018leveraging,zatsarynna2024gated}. Despite the diversity in approaches, some challenges remain.

\subsubsection{Challenges}
\label{sec:forecasting::anticipation:::challenges}

Anticipation models predict future actions in sequences. Thus, \textbf{future action predictions are accumulated} across multiple rounds. As the future may be unpredictable, errors in these predictions will also influence and reduce the quality of subsequent predictions as the sequence's length increases. Although some models forecast the entire sequence \pcite{gong2022future,nawhal2022rethinking}, their temporal context is limited compared to autoregressive approaches.  

Current works rely on \textbf{fixed anticipation time intervals} in which the intermission time $\tau_{1 \rightarrow 2}$ remains constant in training and inference. This significantly limits the applicability of methods in real-world scenarios in which the duration of intervals varies, requiring models to adjust predictions based on conditions such as the speed of execution, difficulty of the action, or the actor's expertise. The majority of existing methods are bound to re-training to accommodate such characteristics.

\subsubsection{Anticipation approaches}
\label{sec:forecasting::anticipation:::methods}

We discuss three action- and object-based future anticipation approaches.

\noindent
\textbf{Embedding similarity maximization}. Representations of future actions can be used as targets for learned embeddings. A large number of methods have thus used future embedding reconstruction tasks to infer future action labels. \tcite{gao2017red} used a recurrent decoder to regress future embeddings with an additional policy for class predictions over time. Interactions between objects and actors \pcite{sun2019relational,luc2018predicting} have been explored in early works. Subsequent methods aimed to either maximize the similarity between future and current embeddings through memory banks \pcite{liu2022hybrid}, optimize latent representations for intended goals \pcite{roy2022action}, learn prototypes \pcite{diko2024semantically}, or use adversarial representations \pcite{gammulle2019predicting}. Other generative approaches use pose information as priors \pcite{villegas2017learning} or focus on the extrapolation of activity trajectories \pcite{chi2023adamsformer}. Autoregressive approaches have recently shown great promise using either contrastive objectives \pcite{wu2020learning}, causal attention \pcite{girdhar2021anticipative}, or audio-visual inputs \pcite{zhong2023anticipative}. As future predictions depend on the usefulness of current observations, works have also integrated uncertainty terms in their predictions. \tcite{vondrick2016anticipating} regressed towards multiple plausible future embeddings, \tcite{abdelsalam2023gepsan} grounded the sequentiality of visual embeddings to language, while \tcite{guo2024uncertainty} defined probabilistic transformer outputs through a top-k prediction loss similar to \tcite{furnari2018leveraging}.

\noindent
\textbf{Long-term anticipation}. The anticipation of the future can also be extended to forecasting multiple upcoming actions over a longer temporal duration. \tcite{bokhari2017long} used a q-learning framework with reward functions for the activity label, and locations where actions are performed. \tcite{nawhal2022rethinking} used a two-stage approach to first infer potential labels and then utilize their logits alongside visual features to predict future action segments. Similarly, \tcite{gong2022future} used learnable latent representations for the future embeddings and cross-attended \pcite{jaegle2021perceiver,lee2019set} them with the observed video embeddings. Generative approaches have also learned future embeddings from pre-defined temporal states \pcite{piergiovanni2020adversarial}, logit sequences \pcite{zhao2020diverse}, cyclic consistency \pcite{abu2021long}, or the expected variance in future representations \pcite{mascaro2023intention,patsch2024long}. Recently, \tcite{mittal2024can} used general language and visual queries to infer prediction through LLMs.

\noindent
\textbf{Next active object}. A recently introduced set of anticipation tasks concerns the study of object-centric future forecasting. Next active object anticipation forecast the objects that will be used in future actions and has been addressed using predictions on the salient regions \pcite{dessalene2021forecasting}, hand position generated representations \pcite{jiang2021predicting}, or autoregressively attending object and visual information \pcite{thakur2024anticipating}. Other methods may also forecast human-object interaction regions \pcite{liu2020forecasting,liu2022joint,roy2024interaction}, object relations \pcite{roy2021action,zatsarynna2021multi}, or time-to-contact estimates \pcite{mur2024aff}.

\subsubsection{Future outlooks}
\label{sec:forecasting::anticipation:::outlooks}

Despite the great advancements and number of methods for each anticipation task, less-studied aspects that enhance the applicability of current models exist. Primarily, although the intention of these approaches is their \textbf{deployment in real-time scenarios}, they are still used and evaluated in offline settings. Aspects such as latency and inference times for these models are largely overlooked with only a limited number of works designing stream-based models \pcite{furnari2022towards,girase2023latency}. The application of these models in real-world settings also requires \textbf{prediction adjustability} as inferring labels or sentences of a specific semantic hierarchy may not always be possible. Instead, the prediction granularity needs to be both adjusted and subsequently refined given the available video information. A greater research challenge concerns \textbf{multi-person anticipation} in which multiple atomic and group actions need to be forecasted. This also includes forecasting actions in social scenarios in which human-human interactions also need to be predicted, possibly with regard to the social context such as interpersonal relations and roles.

\subsection{Video Generation}
\label{sec:forecasting::generation}

Forecasting future actions can be extended from the semantic space to the pixel space. Generating real-world physical phenomena includes a high level of complexity \pcite{finn2016unsupervised}. Capturing simple state transitions of pixels at times does not suffice in learning complex spatiotemporal variations of motions \pcite{wu2021motionrnn}. Several types of models have been used throughout the years to address complex scene dynamics.

\begin{figure*}[ht]
    \centering
    \begin{overpic}[width=\linewidth]{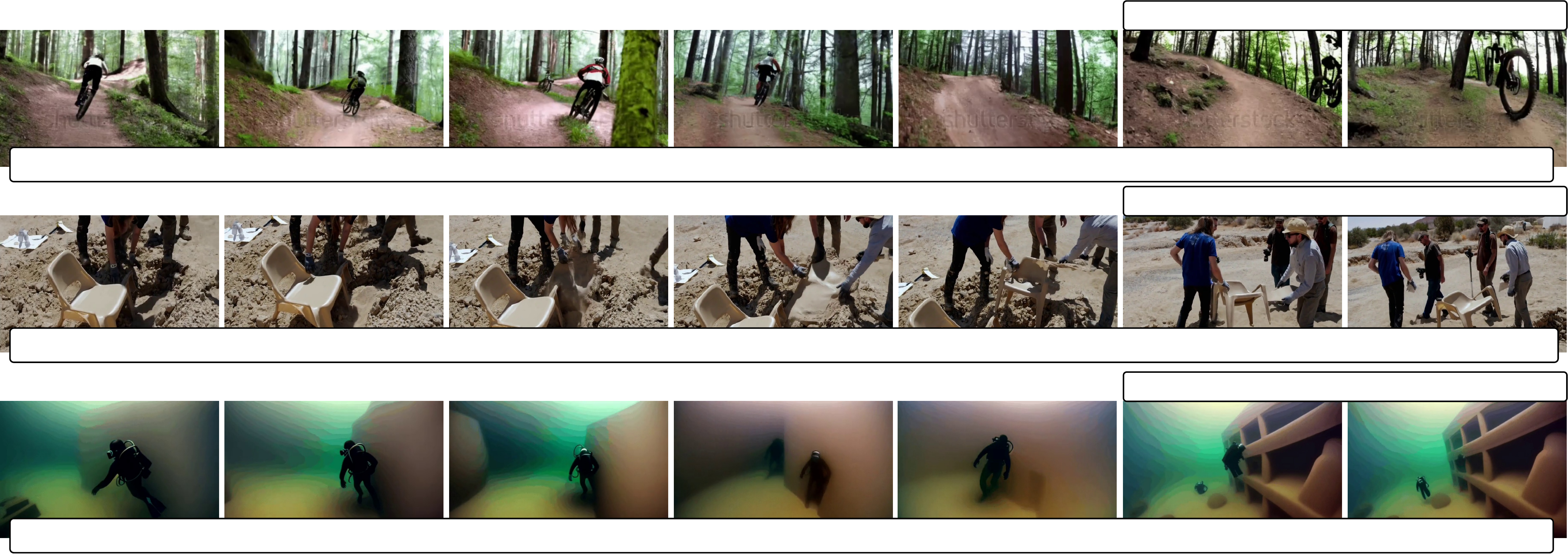}
    \put(72,34){\texttt{Consistency failure}}
    \put(72,22){\texttt{Physics failure}}
    \put(72,10.3){\texttt{Video-prompt alignment failure}}

    \put(2,24.3){FIFO  \pcite{kim2024fifo} \texttt{An exciting mountain bike trail ride through a forest}}
    \put(2,12.8){SORA \pcite{videoworldsimulators2024} \texttt{Archelogists discover a plastic chair in the desert, excavating and dusting it}}
    \put(2,0.7){Gen-L-Video \pcite{wang2023gen} \texttt{A vibrant underwater scene of a scuba diver exploring a shipwreck}}
    \end{overpic}
    \caption{\textbf{Video generation challenges}. Failure cases in video generation can be attributed to (a) poor continuity between frames with appearance or motion changes that do not correspond to the intended concept, (b) failure to capture real-world physics, and (c) poor video and prompt alignment, causing a mismatch between the generated scene and the given description.}
    \label{fig:dlm_fails}
\end{figure*}

\subsubsection{Challenges}
\label{sec:forecasting::generation:::challeneges}

Generating representative future actions is challenging. We visualize three prevalent challenges in video generation in \Cref{fig:dlm_fails}.

\noindent
\textbf{Consistency failures}. Similar to forecasting the semantics of later actions, generative approaches can accumulate errors resulting in a degradation of the frame quality as the video progresses. Methods that either use memory banks \pcite{oshima2024ssm}, coarse to fine representations \pcite{yin2023nuwa}, or generate long sequences in parallel \pcite{zhuang2024vlogger} either require long inference times or are computationally expensive.

\noindent
\textbf{Physics failures}. As scene dynamics and characteristics of the real world are learned implicitly by generative models, out-of-distribution actions or motions may not be effectively synthesized. Most of the metrics and objectives used to quantify model performance are based only on the visual quality of the generated video. However, the scene's realism also extends to feasible motions, actions, and permutations.

\noindent
\textbf{Video-prompt alignment failures}. Conditional generation is a recent research topic of high interest. Text-to-Video (T2V) synthesis relies on binding LLM and noise latent embeddings to control video generation. Misalignments in the joint embedding space will be reflected in the outputs.

\subsubsection{Methods}
\label{sec:forecasting::generation:::methods}

We discuss groups of generative video models next.

\noindent
\textbf{Stochastic models}. This group explores future generation by encoding a variance latent. Variational Autoencoders (VAE) \pcite{kingma2013auto} have been used to stochastically generate trajectories \pcite{walker2016uncertain} and forecast motions representations \pcite{fragkiadaki2017motion}. In the video domain,  \tcite{babaeizadeh2018stochastic} used the generator network from \tcite{finn2016unsupervised} alongside a probabilistically sampled latent to condition next frame generation by the variance of the previous frame. To improve the learned distribution of the generated outcomes, \tcite{denton2018stochastic} used the KL-divergence between generated and previous frame latents to regularize the frame generation and avoid optimization shortcuts that copy previous frames. \tcite{yan2018mt} learned differences between two adjacent views to improve the sampling distribution. Other works aimed to improve the learned variance by the differences between two adjacent views \pcite{franceschi2020stochastic} or through adaptive regularization of the reconstruction objective \pcite{chatterjee2021hierarchical}. Several methods have also employed a hierarchical latent to learn multiple levels of features \pcite{castrejon2019improved,kumar2020videoflow,saxena2021clockwork}. 

\begin{figure}[t]
    \includegraphics[width=\linewidth]{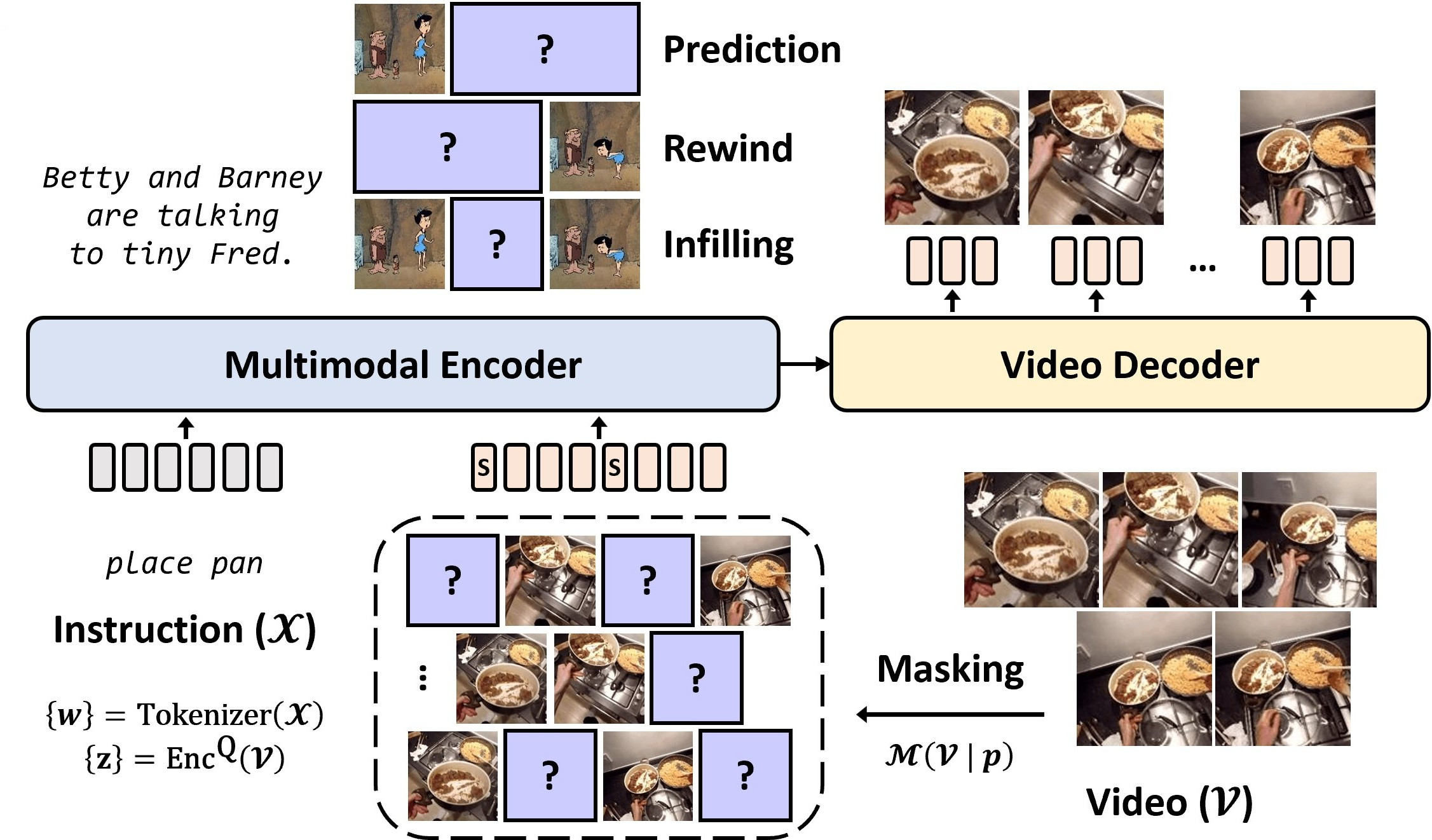}
    \caption{\textbf{Multi-task video generation} model from \tcite{fu2023tell}. Based on a partial video and a text prompt used to condition a codebook, a text-condition video VQGAN generates missing frames.}
    \label{fig:mmvg}
\end{figure}

\noindent
\textbf{Adversarially generated short video sequences}. Generative Adversarial Networks (GANs)~\pcite{goodfellow2014generative} are based on an orthogonal objective between a generator network that synthesizes inputs from latent representations and a discriminator network optimized to distinguish between real and generated inputs. Video approaches extended the dimensionality of convolutional and deconvolutional kernels to space and time. An initial effort by \tcite{vondrick2016generating} was to learn scene dynamics from unlabeled videos with a dual objective of generating static backgrounds and moving foregrounds. \tcite{saito2017temporal} used a similar dual objective by first generating temporal adjacent views of latents and a spatial generator for frames. Works have also incorporated stochastic latent embeddings that can be decoded to the entire video \pcite{lee2018stochastic}, separate spatial and temporal discriminators \pcite{clark2019adversarial}, recurrent units \pcite{gupta2022rv,wang2023styleinv}, and spatio-temporal kernel transformations \pcite{luc2020transformation}. \tcite{menapace2021playable} created an autoregressive approach to generate frames by conditioning the generation with discrete action labels. Other methods have adapted image-based GANs by generating latent trajectories of frame features \pcite{tian2021good} or shifting frame features across time \pcite{munoz2021temporal}. \tcite{yu2022generating} used spatiotemporal coordinate information from latent representations of motion and video diversity. \tcite{fu2023tell} used a variational-based GAN \pcite{esser2021taming} to synthesize past or future frames (temporal outpainting) and current frames (inpainting) of videos based on both cues and textual descriptions. The model's pipeline is shown in \Cref{fig:mmvg}. A partial video is used alongside a codebook of latents to generate the remaining frames. Stop gradients are used to contrastively update the encoder and codebook. A final frame-wise feature-matching function is used to improve the embedding distance of real and generated frames.

Despite the progress in video generation by adversarial models, instability in training and mode collapse are the primary disadvantages of GANs when generating realistic and diverse videos.

\noindent
\textbf{Probabilistic models for video generation}. Denoising Diffusion Probabilistic Models (DDPMs) \pcite{ho2020denoising,sohl2015deep,song2019generative} combine two Markov processes with the first (forward) corrupting the input data to noise within a distribution. The second (backward) process reverses this effect by reconstructing an input from the noisy representation. New inputs not in the training data are generated by sampling the prior distribution. \tcite{ho2022video} directly extended this formulation to video by extending the original U-Net's \pcite{salimans2017pixelcnn++} dimensionality used in the backward step with space-time kernels. Following works \pcite{he2022latent,hong2022cogvideo,blattmann2023align} have moved away from pixel-level diffusion. They instead utilize the semantically rich and lower-dimensional latent space \pcite{rombach2022high} with autoencoders to encode and project video inputs and outputs. The computational efficiency of Latent Diffusion Models (LDM) has enabled a new stream of works to improve temporal alignment of frames \pcite{blattmann2023align,yang2023video} and minimize training data requirements \pcite{nikankin2023sinfusion,wu2023tune}. \tcite{yu2023avideo} combined the two approaches by projecting videos to triplane representations. \tcite{yu2024efficient} adapted image-based models by incorporating low-resolution temporal content latents computed as the weighted sum of frames. Both image-based and low-resolution motion-based models are denoised with a similar training objective conditioned on the context vector. Although video context can improve frame generation, the number of frames these models can generate is still limited.

\noindent
\textbf{Text-conditioned generation}. Language embeddings are increasingly used as a prior for video generation. The objective of these methods is to generate videos from textual descriptions based on visual-language correspondence. \tcite{dorkenwald2021stochastic} used the start and end times as generation-controlling factors. 
Following methods explored T2V generation based on VQVAE codebooks conditioned on language \pcite{han2022show,yan2021videogpt}, or language and motion \pcite{hu2022make}. As LDM approaches rely on latent representations, unified visual-language embedding spaces can also be used to generate videos. LDM methods include joint conditional generation of images and videos \pcite{gupta2023photorealistic}, shifting latent features for parameter-free temporal variance \pcite{an2023latent}, and concatenating frames in spatial grids \pcite{lee2024grid}. \tcite{zeng2024make} showed that
first generating the start and end action states enables models to effectively generate the transitioning frames limiting the dependence on well-formed textual descriptions. A number of recent works \pcite{fei2024dysen,tian2024videotetris,wang2023videocomposer,wang2024recipe,wei2024dreamvideo,zhuang2024vlogger} have employed adapters on image-based LDMs similar to ControlNet \pcite{zhang2023adding}. Given a pre-trained model using input latents and frozen parameters, a copy of the block is created with trainable parameters to fuse conditional latents of any modality type. These are integrated into the frozen model with projection or cross-attention layers.

\noindent
\textbf{Generating long sequences}. Generating long videos is challenging as it requires models to learn long-range temporal dynamics. Initial efforts aimed to mitigate reductions in the generation quality over time through hybrid training schemes \pcite{brooks2022generating} and by concatenating frame-wise codecs over time for frame consistency in generation \pcite{skorokhodov2022stylegan}. \tcite{shen2023mostgan} extended these approaches by using learnable latent vectors to represent motion styles as priors to generate frames. \tcite{harvey2022flexible} explored the conditionality between sampled frames for generating 25min videos with fixed backgrounds. Approaches \pcite{ho2022imagen,singer2023make} have also focused on super-resolution models in tandem with LDMs to generate low-resolution long video sequences which are upsampled to higher resolutions in subsequent steps. Several models have been based on autoregression to accommodate future unpredictability and large video changes. \tcite{weissenborn2020scaling} extended the patch-based generation approach of Subscale Pixel Networks (SPNs)~\pcite{menick2019generating} to spatio-temporal voxels. \tcite{ge2022long} used an autoregressive transformer to generate latent representations for the next frames with a VQVAE as a backbone generator. Other auto-regressive VQVAE-based works explored dimension-specific \pcite{wu2021godiva} and local attention \pcite{liang2022nuwa,wu2022nuwa}. More recently, video generation methods have adapted causal attention encodings fused with previous frame features \pcite{yan2023temporally,villegas2022phenaki}, used cross-attention adapters to include temporal context in image generators \pcite{long2024videodrafter}, and guided the generation with foreground masks \pcite{chang2024look}.

\subsubsection{Future outlooks}
\label{sec:forecasting::generation:::outlooks}

Despite the recent substantial advancements of generative models, rudimentary challenges still exist that, in turn, provide opportunities for future approaches. Crucially, despite the high appearance quality of current models, a standardized \textbf{evaluation and benchmark method} is missing. Approaches based on T2V generation have been shown to replace or fuse concepts, and generate irrelevant objects using specific low-confidence prompts \pcite{du2023stable}. The scope of current evaluation and benchmark methods \pcite{huang2024vbench,liu2024evalcrafter,liu2023fetv,saito2020train,unterthiner2019fvd} is limited to primarily comparing the divergence of generated and real data distributions. Such comparisons do not reflect the extent to which the generated video conforms to the query, and how plausible the output is. The design of domain and characteristic-specific objectives is an interesting research direction. 

Another point of improvement for future works is the implementation of \textbf{generation control based on physical realism}. Although control of the generation process has been explored in many modalities \pcite{zhang2023adding}, the number of works that aim to impose modality-specific characteristics and dynamics constraints in the generation remains small. Video consistency and physics failure cases highlighted in \Cref{fig:dlm_fails} can be addressed by conditional terms that provide implicit information about the visual world. Such information could also be used explicitly by relying on physics simulations that reason about physical objects in the scene \pcite{liu2024physgen}.

The large capacity of generative models has shown great capabilities in simulating complex scenes. The impact of this has been shown in works that can simulate interactions of actors and objects both in the physical world \pcite{yang2023learning} and virtual renderings \pcite{alonso2024diffusion,valevski2024diffusion}. Understanding aspects of the world to generate video frames aligns with a number of downstream tasks that can enable \textbf{generative models to be used as general-purpose models}. 

\definecolor{vidred}{HTML}{f3a9a6}
\definecolor{temptasks}{HTML}{EA6B66}
\definecolor{vislang}{HTML}{ffd3aa}
\definecolor{mm}{HTML}{FFB570}
\definecolor{eap}{HTML}{d1a8c2}
\definecolor{vfp}{HTML}{B5739D}
\definecolor{states}{HTML}{c0d3f0}
\definecolor{vad}{HTML}{7EA6E0}
\definecolor{aa}{HTML}{aed2cc}
\definecolor{gen}{HTML}{67AB9F}

\begin{figure*}[t]
    \centering
    \begin{minipage}[c]{\linewidth}
    \includegraphics[width=\linewidth]{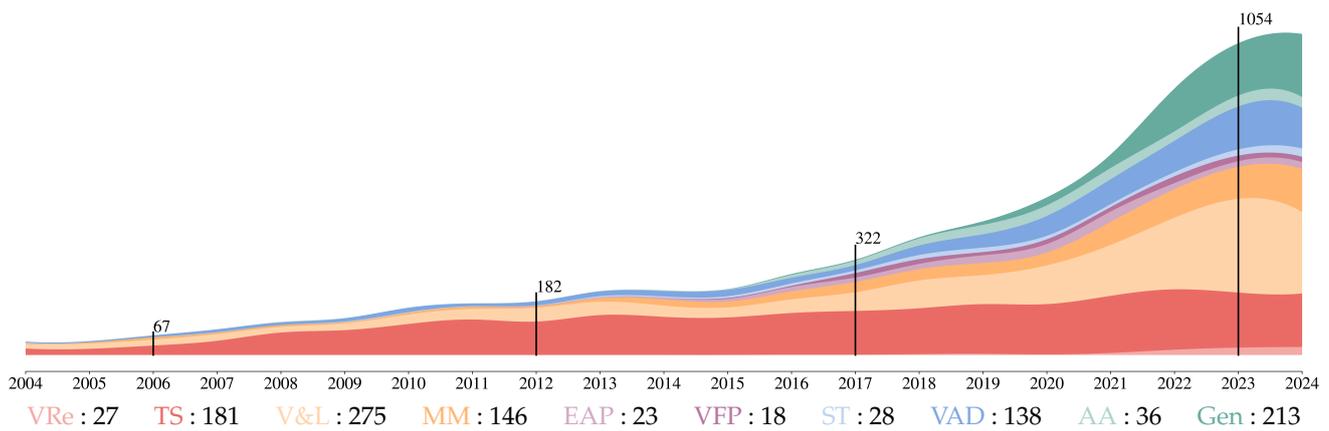}
    \resizebox{\linewidth}{!}{
    \begin{tabular}{c c c c c c c c c c}
         \textcolor{vidred}{VRe} : 27 &
         \textcolor{temptasks}{TS} : 181 &
         \textcolor{vislang}{V\&L} : 275 &
         \textcolor{mm}{MM} : 146 &
         \textcolor{eap}{EAP} : 23 &
         \textcolor{vfp}{VFP} : 18 &
         \textcolor{states}{ST} : 28 &
         \textcolor{vad}{VAD} : 138 &
         \textcolor{aa}{AA} : 36 &
         \textcolor{gen}{Gen} : 213
         \\
    \end{tabular}
    }
    \caption[Caption]{\textbf{Number of action understanding papers per year}. The research focus (bottom to top) includes video reduction approaches \textcolor{vidred}{VRe}, temporal tasks \textcolor{temptasks}{TS}, vision and language methods \textcolor{vislang}{V\&L}, multimodal models \textcolor{mm}{MM}, early action prediction \textcolor{eap}{EAP}, video frame prediction \textcolor{vfp}{VFP}, state-based tasks \textcolor{states}{ST}, video anomaly detection \textcolor{vad}{VAD}, action anticipation \textcolor{aa}{AA}, and video generation \textcolor{gen}{Gen}. The number of relevant papers is approximated from all works citing influential papers with $\geq 300$ citations per group\textcolor{red}{$^*$}. Small disparities are expected as recent works may not be included. The increasing research activity in action understanding is evident.}
    \label{fig:works_over_year}
    \end{minipage}
\end{figure*}

\section{Research directions to explore}
\label{sec:directions}

Progress in video understanding is fast-paced. As shown in \Cref{fig:works_over_year}, vision and language, video generation, and anomaly detection tasks have experienced significant interest in recent years. In tandem, well-established problems such as temporal tasks have remained relevant. We provide a look into the future and explore three main directions of progress beyond the continuation of current trends. We envision ways for future models to reason about abstractions in \Cref{sec:directions::reason}. We then consider the tasks and objectives that future action understanding models will address in \Cref{sec:directions::tasks}. Finally, we discuss efficiency improvements for training and deployment in \Cref{sec:directions::efficiency}.

\subsection{Reasoning semantics}
\label{sec:directions::reason}

With the shift from visual to semantic pattern extraction, the notion of abstraction levels will become more central. We explore future directions for interpreting actions, considering intentions and goals, and adapting to unseen scenarios. 

\subsubsection{From action to understanding}
\label{sec:directions::reason:::action2understanding}

Increasingly, action understanding is concerned not only with what is visually depicted but with reasoning about how the depiction is just one out of a multitude of possible perspectives. While action recognition tasks have driven a significant amount of progress on visual representations, future tasks will require more semantic interpretation. When moving from isolated clips of actions to longer episodes depicting behaviors, modeling long-range temporal dependencies becomes more important. Understanding behavior over time requires more than just aggregated interpretations of brief clips. Simultaneously, the distinction between visual observation and interpretation will become weaker, achieving a less deterministic view of action understanding with potentially multiple possible interpretations. Consequently, the automated analysis of videos will shift from objectively measuring or labeling, to a more subjective, context-dependent interpretation. In turn, this will require novel ways of training and evaluation, for example by including humans \pcite{kaufmann2023survey}.

One perspective on context is to include the intentions of those depicted. Despite VLMs' great progress in learning procedural steps in tasks through natural language-guided embeddings \pcite{li2025llama,li2024mini,wu2024longvideobench}, their reliance on visual information remains partial. \tcite{al2024unibench} showed that scaling models and data sizes do not offer substantial reasoning performance gains for vision tasks despite strong performance in skill-based tasks. Thus, a new avenue for future approaches is the design of open-world models from multi-level semantics that reflect human intentions and goals. Capturing information about the visual world in a scene may not necessarily require to be associated with language embeddings. Instead, relations could correspond to different pairs or groups of modalities adaptively. Longitudinal data specific to individuals may also be used to tailor the model's understanding of the world based on the user's goals, objectives, intentions, and interactions. Such treatment gives rise to a person- or case-specific perspective, bringing general action understanding to the consumer.

\blfootnote{\noindent \textcolor{red}{$^*$} From Google Scholar as of the 28th of October 2024.}

\subsubsection{Novel problem adaptation}
\label{sec:directions::reason:::novel}

Zero-shot performance has improved significantly over the past years. This is especially evident in language- and semantic-based video tasks aided by LLMs' large capacity and context. However, limitations remain in tasks orthogonal to pretext SSL \pcite{liu2024mmbench}. Recent approaches such as modality and probabilistic adapters \pcite{chen2024efficient,lin2023vision,sung2022vl,upadhyay2023probvlm,lu2024improving}, information gating \pcite{zhang2024llama}, visual prompt learning \pcite{khattak2023maple}, knowledge distillation \pcite{mistretta2024improving}, and model caching \pcite{zhang2021tip} have improved zero-shot downstream task performance by adjusting pre-trained models. However, only a few structural elements or objectives in models explicitly improve zero-shot performance for unseen distinct tasks. Unified models that can be used as mixture of experts controllers \pcite{bao2022vlmo,lin2024moe,wang2022image,yu2024boosting} are promising to bridge this gap. Sparsely trained models in mixture of experts settings allow for faster inference times with only task-relative sub-models using conditional computations \pcite{bengio2013estimating,jacobs1991adaptive}. When such mixtures can be linked to different levels of abstraction, re-use of models across levels also becomes feasible. The integration of experts can be done regardless of the backbone architecture.

\subsection{Better task definitions}
\label{sec:directions::tasks}

Training models does not necessarily guarantee that temporal dynamics of videos are learned at a foundational level. Future research may revisit and integrate beneficial properties for representation learning. Additionally, future works can explore performance measurements beyond simple metrics, instead focusing on explanation by observing embedding distributions and feature correspondences.

\subsubsection{Objectives}
\label{sec:directions::tasks:::objectives}

Distributions learned from representation-based objectives can be effectively used as priors to downstream tasks \pcite{janocha2017loss,larochelle2009exploring}. Drawing inspiration from \pcite{bengio2013representation}, a number of widely-accepted target properties are discussed below.

\noindent
\textbf{Temporal and spatial coherence}. Temporally or spatially proximal instances should correspond to similar representations. This notion can be extended to maintain proportional distances across both the pixel and embedding spaces. This coherence prior has been explored in objectives relating to time such as cyclic consistency \pcite{dwibedi2018temporal,donahue2024learning,haresh2021learning}, video procedural learning \pcite{chen2022frame,sermanet2018time}, DTW \pcite{dvornik2021drop,hadji2021representation}, and cross-frame stochasticity \pcite{zhang2023modeling}. These priors can be explored in a more general context as pre-training tasks similar to SSL while being tailored to the nature of videos. 

\noindent
\textbf{Abstractions and hierarchies}. Beyond fine-grained categories and semantics, most current works do not explicitly learn levels of abstraction. They are primarily limited to implicit connections between specific types \pcite{li2024deal} that often lead to spurious correlations \pcite{chen2020counterfactual,kim2023exposing,tian2024argue} as well as task- and instance-based misalignment \pcite{zhang2024rethinking}. Objectives that enforce abstraction hierarchies can potentially mitigate such misalignments. Promising efforts include partial order relations \pcite{alper2024emergent}, prototype learning \pcite{ramesh2022hierarchical}, hyperbolic representations \pcite{mettes2024hyperbolic}, and scene graphs \pcite{li2024scene}. As models become more polysemantic, the use of natural hierarchies and abstractions is expected to become more prevalent.

\noindent
\textbf{Natural clustering and manifolds}. Local representations tend to preserve similar polysemantic characteristics. Several works have shown that real data are not represented within the totality of the feature space but instead form dense concentrations in specific regions \pcite{genovese2012minimax,jiang2018trust,liang2022mind}. Using the tangent space of these distributions as a prior has shown promise in vision tasks such as generation \pcite{he2023manifold}, model explanation \pcite{bordt2023manifold}, anomaly detection \pcite{shin2023anomaly}, and corruption robustness \pcite{chen2022vita}. However, using the tangent space from real data distributions as an objective-steering prior remains largely an open question for large-scale multimodal action understanding models.

\subsubsection{Limitations in performance beyond metrics} \label{sec:directions:tasks:metrics}

Much of the progress in the domain of action understanding originates from comparing model outputs on benchmark data. While reported performance provides insights into the relative merits of models, it does not provide a good understanding of typical failure modes. Recent image-based \pcite{kowal2024understanding,kowal2024visual,park2023self,walmer2023teaching} and video-based \pcite{kowal2024understanding,stergiou2023leaping} visualization approaches provide human-interpretable insights into predictions at the instance level. However, understanding how semantic interpretations of actions are addressed, remains largely unexplored. This limits understanding the generalization ability to novel domains and tasks. Uptake of recent explainable AI trends \pcite{minh2022explainable} into computer vision model development can prompt the development of better measures for the capabilities and limitations of novel models.

Beyond benchmarking models on tasks and metrically evaluating performance, understanding the distributions and learned correlations provides new research opportunities. In-Context Learning (ICL) \pcite{brown2020language,hoffmann2022empirical} and Chain-of Thought (CoT) \pcite{wei2022chain} prompting are promising directions for LLMs and VLMs. \tcite{bansal2023rethinking} has shown that LLMs' capabilities are influenced by just a small number of attention or feed-forward layers, which are highly task-specific. Both \tcite{baldassini2024makes,chen2024understanding} showed that ICL in VLMs primarily relies on text information. Disparities between target and learned features can occur due to shortcuts learned by models. Common factors that can lead to shortcuts include contrastive loss' multiple local minima \pcite{robinson2021can}, suppression of visual information by language \pcite{li2023addressing}, and low mutual information between latent representations and real data \pcite{adnan2022monitoring}. Recently, \tcite{bleeker2024demonstrating} showed that introducing unique information distal to the overall training distribution favors VLMs' reliance on shortcuts for models trained on contrastive objectives. Such insights provide opportunities for exploring objectives and models with better multimodal and data-varying generalization capabilities.

\subsection{Efficiency}
\label{sec:directions::efficiency}

Model efficiency is essential for real-time application. Given the rapid deployment of models in a multitude of applications, we also highlight privacy risks alongside opportunities for domain specialization.  

\subsubsection{From research to deployment}
\label{sec:directions::efficiency:::deployment}

The increased variety of action understanding tasks also comes with the potential of improving actual deployment. Current and future models achieve performance and robustness levels that allow them to automate processes such as video data curation and surveillance. Novel applications based on behavior analysis can also benefit from these advances. Moving from benchmarks to the real world requires attention to computational efficiency. While the accuracy of current models is remarkable, performance comes at a cost. The trend of increasing model sizes, partly because of the focus on foundation models, largely prohibits the use of these models in computationally constrained operational settings. Attempts to reduce the computational complexity of trained models through pruning \pcite{iofinova2023bias}, knowledge distillation \pcite{mistretta2024improving}, or domain-specific adapters \pcite{hu2021lora} are not without limitations. The generalization performance gap between the currently best-performing models and those that can run on consumer hardware is significant. Several recent works \pcite{dao2022flashattention,gu2023mamba,poli2023hyena} propose novel processing paradigms that have the potential to scale better. Future work should address whether advances in multimodal training can transfer across both settings and models.

\subsubsection{Generalizable priors}
\label{sec:directions::efficiency:::priors}

Modern models are primarily trained on large-scale uncurated datasets aimed at multi-domain generalization. However, training distributions can include noise or be insufficiently rich for domain-specific datasets. Recent approaches have aimed to reduce training data requirements by including distribution priors at training. \tcite{kahana2022improving} aimed to improve zero-shot performance with a joint objective that matches label distributions while minimizing the divergence to original zero-shot predictions. \tcite{gao2022pyramidclip} utilized multiple levels of abstraction to contrast language and visual semantics and improve training efficiency. \tcite{nag2024safari} used a weakly-supervised approach to refine pseudo-object masks with cross-modal alignment in low-annotation settings. Approaches specifically utilizing priors in videos include concept distillation from normalized language embeddings \pcite{ranasinghe2023language}, and motion-specific alignment between video and textual descriptions of movements \pcite{zhang2024enhanced}. Distilling learned information to then be used as an optimization prior is a promising route for efficient training by reducing resource requirements. It can also impose a constraint based on the nature of expected motions with potential benefits in model convergence.

\subsubsection{Privacy and specialization}
\label{sec:directions::efficiency:::privacy}

Vision-based models are susceptible to attacks that either invert their gradients to reconstruct inputs \pcite{hatamizadeh2022gradvit} or discover intermediate representations \pcite{fang2023gifd}. Such attacks can compromise potentially proprietary training data, and reveal identifiable information. Inputs and features from VLMs can also be inferred through learnable vision-language triggers \pcite{bai2024badclip}, inference-time adversarial perturbations in frames \pcite{li2024fmm}, backdoor attacks through adversarial patches in training \pcite{carlini2022poisoning}, and injecting malicious prompts during instruction tuning \pcite{liang2024vl}. \tcite{kariyappa2023cocktail} showed that semantics from the original data can still be recovered even in distributed settings over large batches. Such vulnerabilities can be exploited across downstream tasks. Thus, evaluating model robustness is an important topic that the community should attend to.

The importance of privacy can also be understood through the current shift toward domain-expert sub-models integrated into general-purpose frameworks. \tcite{shen2024tag} showed that LLM specialization on domain-specific tasks significantly improves zero-shot generalization in related domains. Visual instructional tuning \pcite{bai2024generalist} and evolutionary instruction-based prompting \pcite{luo2024mmevol} have also shown promising results for vision-language models. Enhancing video-based models with domain specialization requires further exploration, for example through singular general models of high capacity, or multiple models in holistic frameworks.

\section{Conclusion}
\label{sec:conclusion}

Video action understanding includes a diverse set of tasks. These previously isolated tasks are increasingly overlapping in terms of the deployed models, utilized training data, and used evaluation protocols. To this end, we provide a comprehensive review of the broad domain of video action understanding. We discussed the main challenges, relevant datasets, and seminal works with an emphasis on recent (multimodal) advancements across tasks, and future research directions. We explicitly included multimodal advances. We focused on three temporal scopes from which tasks and approaches understand actions performed. We discussed recognition tasks that use complete observations of actions to infer fine- or coarse-grained labels. We then overviewed predictive tasks from partial observations of actions. Finally, we outlined forecasting tasks with anticipation models that infer general scene knowledge and forecast future actions not yet performed. Using time as a stepping stone, we outline current limitations and promising research directions to further advance the scope, robustness, and deployment of action understanding research.

\noindent
\textbf{Data availability}
We do not use or generate datasets. Dataset statistics used for comparisons are sourced from the respective papers referenced below.

{
\bibliography{egbib}
}

\end{document}